\documentclass{article}

    \PassOptionsToPackage{numbers, compress}{natbib}

\usepackage[main, final]{neurips_2026}

\usepackage[utf8]{inputenc} 
\usepackage[T1]{fontenc}    
\usepackage{hyperref}       
\usepackage{url}            
\usepackage{booktabs}       
\usepackage{amsfonts}       
\usepackage{nicefrac}       
\usepackage{microtype}      
\usepackage[dvipsnames,svgnames,x11names,hyperref]{xcolor}         
\usepackage{amsmath, amssymb, amsthm, mathtools}
\usepackage{algorithm}
\usepackage{algpseudocode}
\newtheorem{theorem}{Theorem}
\newtheorem{lemma}{Lemma}
\newtheorem{remark}{Remark}
\usepackage{thm-restate}

\newtheorem{assumption}{Assumption}
\usepackage{enumitem}
\usepackage{subcaption}
\usepackage{graphicx} 
\usepackage{multirow} 
\usepackage{booktabs} 
\usepackage{pifont}

\usepackage{thmtools}
\usepackage{thm-restate}
\usepackage{amsthm}
\usepackage[capitalise,nameinlink]{cleveref}
\usepackage[textsize=tiny]{todonotes}
\theoremstyle{definition}
\newtheorem{definition}[theorem]{Definition}
\usepackage[textsize=tiny]{todonotes}
\newcommand{\tO}{\widetilde{\mathcal{O}}} 
\crefname{assumption}{Assumption}{Assumptions}

\hypersetup{
    colorlinks=true,
    citecolor=NavyBlue,
    linkcolor=NavyBlue,
    urlcolor=NavyBlue,
}

\title{Provably Efficient Representation Learning for Low-Rank CMDPs}

\author{%
  Kaixuan Liu \\
  Emory University \\
  \texttt{kaixuan.liu@emory.edu} \\
  \And
  Guojun Xiong\thanks{Correspondence to Guojun Xiong \texttt{<gjxiong@sjtu.edu.cn>}.} \\
  Shanghai Jiao Tong University \\
  \texttt{gjxiong@sjtu.edu.cn} \\
  \And
  Shengpu Tang \\
  Emory University \\
  \texttt{shengpu.tang@emory.edu} \\
  \AND
  Wanyun Si \\
  University of Wisconsin-Madison \\
  \texttt{wsi7@wisc.edu} \\
  \And
  Jian Li \\
  Stony Brook University \\
  \texttt{jian.li.3@stonybrook.edu} \\
}

\begin{document}

\maketitle

\begin{abstract}
We study representation learning in low-rank Constrained Markov Decision Processes (CMDPs), where both value and reward functions are approximated by a set of unknown representation vectors, and the transition dynamics admit a low-rank factorization. The objective is to maximize expected cumulative rewards while complying with constraints on the expected cumulative utility. To achieve this, we propose \texttt{MFRLC} (Model-Free Representation Learning for Low-Rank CMDP), a model-free algorithm that uses a primal-dual scheme to effectively balance reward regret and constraint violations. To the best of our knowledge, \texttt{MFRLC} is the first representation-learning method for low-rank CMDPs that combines the Least-Squares Value Iteration with Upper Confidence Bound (LSVI-UCB) and primal-dual techniques. \texttt{MFRLC} further enhances value function estimation with bonuses for function approximation and learns the mapping from representations to value functions directly, avoiding explicit estimation of the transition kernel. We prove that, with high probability, \texttt{MFRLC} achieves $\widetilde O(K^{3/4})$ reward regret and cumulative constraint violation over $K$ deployment rounds, independent of the size of the state space. To our knowledge, these are the first sublinear regret and violation guarantees for CMDPs with unknown features.
\end{abstract}

\section{Introduction}

Safe Reinforcement Learning (SRL) has been widely studied and applied 
in real-world settings. For example, SRL addresses collision avoidance 
in autonomous driving \citep{ono2015chance} and resource allocation in 
healthcare \citep{yu2021reinforcement}, underscoring its importance in 
safety-critical domains. To address these challenges, a standard 
framework for SRL is the Constrained Markov Decision Processes (CMDPs). 
In a CMDP, the agent learns a policy that maximizes expected cumulative 
rewards while satisfying specified underlying constraints. Designing 
efficient learning-based algorithms for CMDPs has attracted considerable 
attention. Model-based methods~\citep{efroni2020exploration} learn an 
explicit CMDP model, but incur substantial storage and computational 
overhead in large-scale or continuous state spaces. Model-free methods 
avoid this: tabular variants~\citep{wei2023provably,ding2021provably} 
store value functions directly, while recent works adopt linear function 
approximation (LFA) to handle large state-action 
spaces~\citep{ghosh2022provably}, achieving provably near-optimal 
$\tO(K^{1/2})$ regret and violation rates.

A central limitation of existing LFA-based methods is the assumption 
that the underlying features are known \textit{a priori} 
\citep{ding2021provably,ghosh2022provably}. In practice, such features 
are typically unknown, rendering these methods inapplicable. Meanwhile, 
the low-rank MDP framework \citep{agarwal2020flambe} enables 
representation learning without prior feature knowledge, and provably 
efficient algorithms exist for the \emph{unconstrained} setting 
\citep{zhang2022efficient,modi2024model, mhammedi2023efficient}. However, extending representation learning to CMDPs is non-trivial. 
In the unconstrained setting, representation learning and policy 
optimization are largely decoupled; in a CMDP, the learned 
representation must simultaneously support both reward maximization 
and constraint satisfaction, and errors in the representation 
propagate into both objectives in a tightly coupled manner that 
standard primal-dual analyses cannot accommodate.
\begin{table*}[t]
\centering
\caption{Comparison with existing work, where $d$ denotes the feature 
dimension, $H$ the horizon, $K$ the number of deployment rounds, $\gamma$ the Slater gap, 
and $\varepsilon$ the target suboptimality (the learned policy's value is within 
$\varepsilon$ of optimal with high probability). Rates hide polylog factors via 
$\widetilde O(\cdot)$. \textbf{LFA} denotes linear function approximation; 
{\ding{55}} in the LFA column indicates tabular or non-parametric methods. Bounds stated in terms of $\varepsilon$ are PAC sample-complexity guarantees for unconstrained problems and are not directly comparable to regret.}
\label{tab:theory-comparison}
{\small
\setlength{\tabcolsep}{5.5pt}
\renewcommand{\arraystretch}{1.05}
\resizebox{\textwidth}{!}{%
\begin{tabular}{lcccccc}
\toprule
\textbf{Paper} & \textbf{Model Structure} & \textbf{LFA} & 
\textbf{Constraint} & \textbf{RepLearn} & 
\textbf{Regret / Sample Complexity} & \textbf{Violation} \\
\midrule
\citet{jin2020provably} & Linear & \ding{51} & \ding{55} & \ding{55} 
& $\tO(d^{3/2}H^2K^{1/2})$ & $N/A$ \\
\citet{ghosh2022provably} & Linear & \ding{51} & \ding{51} & \ding{55} 
& $\tO(d^{3/2}H^2K^{1/2}/\gamma)$ 
& $\tO(d^{3/2}H^2K^{1/2}/\gamma)$ \\
\citet{ding2023provably} & Tabular & \ding{55} & \ding{51} & \ding{55} 
& $\tO(d^{2/3}H^{5/2}K^{2/3}/\gamma)$ 
& $\tO(d^{2/3}H^{5/2}K^{2/3}/\gamma)$ \\
\citet{zhang2022efficient} & Block & \ding{51} & \ding{55} & \ding{51} 
& $\tO\big(d^{8}H^{9}\,\varepsilon^{-4}\big)$ & $N/A$ \\
\citet{modi2024model} & Low-rank & \ding{51} & \ding{55} & \ding{51} 
& $\tO\big(d^{7}H^{8}\,\varepsilon^{-2}\big)$ & $N/A$ \\
\citet{mhammedi2023efficient} & Low-rank & \ding{51} & \ding{55} & \ding{51} 
& $\tO\big(d^{13}|\mathcal{A}|^{2}H^{6}\,\varepsilon^{-2}\big)$ & $N/A$ \\
\midrule
\textbf{This Work} & \textbf{Low-rank} & \ding{51} & \ding{51} 
& \ding{51} 
& $\tO(d^{2}H^{3}K^{3/4}|\mathcal{A}|^{3/2}/\gamma)$ 
& $\tO(d^{2}H^{3}K^{3/4}|\mathcal{A}|^{3/2}/\gamma)$ \\
\bottomrule

\end{tabular}}
}\small
\setlength{\tabcolsep}{5.5pt}
\renewcommand{\arraystretch}{1.05}
\end{table*}

In this work, we bring representation learning to low-rank CMDPs and 
propose \texttt{MFRLC}, which in each round first learns 
representations via a minimax objective and then fits the value function 
using the learned features within a primal-dual LSVI-UCB framework. 
This constitutes the first approach for online low-rank CMDPs with unknown features. We prove that 
\texttt{MFRLC} achieves the same bound for both cumulative reward 
regret and cumulative constraint violation: 
$\tO(H^3 d^{2}K^{3/4}|\mathcal{A}|^{3/2}/\gamma)$, where $K$ is the 
number of deployment rounds, $H$ is the horizon, $d$ is the representation 
dimension, and $\gamma$ is the Slater gap. The $|\mathcal{A}|^{3/2}$ factor arises from uniform exploration over
actions in the representation learning step, and the $K^{3/4}$
dependence arises from the accumulated statistical error of
representation learning (Remark~\ref{rem:rate}). Both bounds are upper
bounds. No matching lower bound is known for online low-rank MDPs with
unknown features, with or without constraints, so we do not claim that
the $K^{3/4}$ rate or the dependence on $d$, $H$, and $|\mathcal{A}|$ is
optimal. Known-feature CMDPs admit $\tO(K^{1/2})$ regret and
violation~\citep{ghosh2022provably}; whether this rate is attainable with
unknown features remains open.

\section{Problem Formulation}
\label{sec:formulation}

\subsection{CMDP Preliminaries}

We consider an episodic CMDP $\langle\mathcal{S},\mathcal{A},H, r,g, \mathbb{P},d_0\rangle$, where $\mathcal{S}$ is the state space (possibly infinite or continuous), $\mathcal{A}$ the action space (finite, with cardinality $|\mathcal{A}|$), $H$ the fixed episode length, $r=\{r_h\}_{h=1}^H$ and $g=\{g_h\}_{h=1}^H$ the collections of reward and utility functions, respectively, with $r_h, g_h:\mathcal{S}\times\mathcal{A}\to[0,1]$ assumed to be deterministic, $\mathbb{P}=\{P_h\}_{h=1}^H$ the collection of transition kernels with $P_h(s_{h+1}\mid s_h,a_h):\mathcal{S}\times\mathcal{A}\times\mathcal{S}\to[0,1]$ being the probability of moving to $s_{h+1}$ from $s_h$ under $a_h$, and $s_1\sim d_0$ the initial state of each episode.
At each step $h\in[H]$ of round $k\in[K]$, the agent takes action $a_h^k$ at state $s_h^k$, receives reward $r_h(s_h^k,a_h^k)$ and utility $g_h(s_h^k,a_h^k)$, and transitions to $s_{h+1}^k\sim P_h(\cdot\mid s_h^k,a_h^k)$; each episode terminates after $H$ steps. The policy space is $\Delta(\mathcal{A}\mid\mathcal{S},H):=\{\{\pi_h(\cdot\mid\cdot)\}_{h=1}^H:\pi_h(\cdot\mid s)\in\Delta(\mathcal{A}),\,\forall s\in\mathcal{S},\,h\in[H]\}$, where $\Delta(\mathcal{A})$ is the probability simplex over the action space; $\pi^k\in\Delta(\mathcal{A}\mid\mathcal{S},H)$ denotes the policy deployed in round $k$, with $\pi_h^k(\cdot\mid s_h^k):\mathcal{S}\to\Delta(\mathcal{A})$ specifying the action distribution at $s_h^k$. Given $\pi=\{\pi_h\}_{h=1}^H$, the value function (expected cumulative reward from $s_h=s$ under $\pi$) and action-value function (from $(s_h=s,a_h=a)$) are:
\begin{align*}
V^{\pi}_{r,h}(s) = \mathbb{E}_{\pi}\!\left[\sum_{i=h}^H r_i(s_i, a_i) \,\Big|\, s_h = s\right],
\qquad
Q_{r,h}^\pi(s,a) = \mathbb{E}_\pi\!\left[\sum_{i=h}^H r_i(s_i, a_i) \,\Big|\, s_h=s,\, a_h=a\right],
\end{align*}
where the expectation is over $a_i\sim\pi_i(\cdot\mid s_i)$ and $s_{i+1}\sim P_i(\cdot\mid s_i,a_i)$. The utility value and action-value functions $V^{\pi}_{g,h}(s)$, $Q^{\pi}_{g,h}(s,a)$ are defined analogously by replacing $r$ with $g$; for brevity, we write $V^{\pi}_{j,h}(s)$ and $Q^{\pi}_{j,h}(s,a)$ for $j\in\{r,g\}$.

We say a policy $\pi$ is \emph{feasible} if $V^\pi_{g,1}(s_1)\geq b$. The agent aims to learn the optimal feasible policy $\pi^\star$ by solving the following optimization problem:
\begin{equation}\label{eq:obj}
\max_{\pi\in\Delta(\mathcal{A} \mid \mathcal{S},H)} V_{r,1}^\pi(s_1), \quad \text{subject to} \quad V_{g,1}^\pi(s_1)\geq b,
\end{equation}
where the threshold $b \in [0,H]$. Under the online learning setting of an episodic CMDP, the agent interacts with the environment over a sequence of rounds, and the learning performance is measured by regret and constraint violation as follows:
\begin{align*}
    \mathrm{Regret}(K) = \sum_{k=1}^K \left( V^{\pi^\star}_{r,1}(s_1) - V^{\pi^k}_{r,1}(s_1) \right), \quad 
    \mathrm{Violation}(K) := \Big(\sum_{k=1}^K \big(b - V^{\pi^k}_{g,1}(s_1)\big)\Big)_+,
\end{align*}
with $\pi^k$ being the policy deployed in round $k$. Both quantities are random variables depending on the algorithm's internal randomness and the sampled trajectories; our guarantees on them hold with high probability (see Theorem~\ref{thm:mfrlc}). The regret quantifies the difference in total rewards achieved by the agent's policy $\pi^k$ compared to the optimal feasible policy $\pi^\star$ across $K$ rounds, and the constraint violation measures the cumulative shortfall of utility against the predefined threshold $b$ over $K$ rounds.

\paragraph{Access model and interaction accounting.}
We call each index $k$ a \emph{deployment round}. In round $k$, \texttt{MFRLC} executes $\pi^k$ for one trajectory and also collects two auxiliary transitions at every step $h$ for representation learning (Algorithm~\ref{alg:mfrlc}, line~4), each obtained by rolling in with $\pi^{k-1}$ to step $h$ and taking a uniform action, so no generative model is required. Regret and Violation contain one term per deployed policy and do not charge the auxiliary roll-ins; our guarantees are deployment-regret guarantees with total interaction complexity $K(2H+1)$ trajectories and $K(H^2+2H)$ transitions.

\subsection{Low-Rank MDP and Representation Learning}

Conventional work on learning MDPs typically assumes a general linear MDP framework \citep{jin2020provably,melo2007q,azizzadenesheli2018efficient}, where features are considered known, and function approximation methods are used to maximize rewards without considering constraints. While this linear MDP assumption has been effectively extended to CMDPs \citep{amani2021safe,ghosh2023achieving}, the transition to low-rank MDPs introduces additional layers of complexity, defined as follows.

\begin{definition}[Low-rank MDP \citep{agarwal2020flambe}]
\label{def:lowrank}
Consider a transition kernel $P_h^\star: \mathcal{S} \times \mathcal{A} \to \Delta(\mathcal{S})$ which can be decomposed into low-rank components of dimension $d \in \mathbb{N}$. Specifically, there exist functions $\phi^\star=\{\phi_h^\star\}_{h=1}^H$ and $\mu^\star=\{\mu_h^\star\}_{h=1}^H$ with $\phi_h^\star:\mathcal{S}\times\mathcal{A}\to\mathbb{R}^d$ and $\mu_h^\star: \mathcal{S} \to \mathbb{R}^d$, such that for all $s, s' \in \mathcal{S}$, $a \in \mathcal{A}$, and $h\in[H]$:
\begin{align*}
 P_{h}^\star(s'\mid s,a)=\langle\phi_{h}^\star(s,a), \mu_{h}^\star(s')\rangle.
\end{align*}
Furthermore, there exist vectors $\theta^\star_{r,h}, \theta^\star_{g,h} \in \mathbb{R}^d$ such that for any $(s,a)\in\mathcal{S}\times\mathcal{A}$:
\begin{align*}
r_h(s,a)=\langle\phi_{h}^{\star}(s,a),\theta^{\star}_{r,h}\rangle,\quad g_h(s,a)=\langle\phi_{h}^{\star}(s,a),\theta^{\star}_{g,h}\rangle.
\end{align*}
The feature function satisfies $\|\phi_{h}^\star(s, a)\|_2 \leq 1$ for all state-action pairs $(s, a)$, and for any function $f: \mathcal{S} \to [0, 1]$, $\left\|\int \mu_h^\star(s) f(s) \, \mathrm{d}s \right\|_2 \leq \sqrt{d}$. An MDP $\mathcal{M}$ is considered low-rank if the transition kernel $P_h$ at each time step $h \in [H]$ admits such a decomposition.
\end{definition}

Under the low-rank setting, the feature representations $\phi^\star$ are no longer presumed known. This necessitates the development of algorithms that are not only capable of learning these features but also ensure that the resulting policy adheres to the constraints imposed by the CMDP framework.

\begin{assumption}[Realizable feature class]
\label{assum:realiz}
The learner has access to a finite feature class $\Phi=\Phi_1\times\cdots\times\Phi_H$, where each $\phi_h\in\Phi_h$ maps $\mathcal{S}\times\mathcal{A}$ to $\mathbb{R}^d$ with $\|\phi_h(s,a)\|_2\leq 1$ for all $(s,a)$, and $\Phi$ contains the true representation of Definition~\ref{def:lowrank}, that is, $\phi^\star\in\Phi$.
\end{assumption}
Assumption~\ref{assum:realiz} is the standard setting of model-free representation learning in low-rank MDPs~\citep{agarwal2020flambe,modi2024model}. Finiteness enters only through $\log|\Phi|$ and can be replaced by a covering number, while realizability is exact: misspecified classes with $\phi^\star\notin\Phi$ are outside our guarantees. Requiring the utility, like the reward, to be linear in $\phi^\star$ is the natural CMDP counterpart of the low-rank MDP assumption; it is structural and does not cover reward or utility functions unrelated to the low-rank dynamics. We do not assume any map from representations to policies recovers $\pi^\star$: the analysis compares directly against an arbitrary optimal feasible policy of~\eqref{eq:obj}.
Importantly, we adopt a model-free approach in which the policy is 
learned without explicitly modeling the environment, and there is no 
need to recover the exact values of $\mu^\star$ or $\theta^\star$.

\section{Representation Learning for Low-Rank CMDPs}
\label{sec:replearn}

In this section, we address the challenge of model-free representation learning within low-rank CMDPs. Our algorithm, \texttt{MFRLC} (Algorithm~\ref{alg:mfrlc}), builds on the Upper Confidence Bound (UCB) variant of Least-Squares Value Iteration (LSVI) and utilizes a primal--dual approach to effectively manage the CMDP's optimization requirements.

Before delving into the details of the algorithm, it is helpful to introduce the \emph{composite} (Lagrangian) value function, which encodes the reward--constraint trade-off into a single scalar objective.
\begin{definition}[Composite value function; cf.\ \citealp{paternain2019constrained,ding2021provably}]\label{def2}
For any dual variable $Y\geq 0$ and policy $\pi$, define the composite value function and action-value function as
\begin{align*}
V_h^{\pi,Y}(\cdot)=V_{r,h}^\pi(\cdot)+Y\,V_{g,h}^\pi(\cdot),\quad
Q_{h}^{\pi,Y}(s,a)=Q_{r,h}^{\pi}(s,a)+Y\,Q_{g,h}^{\pi}(s,a).
\end{align*}
Equivalently, $V^{\pi,Y}_h$ and $Q^{\pi,Y}_h$ are the value and action-value functions of policy $\pi$ in the auxiliary MDP whose stage reward is the Lagrangian $r_h + Y g_h$.
\end{definition}
We operate on the composite values $(V_h^{\pi,Y},Q_h^{\pi,Y})$ throughout the analysis, with $Y$ instantiated to the dual iterate $Y_k$ at round $k$. Algorithm~\ref{alg:lsvi} adds the same per-objective bonus $\hat b_h$ to both $Q^k_{r,h}$ and $Q^k_{g,h}$, so the composite optimism bonus is $\hat b^{(k)}_h:=(1+Y_k)\hat b_h$.

\setlength{\textfloatsep}{8pt plus 2pt minus 2pt}
\setlength{\floatsep}{8pt plus 2pt minus 2pt}

\subsection{Algorithm Overview}

\begin{algorithm}[tb]
\caption{Model-Free Representation Learning with Low-rank CMDP (\texttt{MFRLC})}
\label{alg:mfrlc}
\begin{algorithmic}[1]
   \Require Feature classes $\{\Phi_h\}_{h=1}^H$, regularization $\lambda_k$, Slater parameter $\gamma>0$
   \Ensure Sequence of policies $\{\pi^k\}_{k=1}^K$
   \State \textbf{Initialize:} $Y_1=0$, $\xi=2H/\gamma$, $\eta=\xi/\sqrt{KH^{2}}$;
          set $\pi^0$ to any reference policy and replay buffers
          $\mathcal{D}_h=\varnothing$, $\mathcal{D}_h'=\varnothing$ for all $h\in[H]$.
   \For{$k=1$ \textbf{to} $K$}
      \For{$h=1$ \textbf{to} $H$} \Comment{Exploration: branching from $d_h^{\pi^{k-1}}$}
         \State Sample $s_h^k\sim d_h^{\pi^{k-1}}$,
                $\tilde a_h^k\sim U(\mathcal{A})$,
                $s_{h+1}^k\sim P^\star_h(\cdot\mid s_h^k,\tilde a_h^k)$,
                $\tilde s\sim d_h^{\pi^{k-1}}$,
                $\tilde a\sim U(\mathcal{A})$,
                $\tilde s'\sim P^\star_h(\cdot\mid \tilde s,\tilde a)$
         \State $\mathcal{D}_h \gets \mathcal{D}_h \cup \{(s_h^k,\tilde a_h^k,s_{h+1}^k)\}$,\quad
                $\mathcal{D}_h' \gets \mathcal{D}_h' \cup \{(\tilde s,\tilde a,\tilde s')\}$
         \State $\hat{\phi}_h^k \gets
                \textsc{RepLearn}\big(\mathcal{D}_h,\mathcal{D}_h',\Phi_h,\lambda_k\big)$
                \Comment{Algorithm~\ref{alg:replearn}}
      \EndFor
      \State $\big\{Q_{j,h}^k,V_{j,h}^k\big\}_{j\in\{r,g\},h\in[H]},\,
             \{\pi_h^k\}_{h=1}^H \gets
             \textsc{LSVI-UCB}\big($
      \Statex \hspace{2.2em}$\big\{(s_h^\tau,\tilde a_h^\tau,r_h^\tau,g_h^\tau,
             s_{h+1}^\tau,\hat\phi_h^\tau)\big\}_{h\in[H],\,\tau\le k},\,Y_k\big)$
      \For{$h=1$ \textbf{to} $H$} \Comment{Execute $\pi^k$ for next-episode trunk}
         \State Take $a_h^k\sim \pi_h^k(\cdot\mid s_h^k)$ and observe $s_{h+1}^k$
      \EndFor
      \State $Y_{k+1} \gets \max\!\big\{\min\!\big\{Y_k+\eta\,(b-V_{g,1}^k(s_1)),\,\xi\big\},\,0\big\}$
             \Comment{Primal--dual update}
   \EndFor
\end{algorithmic}
\end{algorithm}

\texttt{MFRLC} alternates three steps in each round. 
\textbf{Representation Learning} (lines 4--6) samples under $d_h^{\pi^{k-1}}$ 
with kernel $P_h^\star$, stores data in replay buffers $\mathcal D_h, \mathcal D_h'$, 
and learns $\hat\phi_h^k\in\Phi_h$ via a minimax game against a discriminator 
$v\in\mathcal V_h$, controlled by the regularizer $\lambda_k$, iteration cap $M$, 
and stopping threshold $\ell_k$. 
\textbf{Policy Update} (line 8) takes $\hat\phi^k$ and $Y_k$ and runs LSVI-UCB 
(Algorithm~\ref{alg:lsvi}, Appendix~\ref{app:lsvi}) to produce optimistic per-objective 
estimates $\{Q^k_{r,h}, Q^k_{g,h}, V^k_{r,h}, V^k_{g,h}\}$ and the softmax policy
\begin{equation}\label{eq:softmax}
\pi^k_h(a\mid \cdot) \;=\; \mathrm{softmax}_\alpha\!\left(Q^k_{r,h}(\cdot,a)+Y_k\,Q^k_{g,h}(\cdot,a)\right),
\end{equation}
where $Y_k$ enforces the constraint while reward is maximized. 
\textbf{Primal--Dual Update} (line 12) updates $Y_k$ using the optimistic value 
$V_{g,1}^{k}(s_1)$ in place of $V_{g,1}^{\pi^k}(s_1)$, based on the observed 
constraint slack, and carries it into the next round.

\begin{algorithm}[tb]
\caption{Representation Learning (\textsc{RepLearn})}
\label{alg:replearn}
\begin{algorithmic}[1]
   \Require Feature class $\Phi_h$, datasets $\mathcal{D}_h, \mathcal{D}_h'$, discriminator class $\mathcal{V}_h$, regularization $\lambda_k$, iterations $M$, threshold $\ell_k$, confidence $\delta$, dimension $d$.
   \Ensure Representation $\hat\phi_h^k$
   \State \textbf{Parameters:}
          $\lambda_k = O\!\left(d\ln\tfrac{|\Phi|k}{\delta}\right)$,\
          $\ell_k = O\!\left(d^2\sqrt{\tfrac{\ln(|\Phi|/\delta)}{k}}\right)$
   \State \textbf{Notation:} $\mathcal{L}_{\lambda_k,\mathcal{D}}(\phi,w,v)$ is the empirical objective~\eqref{eq:emploss} over buffer $\mathcal{D}$; $w$ (and $\tilde w$) are predictor weights for feature $\phi$; $v$ ($v^m$) $\in\mathcal{V}_h$ is the discriminator (at step $m$).
   \State \textbf{Initialize:} $\phi^0 \in \Phi_h$ arbitrarily
   \For{$m = 0$ \textbf{to} $M-1$}
      \State Compute worst-case discriminator with bias correction:
      \State $\displaystyle v^m \gets \operatorname*{arg\,max}_{v\in\mathcal{V}_h,\,\tilde\phi\in\Phi_h}\Big[\min_w \mathcal{L}_{\lambda_k,\mathcal{D}_h}(\phi^m,w,v) - \min_{\tilde w}\mathcal{L}_{\lambda_k,\mathcal{D}_h'}(\tilde\phi,\tilde w,v)\Big]$
      \If{the maximization objective at $v^m$ is at most $\ell_k$}
         \State \textbf{Break} and \textbf{Return} $\phi^m$
      \EndIf
      \State Update representation jointly against past discriminators:
      \State \hspace{1em}$\displaystyle \phi^{m+1} \gets
             \operatorname*{arg\,min}_{\phi\in\Phi_h,\,\{w_i\}_{i=0}^{m}}
             \sum_{i=0}^{m}\mathcal{L}_{\lambda_k,\mathcal{D}_h}(\phi,w_i,v^i)$
   \EndFor
\end{algorithmic}
\end{algorithm}

\paragraph{Representation Learning}\label{subsec:replearn}

In the context of low-rank MDPs, where $\phi^\star$ is unknown, we 
must learn this representation. Inspired by \citet{zhang2022efficient}, 
we define $\mathcal{V}_h = \mathcal{F}_h \cup \mathcal{G}_h$, where 
$\mathcal{F}_h$ captures the model-mismatch between $\hat{P}$ and 
$P^\star$ under the previous-episode policy $\pi^{k-1}$ (Lemma~\ref{lem:optimism}) 
and $\mathcal{G}_h$ captures the composite $Q$-function (Lemma~\ref{lem:policygap}):
\begin{align}\label{eq:discF}
&\mathcal{F}_h := \Big\{ s\mapsto\mathbb{E}_{a \sim \pi^{k-1}_{h}(\cdot|s)}\!\!\left[ \langle \phi(s, a), \theta \rangle - \langle \phi'(s, a), \theta' \rangle \right] : \phi, \phi' \in \Phi_{h+1},\ \max(\|\theta\|_\infty, \|\theta'\|_\infty) \leq 1 \Big\},
\end{align}

\begin{align}\label{eq:discG}
\mathcal{G}_h := \bigg\{ s\mapsto \max_{a\in\mathcal A}\bigg(\frac{r_{h+1}(s,a)+g_{h+1}(s,a)+\min\{\langle\phi(s,a),w\rangle,2\}}{2H+1} + \langle\phi(s,a),w'\rangle\bigg) \bigg\},
\end{align}
where the set ranges over $\phi \in \Phi_{h+1}$, $\|w\|_\infty \leq c$, $\|w'\|_\infty \leq 1$. Reward and utility enter the representation objective through $\mathcal{G}_h$, which contains $r_{h+1}+g_{h+1}$; our guarantee requires no separate supervised loss on the immediate reward or utility.

To learn $\phi^\star$, we solve the following optimization problem, where $\rho_h^k$ denotes the sampling distribution induced by the replay buffer $\mathcal{D}_h$ at the start of episode $k$:
\begin{align} \label{eq:repobj}
\arg\min_{\phi^k_h}\;\max_{v\in\mathcal V_h}\;\min_{w}\;\mathbb{E}_{(s,\tilde a)\sim\rho_h^k}\!\Big[\big(\langle\phi^k_h(s,\tilde a),w\rangle-\,\mathbb{E}_{s'\sim P^\star_h(\cdot|s,\tilde a)}\![v(s')]\big)^2\Big].
\end{align}
This objective seeks a feature $\phi$ that accurately approximates the Bellman backup $\mathbb E[v(s')|s,\tilde a]$ as a linear function $\langle\phi(s,\tilde a), w\rangle$, uniformly over $v\in\mathcal V_h$. The min-max-min structure ensures robustness: the inner $\min_w$ is a regularized least-squares problem with a closed-form solution (Appendix~\ref{app:replearn}, complexity $\mathcal O(nd^2+d^3)$); the middle $\max_v$ identifies the worst-case $v\in\mathcal V_h$ for the current $\phi$; the outer $\min_{\phi^k_h}$ chooses the representation minimizing this worst-case error \citep{modi2024model}. The non-linear softmax operator and UCB bonus appearing in the LSVI-UCB output (Algorithm~\ref{alg:lsvi}) are not contained in $\mathcal{G}_h$; their contribution is handled separately via the softmax sub-optimality bound (Lemma~\ref{lem:proof-softmax}) and the elliptical-potential argument applied to the bonus term (Appendix~\ref{app:pseudoregret}).

Since~\eqref{eq:repobj} involves a population expectation, direct implementation is challenging. We replace it by its sample-based counterpart over $\mathcal D_h$:
\begin{align}\label{eq:emploss}
\mathcal{L}_{\lambda_k,\mathcal{D}_h}(\phi^k_h,v,w)
&= \mathbb{E}_{\mathcal{D}_h}\!\Big[\big(\langle\phi^k_h(s,\tilde a),w\rangle - v(s')\big)^2\Big]+\lambda_k\|w\|_2^2.
\end{align}
Because $\mathcal{L}_{\lambda_k,\mathcal{D}_h}$ contains a conditional-variance term that depends on $v$, maximizing it directly would favor high-variance discriminators. Algorithm~\ref{alg:replearn} cancels this term in expectation by subtracting the minimized loss $\mathcal L_{\lambda_k,\mathcal D_h'}(\tilde\phi,\tilde w,v)$ on the independent buffer $\mathcal{D}'_h$ (Appendix~\ref{app:replearn}).

\paragraph{From theory to implementation}
We distinguish three levels of the representation-learning step:
\[
\text{objective~\eqref{eq:repobj}}
\;\longrightarrow\;
\text{bias-corrected \textsc{RepLearn} (Algorithm~\ref{alg:replearn})}
\;\longrightarrow\;
\text{contrastive EMA surrogate}.
\]
Theorem~\ref{thm:mfrlc} covers the first two levels. The experiments in Section~\ref{sec:experiments} use the third: the minimax search over $\mathcal{V}_h$ is replaced by contrastive next-state prediction with an exponential-moving-average (EMA) target encoder. The surrogate uses the same transition triples $(s,\tilde a,s')$ to learn predictive features, but it does not perform the worst-case discriminator search and does not certify uniform Bellman prediction over $\mathcal{V}_h$, so it is not covered by Theorem~\ref{thm:mfrlc}.

\paragraph{Computation}
The inner minimization over $w$ is ridge regression with cost $O(nd^2+d^3)$; the outer problems over $\phi$ and $v$ are treated as optimization oracles, exactly solvable for finite classes but without a polynomial-time guarantee for neural classes (Appendix~\ref{app:replearn-cost}).

\paragraph{Policy Update}

Given $\hat\phi^k$ and $Y_k$, the LSVI-UCB subroutine (Algorithm~\ref{alg:lsvi}, Appendix~\ref{app:lsvi}) computes optimistic estimates $Q^k_{j,h}, V^k_{j,h}$ for $j\in\{r,g\}$ and the softmax policy~\eqref{eq:softmax}. The executed policy $\pi^k$ is defined directly as this optimistic softmax policy computed from $\hat\phi^k$ and $Y_k$. The composite $Q^k_{r,h} + Y_k Q^k_{g,h}$ is the LSVI-UCB optimistic estimate of the Bellman-optimal $Q$ for the auxiliary MDP with stage reward $r_h + Y_k g_h$ (cf.\ Definition~\ref{def2}), with $Y_k$ updated online in the next step.

\paragraph{Dual Update}
Problem~\eqref{eq:obj} admits the Lagrangian formulation:
\begin{equation}\label{eq:saddle}
\max_{\pi\in\Delta(\mathcal{A}\mid\mathcal{S},H)}\,\min_{Y\in[0,\xi]}\;V^{\pi}_{r,1}(s_1)+Y\big(V^{\pi}_{g,1}(s_1)-b\big),
\end{equation}
where $\xi=2H/\gamma$ is chosen so that $Y^\star\in[0,\xi]$ (Lemma~\ref{lem:dualbound}); restricting $Y$ to $[0,\xi]$ does not affect the saddle-point value. \texttt{MFRLC} does not solve~\eqref{eq:saddle} over $\pi$ directly. In round $k$, the primal step is the optimistic LSVI-UCB policy $\pi^k$ for the Lagrangian reward $r_h+Y_kg_h$, and the dual step is the projected update:
\begin{equation}\label{eq:dualupdate}
Y_{k+1}\;=\;\max\!\big\{\min\!\big\{Y_k+\eta\,(b-V^{k}_{g,1}(s_1)),\,\xi\big\},\,0\big\},
\end{equation}
which uses the optimistic estimate $V^{k}_{g,1}(s_1)$ in place of $V^{\pi^k}_{g,1}(s_1)$. The update direction switches with the sign of $b-V^{k}_{g,1}(s_1)$, balancing reward maximization against feasibility \citep{efroni2020exploration,nedic2009subgradient}.

\section{Main Results}
\label{sec:analysis}

\begin{assumption}[Slater's Condition]\label{assum:slater}
There exists $\gamma>0$ and a policy $\bar\pi\in\Delta(\mathcal{A}\mid\mathcal{S},H)$
such that $V_{g,1}^{\bar\pi}(s_1)\geq b+\gamma$.
\end{assumption}

\begin{lemma}[Boundedness of $Y^\star$]\label{lem:dualbound}
The optimal dual variable $Y^\star$ associated with~\eqref{eq:obj} satisfies
$Y^\star\leq (V_{r,1}^{\pi^\star}(s_1)-V_{r,1}^{\bar\pi}(s_1))/\gamma\leq H/\gamma$,
so $\xi=2H/\gamma\geq 2Y^\star$ as required by Lemma~\ref{lem:aux-strongdual}.
\end{lemma}

The Slater condition is mild and widely adopted
\citep{paternain2019constrained,ding2021provably,efroni2020exploration}.
\begin{theorem}\label{thm:mfrlc}
Fix $p\in(0,1)$. Under Assumptions~\ref{assum:realiz}--\ref{assum:slater},
with probability at least $1-p$, the regret and constraint violation of
\texttt{MFRLC} over $K$ deployment rounds both satisfy:
\begin{align*}
\mathrm{Regret}(K),\,\mathrm{Violation}(K)
\;\leq\; \frac{C}{\gamma}\,H^3 d^2 K^{3/4}|\mathcal{A}|^{3/2}\,
\log^{1/4}\!\left(\frac{dK|\Phi|}{p}\right),
\end{align*}
for an absolute constant $C>0$. Over the $K$ rounds, \texttt{MFRLC} uses
$K(2H+1)$ trajectories in total, including the auxiliary roll-ins of
representation learning, whose rewards and utilities are not charged to
Regret or Violation (Section~\ref{sec:formulation}).
\end{theorem}
The dependence on $1/\gamma$ is the standard Slater-gap penalty of
primal--dual CMDP analyses~\citep{ding2020natural,ding2021provably,ghosh2022provably}.
The bound is independent of $|\mathcal{S}|$ and applies to
continuous state spaces. The $K^{3/4}$ term arises only from the
representation-learning error (Remark~\ref{rem:rate} in
Appendix~\ref{app:violation}). Theorem~\ref{thm:mfrlc} is an upper bound:
we prove no matching lower bound, and the optimal dependence on $K$, $d$, $H$, and
$|\mathcal{A}|$ remains open.
\subsection{Proof Sketch}

\paragraph{Step 1: Decomposition of regret and violation.}
For any $Y\in[0,\xi]$, by combining a Lagrangian shift with the projected dual
update of Algorithm~\ref{alg:mfrlc} (Appendix~\ref{app:decomp}), we obtain:
\begin{align}
&\sum_{k=1}^K\!\big(V_{r,1}^{\pi^\star}(s_1)-V_{r,1}^{\pi^k}(s_1)\big)
+Y\sum_{k=1}^K\!\big(b-V_{g,1}^{\pi^k}(s_1)\big)
\;\leq\; \frac{Y^2}{2\eta}+\frac{\eta H^2 K}{2}\label{eq:decomp}\\
&\quad+\underbrace{\sum_{k=1}^K\!\Big[
  \big(V_{r,1}^{\pi^\star}(s_1)+Y_k V_{g,1}^{\pi^\star}(s_1)\big)
  -\big(V_{r,1}^k(s_1)+Y_k V_{g,1}^k(s_1)\big)
\Big]}_{\text{Pseudo regret}}\notag\\
&\quad+\underbrace{\sum_{k=1}^K\!\Big[
  \big(V_{r,1}^k(s_1)-V_{r,1}^{\pi^k}(s_1)\big)
  +Y\big(V_{g,1}^k(s_1)-V_{g,1}^{\pi^k}(s_1)\big)
\Big]}_{\text{Estimation error}}.\notag
\end{align}
Here $Y$ is a free variable that combines reward regret and constraint violation:
setting $Y=0$ recovers the regret bound, while $Y=\xi$ (combined with strong duality)
yields the violation bound. By feasibility of $\pi^\star$, the Lagrangian-shift slack
term $\sum_k Y_k(b-V_{g,1}^{\pi^\star}(s_1))\leq 0$ is non-positive and dropped. The
Pseudo regret captures the gap between $\pi^\star$ and the LSVI-UCB optimistic
estimate $V^k$, while the Estimation error captures the gap between $V^k$ and the
executed policy's value $V^{\pi^k}$. We control the two terms separately.

\paragraph{Step 2: Bounding the Pseudo regret.}
Since $\pi^k$ is the softmax of the composite $Q^k_{r,h}+Y_k Q^k_{g,h}$, the
analysis must be carried out on the auxiliary MDP with stage reward
$\bar r^{(k)}_h$. We first establish composite optimism on the planning model
$\hat P^k$ induced by $\hat\phi^k$.

\begin{lemma}[Composite optimism]\label{lem:optimism}
Conditioned on the Representation Learning guarantee
(Lemma~\ref{lem:aux-replearn}), with probability $1-\delta$, for every round
$k\in[K]$,
\[
Q^{\pi^\star,Y_k}_{\hat P^k,\,\bar r^{(k)}+\hat b^{(k)}}(s_1,\cdot)
\;-\;Q^{\pi^\star,Y_k}_{P^\star,\,\bar r^{(k)}}(s_1,\cdot)
\;\geq\; -(1+Y_k)\sqrt{|\mathcal{A}|\zeta_k},
\]
where $\zeta_k=\widetilde O(d^2/\sqrt{k})$ is the RepLearn statistical error.
\end{lemma}

\begin{lemma}[Composite policy gap, cumulative form]\label{lem:policygap}
Conditioned on the Representation Learning guarantee, with probability $1-\delta$,
\[
\sum_{k=1}^K\!\Big[
  Q^{\pi^\star,Y_k}_{P^\star,\,\bar r^{(k)}}(s_1,\cdot)
  -Q^{\pi^k,Y_k}_{P^\star,\,\bar r^{(k)}}(s_1,\cdot)
\Big]
\;\leq\;\frac{KH\log|\mathcal A|}{\alpha} + (1+\xi)\sum_{k=1}^K \epsilon_k,
\]
where the cumulative representation error satisfies
$\sum_{k=1}^K\epsilon_k\leq C\,H^2 d^2 K^{3/4}|\mathcal{A}|^{3/2}
\log^{1/4}(dK|\Phi|/\delta)$.
\end{lemma}

\begin{remark}[On the per-episode rate]\label{rem:perepisode}
The cumulative bound on $\sum_k \epsilon_k$ corresponds to an average per-episode
rate $\bar\epsilon_K=\frac{1}{K}\sum_k\epsilon_k=\widetilde O(K^{-1/4})$. We state
the cumulative form because the elliptical-potential argument in
Appendix~\ref{app:pseudoregret} directly yields a sum, and only this sum is needed
downstream.
\end{remark}

The proof of Lemma~\ref{lem:policygap} combines composite optimism
(Lemma~\ref{lem:optimism}), a softmax sub-optimality bound for the composite
planning problem, and the simulation lemma. The $(1+Y_k)$ factor arises because
both $\bar r^{(k)}_h\in[0,1+Y_k]$ and $\hat b^{(k)}_h=(1+Y_k)\hat b_h$ scale linearly
in $1+Y_k$; using $Y_k\leq\xi$, this factor is bounded by $1+\xi$, contributing the
$1/\gamma$ dependence.

Combining Lemma~\ref{lem:policygap} with the V--Q identity
$V_{j,1}^k(s_1)=\sum_a \pi^k_1(a\mid s_1) Q_{j,1}^k(s_1,a)$ and a performance-difference
argument on $P^\star$ (Appendix~\ref{app:pseudoregret}):

\begin{lemma}[Pseudo regret]\label{lem:pseudoregret}
Conditioned on the Representation Learning guarantee, with probability $1-p/4$,
\[
\sum_{k=1}^K\!\Big[
  \big(V_{r,1}^{\pi^\star}(s_1)+Y_k V_{g,1}^{\pi^\star}(s_1)\big)
  -\big(V_{r,1}^k(s_1)+Y_k V_{g,1}^k(s_1)\big)
\Big]
\;\leq\;\frac{KH\log|\mathcal{A}|}{\alpha}+(1+\xi)\sum_{k=1}^K\epsilon_k.
\]
\end{lemma}

\paragraph{Step 3: Bounding the Estimation error.}
The estimation error compares the LSVI-UCB optimistic estimate $V^k_{j,1}$ to the
true value $V^{\pi^k}_{j,1}$ of the executed policy. The starting point is a
self-normalized concentration bound for the empirical Bellman residual.

\begin{lemma}[Concentration of empirical Bellman residual]\label{lem:concentration}
Adapting the self-normalized concentration of \citet{ghosh2022provably} to the
learned-feature setting via a covering argument over the optimistic value-function
class, there exists a constant $C_2$ such that for any $p\in(0,1)$, the event:
\[
\mathcal{E}:=\bigcap_{k\in[K],\,h\in[H],\,j\in\{r,g\}}\!\Bigg\{\,
\Big\|\sum_{\tau=1}^{k-1}\hat\phi_h^\tau\!\big(V_{j,h+1}^k(s_{h+1}^\tau)
-\mathbb{P}_h V_{j,h+1}^k(s_h^\tau,\tilde a_h^\tau)\big)\Big\|_{(\Sigma_h^k)^{-1}}
\leq C_2\,dH\sqrt{\chi}\,
\Bigg\}
\]
holds with $\chi=\log\!\big(4(C_1+1)\log|\mathcal{A}|\,dT|\Phi|/p\big)$, and
$\Pr(\mathcal{E})\geq 1-p/4$.
\end{lemma}

\begin{remark}[From fixed to learned features]\label{rem:learned-feat}
\citet{ghosh2022provably} states this concentration for a fixed feature map. Here
both $\hat\phi_h^\tau$ and $V^k_{j,h+1}$ are data-dependent, so we cover the
optimistic value-function class (parameterized by $\hat\phi\in\Phi$ and bounded
LSVI weights), which contributes the $\log|\Phi|$ factor inside $\chi$. The
detailed argument is deferred to Appendix~\ref{app:estimation}.
\end{remark}

We then control the per-step gap between $Q^k_{j,h}$ (the LSVI-UCB optimistic
estimate) and $Q^{\pi^k}_{j,h}$ (the true action-value of the executed policy).
The key technical step is showing that both quantities admit linear approximations
in the learned features $\hat\phi^k$: $Q^k_{j,h}$ by construction, since the
LSVI-UCB output is explicitly linear in $\hat\phi^k$; and $Q^{\pi^k}_{j,h}$ via
the RepLearn guarantee, which ensures $\hat\phi^k$ approximately spans the
Bellman backups of any function in $\mathcal{V}_h$.

\begin{lemma}[Per-step backup error]\label{lem:backup}
Set $\beta=C_1\,dH\sqrt{\iota}$ with $\iota=\log(\log|\mathcal{A}|\cdot 4dT|\Phi|/p)$.
On the event $\mathcal{E}$, for every round $k\in[K]$, step $h\in[H]$,
$j\in\{r,g\}$, and $(s,a)\in\mathcal{S}\times\mathcal{A}$, there exist weights
$w_{j,h}^{\pi^k}$ approximately projecting $Q^{\pi^k}_{j,h}$ onto
$\mathrm{span}(\hat\phi^k_h)$ such that:
\[
Q_{j,h}^k(s,a)-Q_{j,h}^{\pi^k}(s,a)
=\mathbb{E}_{s'\sim P_h^\star(\cdot\mid s,a)}\!\left[V_{j,h+1}^k(s')-V_{j,h+1}^{\pi^k}(s')\right]+\Delta_h^k(s,a),
\]
where $|\Delta_h^k(s,a)|\leq\beta\sqrt{\hat\phi^k_h(s,a)^\top(\Sigma_h^k)^{-1}\hat\phi^k_h(s,a)}$
plus a RepLearn projection residual that accumulates into $\sum_k\epsilon_k$.
\end{lemma}

Unrolling Lemma~\ref{lem:backup} via $V^k_{j,h}(s)=\sum_a \pi^k_h(a\mid s)Q^k_{j,h}(s,a)$,
summing telescopically over $h$, and applying the elliptical potential lemma over
$k$, we obtain:

\begin{lemma}[Estimation error]\label{lem:estimation}
With probability at least $1-p/2$,
\[
\sum_{k=1}^K\!\big(V_{r,1}^k(s_1)-V_{r,1}^{\pi^k}(s_1)\big)
+Y\sum_{k=1}^K\!\big(V_{g,1}^k(s_1)-V_{g,1}^{\pi^k}(s_1)\big)
\;\leq\; O\!\Big((Y+1)\sqrt{d^3 H^3 T\iota^2}\Big)+(1+\xi)\sum_{k=1}^K\epsilon_k,
\]
where $T=KH$.
\end{lemma}

\paragraph{Step 4: Combining Pseudo regret and Estimation error.}
Plugging Lemmas~\ref{lem:pseudoregret} and~\ref{lem:estimation}
into~\eqref{eq:decomp} with $\eta=\xi/\sqrt{KH^2}$ and
$\alpha=\frac{K\log|\mathcal{A}|}{2(1+\xi+H)}$:
\begin{align*}
&\sum_{k=1}^K\!\big(V^{\pi^\star}_{r,1}(s_1)-V^{\pi^k}_{r,1}(s_1)\big)
+Y\sum_{k=1}^K\!\big(b-V^{\pi^k}_{g,1}(s_1)\big)\leq\;\frac{Y^2}{2\eta}+\frac{\eta H^2 K}{2}+\frac{KH\log|\mathcal{A}|}{\alpha}\\
&+2(1+\xi)\sum_{k=1}^K\epsilon_k+O\!\Big((Y+1)\sqrt{d^3 H^3 T\iota^2}\Big).
\end{align*}
For these parameter choices, $\eta H^2 K/2$, $KH\log|\mathcal{A}|/\alpha$, and the
estimation-error term are all dominated by $2(1+\xi)\sum_k\epsilon_k$, which evaluates
to $\widetilde O\!\left(\frac{H^3 d^2 K^{3/4}|\mathcal{A}|^{3/2}}{\gamma}\right)$. Setting $Y=0$ yields
the regret bound. Setting $Y=\xi$ and applying Lemma~\ref{lem:aux-strongdual}
(strong-duality conversion of \citealt{ding2021provably}) with $\xi\geq 2Y^\star$
converts this into the cumulative violation bound, matching
Theorem~\ref{thm:mfrlc}.

\section{Experiments}
\label{sec:experiments}

\begin{figure}[t]
  \centering
  \includegraphics[width=0.62\columnwidth]{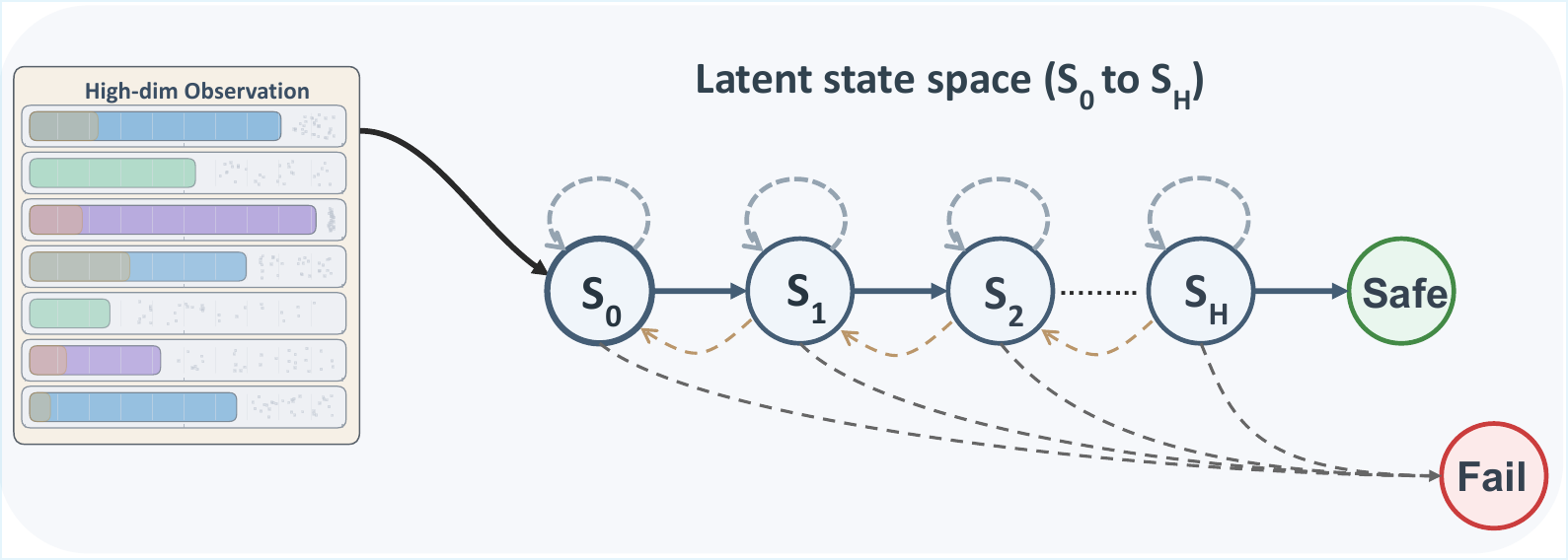}
  \caption{Constrained Latent ComboLock}
  \label{fig:env}
\end{figure}

We introduce the \textbf{Constrained Latent ComboLock (CLCL)}, a new 
benchmark we designed to test agents on difficult, long-horizon visual 
safety problems (Figure~\ref{fig:env}). Our experiments on CLCL 
validate the strengths of our proposed method \texttt{MFRLC}.

\begin{figure}[t]
  \centering
  \includegraphics[width=0.92\columnwidth]{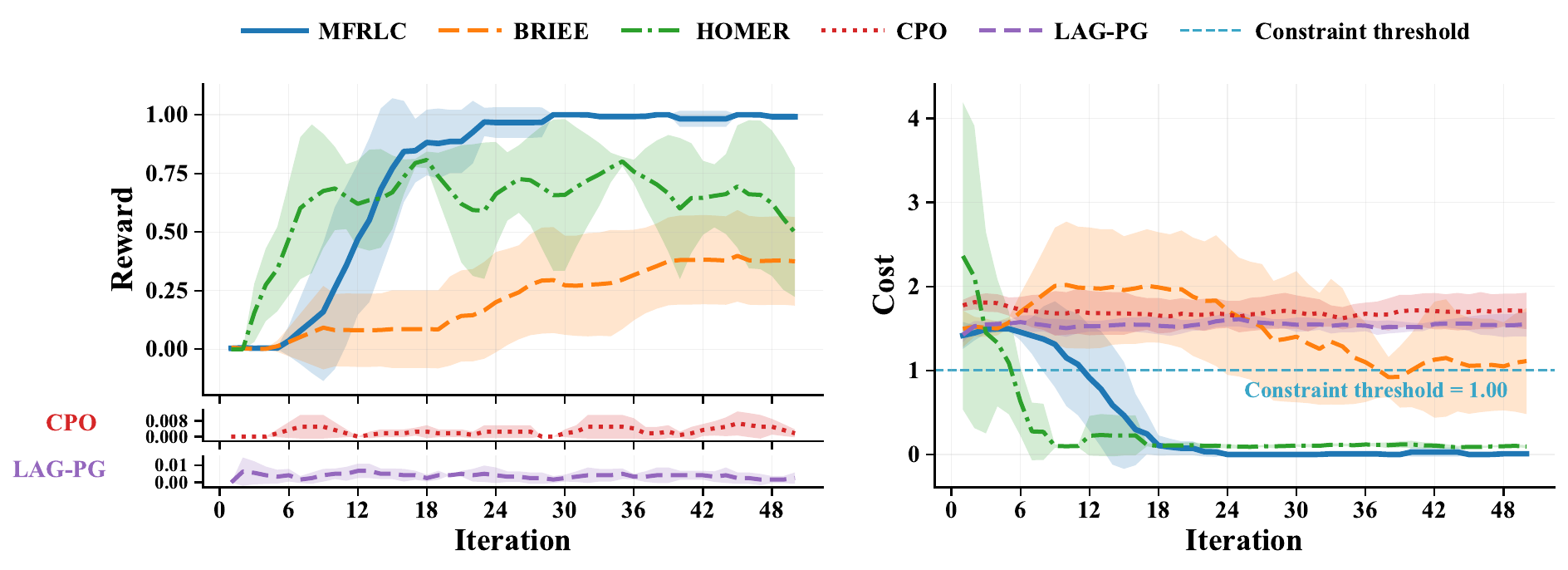}
  \caption{Reward and Cost vs.\ iteration on CLCL (H=6)}
  \label{fig:reward_cost_6}
\end{figure}

\paragraph{Environment Setup}
Experiments use the \textbf{CLCL} (Fig.~\ref{fig:env}), a CMDP with horizon $H = 6$, high-dimensional observations, and a latent state space. The latent states form a chain $s_0 \to s_1 \to \cdots \to s_{H}$ with two absorbing terminals: a $S_{\mathrm{Safe}}$ state and a $S_{\mathrm{Fail}}$ state. The agent must learn a sequence of actions (the ``combo'') to reach the Safe state, but never observes the latent state $s_h$; it receives only an observation $o_h \in \mathbb{R}^{8\times84\times84}$ from a rendering function, mapped for learning into a feature class $\Phi$ of dimension $d=64$. Observations visually encode progress and timestep, with added noise to increase perceptual difficulty. At each state $s_h$, the agent chooses from the action set $\mathcal{A} = \{a_{\mathrm{forward}}, a_{\mathrm{hazard}}, a_{\mathrm{stay}}\}$. Taking $a_{\mathrm{forward}}$ transitions the agent from state $s_h$ to $s_{h+1}$. Taking $a_{\mathrm{stay}}$ causes the agent to remain in $s_h$ without cost or reward. Taking $a_{\mathrm{hazard}}$ incurs a cost of $1$ and causes transition to $S_{\mathrm{Fail}}$ with probability $p_{\mathrm{hazard}}$, or return to $s_{h-1}$ with probability $1-p_{\mathrm{hazard}}$. A reward of $1$ is given only upon successfully transitioning from $s_{H-1}$ to $S_{\mathrm{Safe}}$; all other rewards are zero. The objective is to maximize expected reward while ensuring the total expected cost remains below $1$. Full setup details and hyperparameters are in Appendix~\ref{app:experiments}.

\paragraph{Baselines}
We compare our proposed method against a suite of reinforcement learning algorithms. Our baselines include \textsc{Lagrangian Policy Gradient (LAG-PG)} \citep{Ray2019SafetyGym}, a standard on-policy actor-critic with GAE for reward and cost; \textsc{Constrained Policy Optimization (CPO)} \citep{Achiam2017CPO}, an on-policy trust-region method for CMDPs; \textsc{HOMER} \citep{misra2020kinematic}, a hierarchical option-based algorithm; and \textsc{Block-structured Representation learning with Interleaved Explore Exploit (BRIEE)} \citep{zhang2022efficient}, which focuses on representation learning coupled with a policy optimization. Since \textsc{HOMER} and \textsc{BRIEE} were originally designed for unconstrained MDPs, we add a Lagrangian framework for fair comparison; details are in Appendix~\ref{app:baselines}.

\paragraph{Results}
\texttt{MFRLC} achieves the best overall performance, balancing reward maximization with constraint satisfaction. In the CLCL task with horizon $H=6$ (Fig.~\ref{fig:reward_cost_6}), \texttt{MFRLC} is the only algorithm to achieve an optimal reward approaching $1.0$ while satisfying the safety constraint. \textsc{HOMER} satisfies the constraint sooner but converges to a suboptimal reward of $\approx 0.8$; \textsc{BRIEE} plateaus at $\approx 0.4$ and violates the constraint; \textsc{CPO} and \textsc{LAG-PG} fail to achieve meaningful reward and consistently violate safety. Representation learning alone is thus insufficient: only \texttt{MFRLC}, by jointly learning features and a constraint-aware policy, succeeds on both objectives (details in Appendix~\ref{app:experiments}).

\paragraph{Additional Settings}
Table~\ref{tab:extended} adds a larger action space and a longer horizon (Appendix~\ref{app:extended}). With $|\mathcal{A}|=5$, where discovering the rewarding sequence under uniform exploration is $(5/3)^6\approx 21.4$ times less likely, \texttt{MFRLC} attains the highest reward and lowest violation. At $H=8$, \texttt{MFRLC} obtains the highest reward; only \textsc{CPO} has lower violation, at near-zero reward. Its estimated cumulative regret is $15{,}382\pm942$ over $15{,}552$ deployed episodes; we do not read these finite-run numbers as verifying the $K^{3/4}$ rate. At $H=12$, \texttt{MFRLC} reaches only $0.021$ reward, a failure mode under our budget.

\begin{table}[t]
\centering
\small
\caption{Additional CLCL settings (mean $\pm$ s.e., 5 seeds). Best $R$: maximum reward per run; Viol.: cumulative violation. Compare within each setting.}
\label{tab:extended}
\begin{tabular}{llcc}
\toprule
Setting & Method & Best $R$ & Viol. \\
\midrule
\multirow{3}{*}{$H=6$, $|\mathcal{A}|=5$} & \texttt{MFRLC} & $\mathbf{0.567}\pm0.174$ & $\mathbf{711.9}\pm30.8$ \\
 & \textsc{BRIEE} & $0.105\pm0.099$ & $1063.1\pm26.3$ \\
 & \textsc{LAG-PG} & $0.008\pm0.005$ & $1614.7\pm137.7$ \\
\midrule
\multirow{5}{*}{$H=8$, $|\mathcal{A}|=3$} & \texttt{MFRLC} & $\mathbf{0.692}\pm0.136$ & $207.4\pm13.0$ \\
 & \textsc{BRIEE} & $0.505\pm0.005$ & $683.4\pm2.7$ \\
 & \textsc{LAG-PG} & $0.017\pm0.004$ & $747.4\pm191.5$ \\
 & \textsc{HOMER} & $0.004\pm0.004$ & $936.5\pm31.0$ \\
 & \textsc{CPO} & $0.012\pm0.005$ & $\mathbf{102.2}\pm75.8$ \\
\midrule
$H=12$, $|\mathcal{A}|=3$ & \texttt{MFRLC} & $0.021\pm0.013$ & $1206.9\pm76.2$ \\
\bottomrule
\end{tabular}
\end{table}

\section{Limitations}
\label{sec:limitations}
Our results have four main limitations. (i) Theorem~\ref{thm:mfrlc} is an upper bound; we prove no lower bound and make no optimality claim about the $K^{3/4}$ rate or the dependence on $d$, $H$, $|\mathcal{A}|$, and $\gamma$. (ii) The outer minimax problem of Algorithm~\ref{alg:replearn} is solved by an optimization oracle; exact search is possible for finite classes, but we give no polynomial-time guarantee for neural classes. (iii) Regret and Violation are charged only to deployed policies; the auxiliary roll-ins of representation learning are counted in the $K(2H+1)$ trajectory complexity but not in the regret or violation sums. (iv) Our experiments use a contrastive EMA surrogate that is not covered by Theorem~\ref{thm:mfrlc}, the evaluation is restricted to CLCL, and \texttt{MFRLC} fails to obtain meaningful reward at $H=12$ under our training budget. The analysis also assumes a finite action space, which incurs the $|\mathcal{A}|^{3/2}$ factor through uniform exploration, exact realizability $\phi^\star\in\Phi$, and reward and utility linear in $\phi^\star$.

\section{Conclusion}
We presented \texttt{MFRLC}, a model-free algorithm for online low-rank
CMDPs with unknown features. By combining LSVI-UCB, a primal--dual scheme,
and a minimax representation-learning objective, \texttt{MFRLC} jointly
learns features and constraint-aware policies and achieves
$\widetilde{O}(K^{3/4})$ reward regret and cumulative constraint violation
over $K$ deployment rounds. Improving this rate, and certifying a scalable
representation learner against the guarantee of
Algorithm~\ref{alg:replearn}, remain open.

\begin{ack}
\textbf{Funding.} ST is supported by the Emory URC Research Award and the Disruptive Discovery Seed Program by Emory University’s Office of the Provost. The content in this document are those of the authors and should not be interpreted as necessarily representing the views of the funding sources. \textbf{Competing interests.} The authors declare no competing interests.
\end{ack}

\appendix

\clearpage
\bibliographystyle{plainnat}
\bibliography{ref}

@article{ono2015chance,
  title={Chance-constrained dynamic programming with application to risk-aware robotic space exploration},
  author={Ono, Masahiro and Pavone, Marco and Kuwata, Yoshiaki and Balaram, J},
  journal={Autonomous Robots},
  volume={39},
  pages={555--571},
  year={2015},
  publisher={Springer}
}

@article{yu2021reinforcement,
  title={Reinforcement learning in healthcare: A survey},
  author={Yu, Chao and Liu, Jiming and Nemati, Shamim and Yin, Guosheng},
  journal={ACM Computing Surveys (CSUR)},
  volume={55},
  number={1},
  pages={1--36},
  year={2021},
  publisher={ACM New York, NY}
}

@article{efroni2020exploration,
  title={Exploration-exploitation in constrained {MDPs}},
  author={Efroni, Yonathan and Mannor, Shie and Pirotta, Matteo},
  journal={arXiv preprint arXiv:2003.02189},
  year={2020}
}

@inproceedings{ding2021provably,
  title={Provably efficient safe exploration via primal-dual policy optimization},
  author={Ding, Dongsheng and Wei, Xiaohan and Yang, Zhuoran and Wang, Zhaoran and Jovanovic, Mihailo},
  booktitle={International Conference on Artificial Intelligence and Statistics},
  pages={3304--3312},
  year={2021},
  organization={PMLR}
}

@article{ghosh2022provably,
  title={Provably efficient model-free constrained {RL} with linear function approximation},
  author={Ghosh, Arnob and Zhou, Xingyu and Shroff, Ness},
  journal={Advances in Neural Information Processing Systems},
  volume={35},
  pages={13303--13315},
  year={2022}
}

@inproceedings{zhang2022efficient,
  title={Efficient reinforcement learning in block {MDPs}: A model-free representation learning approach},
  author={Zhang, Xuezhou and Song, Yuda and Uehara, Masatoshi and Wang, Mengdi and Agarwal, Alekh and Sun, Wen},
  booktitle={International Conference on Machine Learning},
  pages={26517--26547},
  year={2022},
  organization={PMLR}
}

@article{agarwal2020flambe,
  title={{FLAMBE}: Structural complexity and representation learning of low rank {MDPs}},
  author={Agarwal, Alekh and Kakade, Sham and Krishnamurthy, Akshay and Sun, Wen},
  journal={Advances in neural information processing systems},
  volume={33},
  pages={20095--20107},
  year={2020}
}

@article{uehara2021representation,
  title={Representation learning for online and offline {RL} in low-rank {MDPs}},
  author={Uehara, Masatoshi and Zhang, Xuezhou and Sun, Wen},
  journal={arXiv preprint arXiv:2110.04652},
  year={2021}
}

@inproceedings{ren2022free,
  title={A free lunch from the noise: Provable and practical exploration for representation learning},
  author={Ren, Tongzheng and Zhang, Tianjun and Szepesv{\'a}ri, Csaba and Dai, Bo},
  booktitle={Uncertainty in Artificial Intelligence},
  pages={1686--1696},
  year={2022},
  organization={PMLR}
}

@article{modi2024model,
  title={Model-free representation learning and exploration in low-rank {MDPs}},
  author={Modi, Aditya and Chen, Jinglin and Krishnamurthy, Akshay and Jiang, Nan and Agarwal, Alekh},
  journal={Journal of Machine Learning Research},
  volume={25},
  number={6},
  pages={1--76},
  year={2024}
}

@inproceedings{xu2021crpo,
  title={{CRPO}: A new approach for safe reinforcement learning with convergence guarantee},
  author={Xu, Tengyu and Liang, Yingbin and Lan, Guanghui},
  booktitle={International Conference on Machine Learning},
  pages={11480--11491},
  year={2021},
  organization={PMLR}
}

@article{ding2020natural,
  title={Natural policy gradient primal-dual method for constrained {M}arkov decision processes},
  author={Ding, Dongsheng and Zhang, Kaiqing and Basar, Tamer and Jovanovic, Mihailo},
  journal={Advances in Neural Information Processing Systems},
  volume={33},
  pages={8378--8390},
  year={2020}
}

@inproceedings{ding2023provably,
  title={Provably efficient primal-dual reinforcement learning for {CMDPs} with non-stationary objectives and constraints},
  author={Ding, Yuhao and Lavaei, Javad},
  booktitle={Proceedings of the AAAI Conference on Artificial Intelligence},
  volume={37},
  number={6},
  pages={7396--7404},
  year={2023}
}

@article{wei2021provably,
  title={A provably-efficient model-free algorithm for constrained {M}arkov decision processes},
  author={Wei, Honghao and Liu, Xin and Ying, Lei},
  journal={arXiv preprint arXiv:2106.01577},
  year={2021}
}

@inproceedings{jin2020provably,
  title={Provably efficient reinforcement learning with linear function approximation},
  author={Jin, Chi and Yang, Zhuoran and Wang, Zhaoran and Jordan, Michael I},
  booktitle={Conference on learning theory},
  pages={2137--2143},
  year={2020},
  organization={PMLR}
}

@article{paternain2019constrained,
  title={Constrained reinforcement learning has zero duality gap},
  author={Paternain, Santiago and Chamon, Luiz and Calvo-Fullana, Miguel and Ribeiro, Alejandro},
  journal={Advances in Neural Information Processing Systems},
  volume={32},
  year={2019}
}

@inproceedings{melo2007q,
  title={{Q}-learning with linear function approximation},
  author={Melo, Francisco S and Ribeiro, M Isabel},
  booktitle={International Conference on Computational Learning Theory},
  pages={308--322},
  year={2007},
  organization={Springer}
}

@inproceedings{azizzadenesheli2018efficient,
  title={Efficient exploration through {B}ayesian deep {Q}-networks},
  author={Azizzadenesheli, Kamyar and Brunskill, Emma and Anandkumar, Animashree},
  booktitle={2018 Information Theory and Applications Workshop (ITA)},
  pages={1--9},
  year={2018},
  organization={IEEE}
}

@inproceedings{amani2021safe,
  title={Safe reinforcement learning with linear function approximation},
  author={Amani, Sanae and Thrampoulidis, Christos and Yang, Lin},
  booktitle={International Conference on Machine Learning},
  pages={243--253},
  year={2021},
  organization={PMLR}
}

@article{nedic2009subgradient,
  title={Subgradient methods for saddle-point problems},
  author={Nedi{\'c}, Angelia and Ozdaglar, Asuman},
  journal={Journal of optimization theory and applications},
  volume={142},
  pages={205--228},
  year={2009},
  publisher={Springer}
}

@inproceedings{misra2020kinematic,
  title={Kinematic state abstraction and provably efficient rich-observation reinforcement learning},
  author={Misra, Dipendra and Henaff, Mikael and Krishnamurthy, Akshay and Langford, John},
  booktitle={International conference on machine learning},
  pages={6961--6971},
  year={2020},
  organization={PMLR}
}

@inproceedings{wei2023provably,
  title={Provably efficient model-free algorithms for non-stationary {CMDPs}},
  author={Wei, Honghao and Ghosh, Arnob and Shroff, Ness and Ying, Lei and Zhou, Xingyu},
  booktitle={International Conference on Artificial Intelligence and Statistics},
  pages={6527--6570},
  year={2023},
  organization={PMLR}
}

@inproceedings{
ghosh2023achieving,
title={Achieving Sub-linear Regret in Infinite Horizon Average Reward Constrained {MDP} with Linear Function Approximation},
author={Arnob Ghosh and Xingyu Zhou and Ness Shroff},
booktitle={The Eleventh International Conference on Learning Representations },
year={2023},
url={https://openreview.net/forum?id=zZhX4eYNeeh}
}

@inproceedings{Achiam2017CPO,
  title={Constrained Policy Optimization},
  author={Joshua Achiam and David Held and Aviv Tamar and Pieter Abbeel},
  booktitle={ICML},
  year={2017},
  url={https://proceedings.mlr.press/v70/achiam17a/achiam17a.pdf}
}

@misc{Ray2019SafetyGym,
  title={Benchmarking Safe Exploration in Deep Reinforcement Learning},
  author={Alex Ray and Joshua Achiam and Dario Amodei},
  year={2019},
  eprint={1910.01708},
  archivePrefix={arXiv},
  url={https://cdn.openai.com/safexp-short.pdf}
}

@article{abbasi2011improved,
  title={Improved algorithms for linear stochastic bandits},
  author={Abbasi-Yadkori, Yasin and P{\'a}l, D{\'a}vid and Szepesv{\'a}ri, Csaba},
  journal={Advances in neural information processing systems},
  volume={24},
  year={2011}
}

@inproceedings{dani2008stochastic,
  title={Stochastic linear optimization under bandit feedback},
  author={Dani, Varsha and Hayes, Thomas P and Kakade, Sham M},
  booktitle={21st Annual Conference on Learning Theory},
  number={101},
  pages={355--366},
  year={2008}
}

@article{mhammedi2023efficient,
  title={Efficient model-free exploration in low-rank {MDPs}},
  author={Mhammedi, Zak and Block, Adam and Foster, Dylan J and Rakhlin, Alexander},
  journal={Advances in Neural Information Processing Systems},
  volume={36},
  pages={66782--66817},
  year={2023}
}

@book{puterman1994markov,
  title={Markov Decision Processes: Discrete Stochastic Dynamic Programming},
  author={Puterman, Martin L.},
  year={1994},
  publisher={Wiley},
  address={New York}
}

\clearpage

\vspace{1cm}
\appendix
\crefalias{section}{appendix}
\crefalias{subsection}{appendix}
\begin{center}
    {\LARGE \textbf{Supplementary Materials}}
\end{center}

In this supplementary material, we provide additional details omitted from the main paper for brevity. Throughout the appendix, ``episode $k$'' refers to deployment round $k$ (Section~\ref{sec:formulation}).

\section{LSVI-UCB Algorithm Details}
\label{app:lsvi}

\subsection{LSVI-UCB Algorithm}

Algorithm~\ref{alg:lsvi} provides the full pseudocode for our constrained
Least-Squares Value Iteration with Upper Confidence Bounds (LSVI-UCB) procedure.
This algorithm is invoked in the policy-update step of Algorithm~\ref{alg:mfrlc}
(line~8) at each episode $k$, taking as inputs (i) the trajectory data collected up
to episode $k$ via the exploration step (line~4 of Algorithm~\ref{alg:mfrlc}, with
uniform actions $\tilde a_h^\tau$), (ii) the learned representation $\hat\phi^k$, 
and (iii) the current dual variable $Y_k$. It returns the optimistic per-objective
$Q$- and $V$-functions $\{Q^k_{r,h},Q^k_{g,h},V^k_{r,h},V^k_{g,h}\}_{h=1}^H$
together with the policy $\pi^k = \{\pi^k_h\}_{h=1}^H$.

The algorithm proceeds via backward induction from the final horizon step $H$ down
to $1$. At each step $h$, it first constructs the regularized Gram matrix
$\Sigma^k_h$ (line~4), which represents the empirical covariance of the learned
features over past episodes. This matrix is then used to perform a regularized
least-squares regression to find the linear weights for the reward value function
($w^k_{r,h}$) and the utility value function ($w^k_{g,h}$) (lines~5--6). An optimistic
UCB exploration bonus $\hat b^k_h$ is then computed to encourage exploration in
less-visited state--action pairs (line~7). This single bonus is added to both the
reward and utility linear value estimates to produce the optimistic $Q$-values
(lines~8--9). The policy $\pi^k_h$ is derived from a softmax distribution over the
\emph{composite} $Q$-function $Q^k_{r,h}+Y_k Q^k_{g,h}$ (line~10). Finally, the
per-objective state-value functions are updated under this policy (lines~11--12)
and used at the next iteration of the backward induction.

\begin{remark}[Single bonus, composite bonus]\label{rem:bonus-app}
Lines~8--9 add the same per-objective bonus $\hat b^k_h$ to both $Q^k_{r,h}$ and
$Q^k_{g,h}$. Consequently, the composite $Q$-function $Q^k_{r,h}+Y_k Q^k_{g,h}$
used in line~10 inherits a composite bonus of magnitude $(1+Y_k)\hat b^k_h$. The
analysis (Appendices~\ref{app:pseudoregret}--\ref{app:violation}) denotes this
composite bonus by $\hat b^{(k)}_h := (1+Y_k)\hat b^k_h$.
\end{remark}

\begin{algorithm}[h]
\caption{Least-Squares Value Iteration with UCB (LSVI-UCB)}
\label{alg:lsvi}
\begin{algorithmic}[1]
   \Require Trajectory data $\{(s_h^\tau, \tilde a_h^\tau, r_h(s_h^\tau, \tilde a_h^\tau), g_h(s_h^\tau, \tilde a_h^\tau), s_{h+1}^\tau, \hat{\phi}^k_h(s_h^\tau, \tilde a_h^\tau))\}_{h=1,\,\tau=1}^{H,\,k-1}$ from line~4 of Algorithm~\ref{alg:mfrlc}; current dual variable $Y_k$
   \State \textbf{Parameter:} $\alpha = \frac{K\log|\mathcal{A}|}{2(1+\xi+H)}$,\quad $\beta = C_1 dH\sqrt{\log(4\log|\mathcal{A}|\,dT|\Phi|/p)}$,\quad regularization $\lambda$
   \Ensure Optimistic estimates $\{Q^k_{r,h}, Q^k_{g,h}, V^k_{r,h}, V^k_{g,h}\}_{h=1}^H$ and policy $\pi^k = \{\pi^k_h\}_{h=1}^H$
   \State Initialize $V^k_{r,H+1}(\cdot) \equiv 0$,\quad $V^k_{g,H+1}(\cdot) \equiv 0$
   \For{$h = H$ \textbf{to} $1$}
      \State $\Sigma^k_{h} \leftarrow \displaystyle\sum_{\tau=1}^{k-1} \hat{\phi}^k_h(s_h^\tau, \tilde a_h^\tau)\,\hat{\phi}^k_h(s_h^\tau, \tilde a_h^\tau)^{\top} + \lambda I$
      \State $w^k_{r,h} \leftarrow (\Sigma^k_h)^{-1} \displaystyle\sum_{\tau=1}^{k-1} \hat{\phi}^k_h(s_h^\tau, \tilde a_h^\tau)\,\big(r_h(s_h^\tau, \tilde a_h^\tau) + V^k_{r,h+1}(s_{h+1}^\tau)\big)$
      \State $w^k_{g,h} \leftarrow (\Sigma^k_h)^{-1} \displaystyle\sum_{\tau=1}^{k-1} \hat{\phi}^k_h(s_h^\tau, \tilde a_h^\tau)\,\big(g_h(s_h^\tau, \tilde a_h^\tau) + V^k_{g,h+1}(s_{h+1}^\tau)\big)$
      \State $\hat{b}^k_h(s, a) \leftarrow \beta\sqrt{\hat{\phi}^k_h(s,a)^{\top}(\Sigma^k_h)^{-1}\hat{\phi}^k_h(s,a)}$ \quad \textit{// per-objective bonus}
      \State $Q^k_{r,h}(s, a) \leftarrow \min\!\Big(\langle w^k_{r,h},\, \hat{\phi}^k_h(s, a)\rangle + \hat{b}^k_h(s, a),\; H\Big)$
      \State $Q^k_{g,h}(s, a) \leftarrow \min\!\Big(\langle w^k_{g,h},\, \hat{\phi}^k_h(s, a)\rangle + \hat{b}^k_h(s, a),\; H\Big)$
      \State $\pi^k_h(a \mid \cdot) \leftarrow \mathrm{softmax}_\alpha\!\Big(Q^k_{r,h}(\cdot, a) + Y_k\,Q^k_{g,h}(\cdot, a)\Big)$
      \State $V^k_{r,h}(s) \leftarrow \displaystyle\sum_{a\in\mathcal{A}} \pi^k_h(a \mid s)\, Q^k_{r,h}(s, a)$
      \State $V^k_{g,h}(s) \leftarrow \displaystyle\sum_{a\in\mathcal{A}} \pi^k_h(a \mid s)\, Q^k_{g,h}(s, a)$
   \EndFor
\end{algorithmic}
\end{algorithm}

\subsection{Comparator}\label{app:comparator}
Theorem~\ref{thm:mfrlc} compares against an arbitrary optimal feasible stochastic policy $\pi^\star$ of~\eqref{eq:obj}. The policy $\pi^\star$ enters the analysis only through the pseudo-regret term of~\eqref{eq:decomp}, which Lemma~\ref{lem:pseudoregret} bounds by combining composite optimism (Lemma~\ref{lem:optimism}), the softmax sub-optimality bound (Lemma~\ref{lem:proof-softmax}), and the simulation lemma (Lemma~\ref{lem:aux-simulation}). None of these steps requires $\pi^\star$ to be produced by Algorithm~\ref{alg:lsvi} from any representation and dual variable. Lemma~\ref{lem:optimism} bounds the optimistic value of the fixed policy $\pi^\star$ on the planning model, and Lemma~\ref{lem:proof-softmax} holds for an arbitrary comparator policy. The finite-temperature softmax error is already accounted for by the explicit $KH\log|\mathcal{A}|/\alpha$ term, so exact recovery of $\pi^\star$ is unnecessary.

For measurability in the concentration arguments, note that Algorithm~\ref{alg:lsvi} uses the data only through the feature evaluations $\hat\phi^k_h(s^\tau_h,\tilde a^\tau_h)$, the regression targets, and $Y_k$. Hence $\pi^k$ is a deterministic function of $\hat\phi^k$, the data collected before round $k$, $Y_k$, and fixed hyperparameters.

\section{Related Work}

\paragraph{Representation Learning in Low-Rank MDPs.}
Low-rank MDPs \citep{agarwal2020flambe} provide a structural framework where the transition kernel admits a $d$-dimensional factorization, generalizing earlier rich-observation models \citep{misra2020kinematic}. Existing approaches fall into two categories. \emph{Model-based} methods explicitly estimate both the feature map and the next-state factor \citep{uehara2021representation,ren2022free}, requiring realizability assumptions on the full dynamics. \emph{Model-free} methods avoid recovering the next-state factor and instead learn the feature via a min-max objective against a discriminator class: \citet{zhang2022efficient} (BRIEE) operates in block MDPs, \citet{modi2024model} extends to general low-rank MDPs, and \citet{mhammedi2023efficient} (VoX) achieves improved sample complexity via barycentric spanners. \emph{None of these works handles constraints}, leaving the constrained setting open.

\paragraph{Constrained MDPs.}
CMDPs \citep{paternain2019constrained} provide a standard framework for safe RL, where the agent maximizes reward subject to expected cumulative cost constraints. For tabular settings, primal-dual and primal-only methods \citep{efroni2020exploration,ding2020natural,xu2021crpo} establish regret and violation guarantees, with model-free variants \citep{wei2021provably,ding2023provably,wei2023provably} achieving $\widetilde O(\sqrt K)$ rates. To scale beyond tabular settings, \citet{Achiam2017CPO} introduce constrained policy optimization for deep RL, \citet{amani2021safe} treats safe linear bandits, and \citet{ding2021provably,ghosh2022provably,ghosh2023achieving} extend primal-dual policy optimization to linear CMDPs with $\widetilde O(\sqrt K)$ guarantees. \emph{All these works assume the feature map is known a priori}---an assumption that fails when only raw observations (e.g., images) are available. Bridging this gap, \texttt{MFRLC} is, to our knowledge, the first algorithm with provable regret and violation guarantees for CMDPs with unknown features. The unknown-feature setting requires controlling three coupled error-propagation mechanisms absent from prior analyses. First, errors in the estimated utility values affect the dual iterate $Y_k$. Second, the changing $Y_k$ modifies the composite objective $r+Y_kg$, so representation error propagates through both the reward and utility Bellman estimates and is amplified by $1+Y_k$. Third, the resulting policy changes the data used for subsequent representation learning, which feeds back into the value estimates and the dual update. Prior low-rank MDP analyses do not handle constraints, and prior primal--dual CMDP analyses assume known features, so neither faces this coupling.
\section{Representation Learning Algorithm}
\label{app:replearn}

This appendix collects the implementation details of the representation learning step
invoked in line~6 of Algorithm~\ref{alg:mfrlc}. We first specify the discriminator
class $\mathcal{V}_h = \mathcal{F}_h\cup\mathcal{G}_h$, then derive the empirical
surrogate of the population objective~\eqref{eq:repobj}, and finally state the full
iterative procedure (Algorithm~\ref{alg:replearn}) together with its theoretical
guarantee.

\subsection{Discriminator Class}

The two discriminator classes used in Section~\ref{subsec:replearn} are defined as
\begin{equation}\label{eq:Fh}
\mathcal{F}_h := \Big\{ s\mapsto\mathbb{E}_{a \sim \pi^{k-1}_{h}(\cdot|s)}\!\!
\left[ \langle \phi(s, a), \theta \rangle - \langle \phi'(s, a), \theta' \rangle \right] :
\phi, \phi' \in \Phi_{h+1},\ \max(\|\theta\|_\infty, \|\theta'\|_\infty) \leq 1 \Big\},
\end{equation}
\begin{equation}\label{eq:Gh}
\begin{aligned}
\mathcal{G}_h := \bigg\{\, s\mapsto \max_{a\in\mathcal A}\bigg(&
\tfrac{r_{h+1}(s,a)+g_{h+1}(s,a)+\min\{\langle\phi(s,a),w\rangle,\,2\}}{2H+1}
+ \langle\phi(s,a),w'\rangle\bigg) :\\
&\phi \in \Phi_{h+1},\ \|w\|_\infty \leq c,\ \|w'\|_\infty \leq 1 \,\bigg\}.
\end{aligned}
\end{equation}
$\mathcal{F}_h$ captures the model-mismatch function appearing in the proof of
composite optimism (Lemma~\ref{lem:optimism}), evaluated under the previous-episode policy $\pi^{k-1}$ which is well-defined at the start of episode $k$. $\mathcal{G}_h$ captures (after
rescaling by $1+Y_k$) the linear part of the optimistic composite $Q$-function appearing in the
policy-gap bound (Lemma~\ref{lem:policygap}); the non-linear softmax operator and UCB bonus produced by Algorithm~\ref{alg:lsvi} are not contained in $\mathcal{G}_h$ and are handled separately via Lemma~\ref{lem:proof-softmax} and the elliptical-potential argument in Appendix~\ref{app:pseudoregret}. The simple form
above---independent of $Y_k$---defines a discriminator class with statistical
complexity $\log|\mathcal G_h|$ that does not depend on $Y_k$. When applied to
composite quantities of magnitude $1+Y_k$ in the proofs of
Lemmas~\ref{lem:optimism}--\ref{lem:policygap}, the rescaling factor $1+Y_k\le 1+\xi$
appears as an explicit multiplicative constant in the RepLearn statistical error
$\zeta_k$, which is absorbed into the cumulative bound $(1+\xi)\sum_k\epsilon_k$
in Lemma~\ref{lem:policygap}.

\subsection{From Population to Empirical Objective}

Recall the population objective~\eqref{eq:repobj} from the main text:
\[
\arg\min_{\phi\in\Phi_h}\;\max_{v\in\mathcal V_h}\;\min_{w\in\mathbb{R}^d}\;
\mathbb{E}_{(s,\tilde a)\sim\rho_h^k}\!\Big[\big(\langle\phi(s,\tilde a),w\rangle
-\mathbb{E}_{s'\sim P^\star_h(\cdot\mid s,\tilde a)}\![v(s')]\big)^2\Big].
\]
Direct optimization is intractable because the inner expectation
$\mathbb{E}_{s'\sim P^\star_h}[v(s')]$ requires the unknown true kernel. To address
this, we replace the squared deviation with the squared empirical residual:
\begin{equation}\label{eq:Lemp}
\mathcal{L}_{\lambda_k,\mathcal{D}_h}(\phi,v,w) \;:=\;
\mathbb{E}_{(s,\tilde a,s')\sim\mathcal{D}_h}\!\Big[\big(\langle\phi(s,\tilde a),w\rangle - v(s')\big)^2\Big]
\;+\;\lambda_k\|w\|_2^2.
\end{equation}
Expanding around the conditional mean of $v(s')$ gives the variance--bias
decomposition
\begin{equation}\label{eq:variance-decomp}
\begin{aligned}
\mathcal{L}_{\lambda_k,\mathcal{D}_h}(\phi,v,w)
\;=\;&\underbrace{\mathbb{E}_{(s,\tilde a)\sim\rho_h^k}\!\Big[
   \big(\langle\phi(s,\tilde a),w\rangle-\mathbb{E}_{s'\sim P^\star_h}[v(s')]\big)^2\Big]}_{\text{population objective in~\eqref{eq:repobj}}}\\
&+\underbrace{\mathbb{E}_{(s,\tilde a)\sim\rho_h^k}\!\big[\mathbb{V}_{s'\sim P^\star_h}[v(s')]\big]}_{\text{conditional variance term}}
\;+\;\lambda_k\|w\|_2^2.
\end{aligned}
\end{equation}
The conditional-variance term is independent of $\phi$ but is non-trivially correlated
with $v$, so na\"ively minimizing $\mathcal{L}_{\lambda_k,\mathcal{D}_h}$ alone biases
the inner minimization in favor of low-variance discriminators rather than the
true population minimum.

To cancel the variance term, we exploit the fact that the second buffer
$\mathcal{D}_h'$ contains transitions $(s',\tilde a',s'')$ where, for each tuple, $\tilde a'\sim U(\mathcal{A})$ is drawn independently of the original step-$h$ data and $s''\sim P^\star_h(\cdot\mid s',\tilde a')$ shares the same conditional law as the step-$h$ next-state distribution given $(s',\tilde a')$. By
Lemma~\ref{lem:aux-varproxy}, for any $\tilde\phi\in\Phi_h$, $v\in\mathcal{V}_h$, and
$\tilde w\in\mathbb{R}^d$,
\[
\mathbb{E}_{\mathcal{D}_h'}\big[\mathcal{L}_{\lambda_k,\mathcal{D}_h'}(\tilde\phi,v,\tilde w)\big]
\;\ge\;\mathbb{E}_{(s,\tilde a)\sim\rho_h^k}\!\big[\mathbb{V}_{s'\sim P^\star_h}[v(s')]\big],
\]
with equality when $\tilde\phi$ and $\tilde w$ are chosen optimally. Subtracting the
$\tilde w$-minimized version of $\mathcal{L}_{\lambda_k,\mathcal{D}_h'}$ from
$\mathcal{L}_{\lambda_k,\mathcal{D}_h}$ therefore eliminates the conditional variance
in expectation, recovering an unbiased estimator of the population objective up to
the $\lambda_k\|w\|^2$ regularization. This motivates the bias-corrected objective
solved by Algorithm~\ref{alg:replearn}:
\begin{equation}\label{eq:bc-objective}
\operatorname*{arg\,min}_{\phi\in\Phi_h}\;\max_{v\in\mathcal{V}_h}\;\Big\{
\min_{w}\mathcal{L}_{\lambda_k,\mathcal{D}_h}(\phi,v,w)
\;-\;\min_{\tilde w,\,\tilde\phi\in\Phi_h}\mathcal{L}_{\lambda_k,\mathcal{D}_h'}(\tilde\phi,v,\tilde w)
\Big\}.
\end{equation}

\subsection{Iterative Procedure}

Algorithm~\ref{alg:replearn} alternates between two steps: an adversarial
discriminator search that identifies the worst-case $v\in\mathcal{V}_h$ for the
current representation $\phi^m$ (line~6), and a representation update that fits
$\phi^{m+1}$ jointly against all discriminators encountered so far (line~11). The
inner $\max_{\tilde\phi}$ on line~6 implements the bias correction of
\eqref{eq:bc-objective}: it subtracts the best variance proxy that any
$\tilde\phi\in\Phi_h$ can achieve. Termination occurs once no discriminator achieves
an empirical objective above $\ell_k$, indicating that all $v\in\mathcal{V}_h$ are
approximately linearly representable by $\phi^m$ at the prescribed statistical
precision.

\subsection{Theoretical Guarantee and Stopping Rule}

The output $\hat\phi_h^k$ of Algorithm~\ref{alg:replearn} satisfies the
Representation Learning Guarantee (Lemma~\ref{lem:aux-replearn}): with probability
at least $1-\delta$,
\[
\max_{v\in\mathcal V_h}\;
\mathbb{E}_{(s,\tilde a)\sim\rho_h^k}\!\Big[\big(\hat w_v^\top\hat\phi_h^k(s,\tilde a)
-\mathbb{E}_{s'\sim P^\star_h}[v(s')]\big)^2\Big]
\;\le\;\zeta_k \;=\; O\!\big(d^2\sqrt{\log(k|\Phi|/\delta)/k}\big),
\]
where $\hat w_v$ is the inner minimizer corresponding to $v$. Combined with
Theorem~\ref{thm:mfrlc}, this guarantees that the policies computed from
$\hat\phi_h^k$ achieve sublinear regret and constraint violation, ensuring that the
learned representation can be effectively used in policy evaluation and improvement
within the CMDP framework.

\paragraph{Stopping rule and optimization error.}
The theoretical analysis requires the threshold $\ell_k$ to scale as
$\widetilde O(d^2/\sqrt k)$ to match the statistical error rate. An
implementation of Algorithm~\ref{alg:replearn} enforces this by evaluating the
maximization objective on line~6 at each iteration $m$ and terminating once it
falls below $\ell_k$, so the residual optimization error remains negligible
compared to the statistical estimation error and the cumulative bound
$\sum_k\zeta_k$ used in the analysis is not inflated by optimization slack.
Theorem~\ref{thm:mfrlc} applies whenever the optimization steps of
Algorithm~\ref{alg:replearn} attain the residual displayed above. For finite
outer classes this can be certified by exact search, and an approximate solver
is also covered if its optimization error is $O(\zeta_k)$. The contrastive EMA
surrogate used in our experiments (Appendix~\ref{app:nets}) does not certify
this condition and is not covered by Theorem~\ref{thm:mfrlc}.

\subsection{Computational Cost}\label{app:replearn-cost}
For fixed $(\phi,v)$, the inner problem $\min_w\mathcal{L}_{\lambda_k,\mathcal{D}_h}(\phi,v,w)$ is ridge regression. With $n=|\mathcal{D}_h|$ samples, forming the Gram matrix costs $O(nd^2)$ and solving the $d\times d$ system costs $O(d^3)$, so one solve costs $O(nd^2+d^3)$. At iteration $m$ of Algorithm~\ref{alg:replearn}, line~6 evaluates two ridge solves for each pair $(v,\tilde\phi)\in\mathcal{V}_h\times\Phi_h$, and line~11 solves $m+1$ ridge problems for each $\phi\in\Phi_h$. When $\Phi_h$ and $\mathcal{V}_h$ are finite (or replaced by finite covers), every step can therefore be computed exactly, and one call with at most $M$ iterations costs:
\[
O\big((M|\Phi_h||\mathcal{V}_h| + M^2|\Phi_h|)(nd^2+d^3)\big).
\]
This is polynomial in $n$ and $d$ but may be large in the class cardinalities. For continuously parameterized classes such as neural networks, the same empirical objective~\eqref{eq:bc-objective} is explicit, but its outer optimization is non-convex, and we do not prove polynomial-time convergence to the global minimax solution. Given $\hat\phi^k$, Algorithm~\ref{alg:lsvi} performs two backward passes of ridge regression over $k-1$ samples per step, and the projected dual update~\eqref{eq:dualupdate} costs $O(1)$.

\section{Experiment Details}
\label{app:experiments}

This section provides a detailed overview of the experimental setup, including environment specifics, network architectures, training protocols, and a comprehensive list of hyperparameters used for all algorithms to ensure reproducibility.

\subsection{Environment Setup}\label{app:setup}
All experiments are conducted in our custom-designed \textbf{Constrained Latent ComboLock (CLCL)} environment. We formulate constraints as costs bounded above ($\mathrm{cost}\le \mathcal C_{\max} = b$), which is equivalent to the main-text utility convention via $V_g = -\mathrm{cost}$, $b_{\mathrm{utility}} = -\mathcal C_{\max}$, so that $V_g\ge b_{\mathrm{utility}}$ corresponds to the cost staying below threshold.
\begin{itemize}
    \item \textbf{Horizons and action spaces:} The main experiments use $H=6$ and $|\mathcal{A}|=3$. The additional settings in Table~\ref{tab:extended} enlarge the action set to $|\mathcal{A}|=5$ at $H=6$, and use $H\in\{8,12\}$ with $|\mathcal{A}|=3$.
    \item \textbf{Observation Space:} The agent receives high-dimensional observations in the form of $8 \times 84 \times 84$ image tensors. Each tensor is procedurally generated based on the agent's latent state $s_h$ and the current timestep $h$. The eight channels are specifically designed to encode rich information:
    \begin{itemize}
        \item \textbf{Channels 0--2 (Progress and Position):} These channels encode the agent's progress through the latent state chain.

        Channel 0 contains a Gaussian ``spotlight'' whose horizontal position corresponds to the current state $s_h$. Specifically, for a pixel at coordinates $(x, y) \in [0, 1]^2$, the value is generated as:
        \[
        o^{(0)}_{x,y} = \exp\!\left(-20\!\left[\!\left(x - \frac{s_h+1}{H+1}\right)^2 + (y - 0.5)^2\right]\right).
        \]

        Channels 1 and 2 contain sine and cosine patterns whose frequencies change with the agent's progress, providing additional positional information. They are defined as:
        \[
        o^{(1)}_{x,y} = 0.5\sin((s_h+1)\pi x), \qquad o^{(2)}_{x,y} = 0.5\cos((s_h+1)\pi y).
        \]
        These patterns ensure that each depth in the latent chain has a distinct spatial frequency signature.

        \item \textbf{State Textures:} A unique, pseudo-random visual texture is overlaid on the first few channels. The choice of texture depends on the latent state, timestep, and previous action, making each observation visually distinct.
        \item \textbf{Channels 3--4 (Background Patterns):} These channels contain static, high-frequency sine and cosine waves that form a consistent background grid.
        \item \textbf{Channel 5 (Fail Indicator):} This channel acts as a binary flag. It is filled with ones if the agent is in the terminal $S_{\mathrm{Fail}}$ state, and zeros otherwise.
        \item \textbf{Channel 6 (Safe Indicator):} Similar to the fail indicator, this channel is filled with ones if the agent is in the terminal $S_{\mathrm{Safe}}$ state, and zeros otherwise.
        \item \textbf{Channel 7 (Timestep):} This channel's value is constant across the image and represents the normalized current timestep, i.e., $h/H$.
        \item \textbf{Noise:} Finally, Gaussian noise with a standard deviation of 0.1 is added to all channels to increase the perceptual difficulty of the task.
    \end{itemize}
    \item \textbf{Action Space:} The action space is discrete, containing three actions: $\mathcal{A} = \{a_{\mathrm{forward}}, a_{\mathrm{hazard}}, a_{\mathrm{stay}}\}$.
    \item \textbf{Rewards and Costs:} A reward of $+1$ is given only for successfully reaching the $S_{\mathrm{Safe}}$ state from the final state in the chain. A cost of $1$ is incurred for taking the $a_{\mathrm{hazard}}$ action. The cost constraint for all experiments is set to a threshold of $\mathcal{C}_{\mathrm{max}}=1.0$.
\end{itemize}

\subsection{Network Architectures}\label{app:nets}
The core of the visual processing for \textsc{MFRLC} and all visual baselines is a shared ResNet-style encoder.
\begin{itemize}
    \item \textbf{Visual Encoder:} We use a \texttt{VisualEncoder} that maps the $8 \times 84 \times 84$ observation to a 64-dimensional feature vector. It consists of a convolutional stem followed by four ResNet-style \texttt{BasicBlock} layers and a final linear layer.
    \item \textbf{Policy and Value Heads:} For actor-critic baselines (CPO and LAG-PG), this visual encoder serves as a shared backbone. The policy (actor), reward value (critic), and cost value (cost critic) are represented by separate linear heads that take the 64-dimensional feature vector as input.
    \item \textbf{Representation Learning:} For representation-learning methods (\textsc{MFRLC} and BRIEE), the encoder is part of a module with a two-layer \texttt{ProjectionHead} for contrastive learning and a target encoder updated via exponential moving average (EMA). Exact min-max-min optimization over $\Phi_h\times\mathcal V_h$ in Algorithm~\ref{alg:replearn} is intractable for deep encoders, so we replace it with contrastive next-state prediction on transition triples $(o,\tilde a,o')$. This surrogate is not covered by Theorem~\ref{thm:mfrlc} (Section~\ref{sec:replearn}).
    \item \textbf{Feature class $\Phi$:} We instantiate $\Phi$ as a trainable neural state--action representation class. The visual encoder maps each $8\times84\times84$ observation $o$ to a state representation $\psi(o)\in\mathbb{R}^{64}$. We combine $\psi(o)$ with a learned embedding $e_a$ of the discrete action and apply a two-layer action-conditioning network:
    \[
    \phi(o,a)=\mathrm{MLP}\big([\psi(o);e_a]\big)\in\mathbb{R}^{64}.
    \]
    The features $\phi(o,a)$ are the design vectors of the reward and utility ridge regressions in LSVI-UCB, so in practice $\Phi$ is the class of state--action feature maps induced by this parameterization.
\end{itemize}

\subsection{Training Protocol}\label{app:training}
To ensure a fair and rigorous comparison, all algorithms were trained and evaluated under a consistent protocol.
\begin{itemize}
    \item \textbf{Training:} Each algorithm is trained for a fixed number of iterations, as detailed in Table~\ref{tab:hyperparams}. At each iteration, a number of episodes are collected from the environment to update the agent's policy and/or representation.
    \item \textbf{Statistical Significance:} All learning curves presented in the main paper are the mean and standard error computed over 5 independent runs with the specific random seeds listed in Table~\ref{tab:hyperparams}.
    \item \textbf{Compute Resources:} All experiments were run on a single NVIDIA RTX 4060 GPU (8GB VRAM).
\end{itemize}

\subsection{Hyperparameters}\label{app:hyperparams}
The following table details the key hyperparameters used for our proposed method and all baseline algorithms.

\begin{table}[h]
\centering
\small
\setlength{\tabcolsep}{4pt}
\begin{tabular}{lccccc}
\toprule
\textbf{Hyperparameter} & \textbf{MFRLC} & \textbf{BRIEE} & \textbf{HOMER} & \textbf{CPO} & \textbf{LAG-PG} \\
\midrule
Replay Buffer Capacity & 60,000 & 60,000 & 60,000 & N/A (on-policy) & N/A (on-policy) \\
Episodes per Iteration & 48 & 48 & 48 & 48 & 48 \\
Batch Size & 512 & 512 & 512 & N/A (full batch) & N/A (full batch) \\
Learning Rate (LR) & 3e-4 & 3e-4 & 3e-4 & 3e-4 & 3e-4 \\
Dual LR ($\eta$) & 0.05 & 0.05 & 0.05 & 0.02 & 0.05 \\
Feature Dimension & 64 & 64 & 256 (hidden) & 64 (encoder) & 64 (encoder) \\
Discount Factor & 0.99 & 0.99 & 0.99 & 0.99 & 0.99 \\
Random Seeds & \multicolumn{5}{c}{0, 10, 29, 57, 78} \\
$p_{\mathrm{hazard}}$ & \multicolumn{5}{c}{0.15} \\
\bottomrule
\end{tabular}
\caption{Hyperparameters for all algorithms.}
\label{tab:hyperparams}
\end{table}

\subsection{Cumulative Regret and Violation}\label{app:cumulative}
Let $\bar R_t$ and $\bar C_t$ be the mean reward and mean cost over the $48$ deployed episodes of iteration $t$. Since reaching $S_{\mathrm{Safe}}$ yields reward $1$ and $a_{\mathrm{forward}}$ incurs no cost, the optimal feasible value is $V^{\pi^\star}_{r,1}(s_1)=1$. After $T$ iterations, we estimate the cumulative regret and cumulative violation as:
\[
\widehat{\mathrm{Regret}}=48\sum_{t=1}^{T}\big(1-\bar R_t\big),
\qquad
\widehat{\mathrm{Violation}}=\Big[\,48\sum_{t=1}^{T}\big(\bar C_t-\mathcal{C}_{\max}\big)\Big]_+,
\]
which follow the definitions in Section~\ref{sec:formulation} under the cost convention of Appendix~\ref{app:setup}. Table~\ref{tab:extended} reports the resulting cumulative violation for the additional settings. At $H=8$, the estimated cumulative regret of \texttt{MFRLC} is $15{,}382\pm942$ over $K=15{,}552$ deployed episodes ($324$ iterations). These are finite-run measurements; we do not fit or claim an empirical exponent.

\subsection{Additional Settings}\label{app:extended}
The three additional settings in Table~\ref{tab:extended} use the hyperparameters of Table~\ref{tab:hyperparams}, with the same $5$ seeds. The $H=8$ runs use $324$ iterations ($15{,}552$ deployed episodes). The $H=12$ stress test runs longer and evaluates only \texttt{MFRLC}, which reaches a best reward of $0.021\pm0.013$ with cumulative violation $1206.9\pm76.2$; under uniform exploration, discovering the complete rewarding sequence is $3^4=81$ times less likely than at $H=8$.

\subsection{Adapting Baselines for Fair Comparison}\label{app:baselines}
To create a level playing field, it was necessary to adapt the baseline algorithms that were not originally designed for Constrained Markov Decision Processes (CMDPs). We use CPO~\citep{Achiam2017CPO} as-is, since it is natively designed for 
CMDPs; only HOMER and BRIEE require Lagrangian adaptation as detailed below. Specifically, we extended \textsc{HOMER} and \textsc{BRIEE} with a standard Lagrangian multiplier method. The dual variable $\lambda$ is updated using the same primal--dual rule as in Algorithm~\ref{alg:mfrlc}, but since neither HOMER nor BRIEE maintains a separate cost critic, we estimate the cost value $\widehat V_g(s_1)$ used in the dual update as the difference between the augmented value $Q^\lambda$ (which is what these baselines actually learn) and the empirical reward return collected along the rollout, i.e.\ $\widehat V_g(s_1) \approx (\text{empirical reward return} - Q^\lambda(s_1,a_1))/\lambda$ when $\lambda>0$, with a small-$\lambda$ fallback to a Monte-Carlo cost return estimate. The application of the dual variable to the value learning is tailored to the unique architecture of each agent.

\paragraph{BRIEE with Constraints}
The \textsc{BRIEE} algorithm decouples representation learning from planning. We integrate the safety constraint directly into its planning module, which uses a variant of Least-Squares Value Iteration (LSVI). The planner's objective is to solve for an optimal action-value function, $Q^*$, by iteratively applying the Bellman optimality operator. We modify this objective by replacing the standard reward $R(s,a)$ with the augmented reward $R^\lambda(s,a) = R(s,a) - \lambda C(s,a)$.

Specifically, the planner now solves for a constraint-aware value function, $Q^\lambda$, according to the modified Bellman equation:
\[
Q^\lambda_h(s,a) \leftarrow \big( R_h(s,a) - \lambda C_h(s,a) \big) + \mathbb{E}_{s' \sim P_h(\cdot\mid s,a)} \!\left[ \max_{a'} Q^\lambda_{h+1}(s', a') \right].
\]
By solving for the optimal policy under this new objective, the planner learns to inherently avoid actions that are likely to violate the constraint.

\paragraph{HOMER with Constraints}
\textsc{HOMER} is a hierarchical, value-based algorithm that learns from off-policy data. We incorporate the constraint into its value function learning process by modifying the targets used for the temporal-difference (TD) update. The agent's critic is trained to minimize the TD error between its current estimate $Q(s_t, a_t)$ and a target value $y_t$. The standard TD target is $y_t = r_t + \gamma \max_{a'} Q(s_{t+1}, a')$.

To make the agent constraint-aware, we modify this target to use the augmented reward:
\[
y^\lambda_t \leftarrow \big( r_t - \lambda c_t \big) + \gamma \max_{a'} Q^\lambda(s_{t+1}, a').
\]
The critic is then trained by minimizing a loss function based on this new target, such as the mean squared error $\mathcal{L} = \mathbb{E}[(Q^\lambda(s_t, a_t) - y^\lambda_t)^2]$. This ensures that the learned value functions, which guide the agent's policy, account for both the expected future reward and the expected future cost, weighted by the dual variable $\lambda$.
\section{Proof of Decomposition}
\label{app:decomp}

In this section we prove the decomposition of regret and violation used in the
master inequality~\eqref{eq:decomp}.

\begin{lemma}\label{lem:proof-dualgap}
For any $Y\in[0,\xi]$, we have
\[
\sum_{k=1}^K (Y-Y_k)\bigl(b-V_{g,1}^k(s_1)\bigr)
\;\leq\; \frac{Y^2}{2\eta}+\frac{\eta H^2K}{2}.
\]
\end{lemma}

\begin{proof}
The dual update rule of Algorithm~\ref{alg:mfrlc} (line~12) is
\[
Y_{k+1} \;=\; \max\big\{\min\{Y_k + \eta(b - V_{g,1}^k(s_1)),\, \xi\},\, 0\big\}.
\]
Let $\Delta_k := b - V_{g,1}^k(s_1)$. Since this update is the Euclidean projection
of $Y_k + \eta \Delta_k$ onto the interval $[0, \xi]$, for any $Y \in [0, \xi]$,
\begin{align*}
(Y_{k+1} - Y)^2
&\leq (Y_k + \eta \Delta_k - Y)^2 \\
&= (Y_k - Y)^2 + 2\eta \Delta_k (Y_k - Y) + \eta^2 \Delta_k^2.
\end{align*}
Rearranging to bound the inner product,
\[
2\eta \Delta_k (Y - Y_k) \;\leq\; (Y_k - Y)^2 - (Y_{k+1} - Y)^2 + \eta^2 \Delta_k^2.
\]
Substituting $\Delta_k = b - V_{g,1}^k(s_1)$ and summing over $k=1,\dots,K$,
\begin{align*}
2\eta \sum_{k=1}^K (b - V_{g,1}^k(s_1))(Y - Y_k)
&\leq \sum_{k=1}^K \bigl((Y_k - Y)^2 - (Y_{k+1} - Y)^2\bigr) + \eta^2 \sum_{k=1}^K (b - V_{g,1}^k(s_1))^2 \\
&\leq (Y_1 - Y)^2 + \eta^2 H^2 K,
\end{align*}
where the last inequality follows from telescoping, dropping the negative term
$-(Y_{K+1}-Y)^2$, and using $|b - V_{g,1}^k(s_1)| \leq H$. Dividing by $2\eta$ and
noting $Y_1 = 0$,
\[
\sum_{k=1}^K (Y - Y_k)\bigl(b - V_{g,1}^k(s_1)\bigr) \;\leq\; \frac{Y^2}{2\eta} + \frac{\eta H^2 K}{2}.
\qedhere
\]
\end{proof}

\begin{lemma}\label{lem:proof-decomp}
For any $Y\in[0,\xi]$,
\begin{align*}
&\sum_{k=1}^K\bigl(V_{r,1}^{\pi^\star}(s_1)-V_{r,1}^{\pi^k}(s_1)\bigr)
+ Y\sum_{k=1}^K\bigl(b-V_{g,1}^{\pi^k}(s_1)\bigr) \\
&\qquad\leq\; \frac{Y^2}{2\eta}+\frac{\eta H^2K}{2}
+ \underbrace{\sum_{k=1}^K Y_k\bigl(b - V_{g,1}^{\pi^\star}(s_1)\bigr)}_{\text{Slack (non-positive)}} \\
&\qquad\quad
+ \underbrace{\sum_{k=1}^K\Bigl[\bigl(V_{r,1}^{\pi^\star}(s_1)+Y_kV_{g,1}^{\pi^\star}(s_1)\bigr)
- \bigl(V_{r,1}^k(s_1)+Y_kV_{g,1}^k(s_1)\bigr)\Bigr]}_{\text{Pseudo regret}} \\
&\qquad\quad
+ \underbrace{\sum_{k=1}^K\Bigl[\bigl(V_{r,1}^k(s_1)-V_{r,1}^{\pi^k}(s_1)\bigr)
+ Y\bigl(V_{g,1}^k(s_1)-V_{g,1}^{\pi^k}(s_1)\bigr)\Bigr]}_{\text{Estimation error}}.
\end{align*}
\end{lemma}

\begin{proof}
We start by decomposing the Lagrangian term involving the constraint violation.
Using Lemma~\ref{lem:proof-dualgap},
\begin{align*}
Y\sum_{k=1}^K\bigl(b-V_{g,1}^{\pi^k}(s_1)\bigr)
&= \sum_{k=1}^K (Y-Y_k)\bigl(b-V_{g,1}^k(s_1)\bigr)
 + \sum_{k=1}^K Y_k\bigl(b-V_{g,1}^k(s_1)\bigr)\\
&\quad + Y\sum_{k=1}^K \bigl(V_{g,1}^k(s_1)-V_{g,1}^{\pi^k}(s_1)\bigr) \\
&\leq \frac{Y^2}{2\eta} + \frac{\eta H^2 K}{2}
 + \sum_{k=1}^K \bigl(Y_k b - Y_k V_{g,1}^k(s_1)\bigr)\\
&\quad + Y \sum_{k=1}^K \bigl(V_{g,1}^k(s_1) - V_{g,1}^{\pi^k}(s_1)\bigr).
\end{align*}
We rewrite $\sum_{k=1}^K Y_k\bigl(b - V_{g,1}^k(s_1)\bigr)$ by adding and subtracting
$V_{g,1}^{\pi^\star}(s_1)$:
\[
\sum_{k=1}^K Y_k\bigl(b - V_{g,1}^k(s_1)\bigr)
= \sum_{k=1}^K Y_k\bigl(b - V_{g,1}^{\pi^\star}(s_1)\bigr)
+ \sum_{k=1}^K Y_k\bigl(V_{g,1}^{\pi^\star}(s_1) - V_{g,1}^k(s_1)\bigr).
\]
Since $\pi^\star$ is feasible, we have $V_{g,1}^{\pi^\star}(s_1)\geq b$, so
$b - V_{g,1}^{\pi^\star}(s_1)\leq 0$. Combined with $Y_k\geq 0$, the slack term
$\sum_{k=1}^K Y_k\bigl(b-V_{g,1}^{\pi^\star}(s_1)\bigr)$ is non-positive. We retain
it explicitly to make the decomposition transparent; dropping it tightens the bound
in the correct direction (recovering~\eqref{eq:decomp} in the main text).

Adding the reward regret $\sum_{k=1}^K\bigl(V_{r,1}^{\pi^\star}(s_1)-V_{r,1}^{\pi^k}(s_1)\bigr)$
to both sides and rearranging:
\begin{align*}
&\sum_{k=1}^K\bigl(V_{r,1}^{\pi^\star}(s_1)-V_{r,1}^{\pi^k}(s_1)\bigr)
+ Y\sum_{k=1}^K\bigl(b-V_{g,1}^{\pi^k}(s_1)\bigr) \\
&\quad\leq \frac{Y^2}{2\eta} + \frac{\eta H^2 K}{2}
+ \sum_{k=1}^K Y_k\bigl(b - V_{g,1}^{\pi^\star}(s_1)\bigr) \\
&\qquad + \sum_{k=1}^K \Bigl[\bigl(V_{r,1}^{\pi^\star}(s_1) + Y_k V_{g,1}^{\pi^\star}(s_1)\bigr)
 - \bigl(V_{r,1}^k(s_1) + Y_k V_{g,1}^k(s_1)\bigr)\Bigr] \\
&\qquad + \sum_{k=1}^K \Bigl[\bigl(V_{r,1}^k(s_1) - V_{r,1}^{\pi^k}(s_1)\bigr)
 + Y\bigl(V_{g,1}^k(s_1) - V_{g,1}^{\pi^k}(s_1)\bigr)\Bigr].
\end{align*}
Identifying the slack, pseudo regret, and estimation error terms completes the
proof.
\end{proof}
\section{Bounding Pseudo regret}
\label{app:pseudoregret}

\paragraph{Composite reward and notation.}
Throughout this section, fix an episode $k\in\{1,\dots,K\}$ and let
$Y_k\in[0,\xi]$ denote the dual iterate produced by Algorithm~\ref{alg:mfrlc}. We
work with the \emph{composite} stage reward
\[
\bar r^{(k)}_h(s,a) \;:=\; r_h(s,a) + Y_k\, g_h(s,a)\;\in\;[0,1+Y_k],
\]
and the corresponding composite cumulative value, which lies in $[0,(1+Y_k)H]$.
For any policy $\pi$ and any transition kernel $P$, define the composite value and
action-value functions
\[
V^{\pi, Y_k}_{P,h}(s) \;:=\; V^\pi_{P,r,h}(s) + Y_k\, V^\pi_{P,g,h}(s),
\qquad
Q^{\pi, Y_k}_{P,h}(s,a) \;:=\; Q^\pi_{P,r,h}(s,a) + Y_k\, Q^\pi_{P,g,h}(s,a).
\]
By linearity of expectations, $V^{\pi,Y_k}_{P,h}$ and $Q^{\pi,Y_k}_{P,h}$ are
precisely the value and action-value functions of policy $\pi$ in the auxiliary MDP
$(P, \bar r^{(k)})$. We further write
$\hat b^{(k)}_h(s,a) := (1+Y_k)\,\hat b_h(s,a)$ for the composite optimism bonus,
where $\hat b_h(s,a) = \beta\,\|\hat\phi_h(s,a)\|_{\Sigma_{\rho_h,\hat\phi_h}^{-1}}$
is the per-objective bonus computed in Algorithm~\ref{alg:lsvi}. Since
$0\le Y_k\le \xi$, we have $\hat b^{(k)}_h \le (1+\xi)\hat b_h$. The implicit
planning model $\hat P^k_h(s'|s,a):=\langle\hat\phi^k_h(s,a),\hat\mu^k_h(s')\rangle$
is defined as in Section~\ref{sec:analysis}; the density factor $\hat\mu^k_h$ is
recovered from the LSVI weights and is used only as an analysis device.

We define the following mixture distributions to facilitate the analysis. For any
$k,h$, let $\rho_h^k\in\Delta(\mathcal{S}\times\mathcal{A})$ be the data
distribution induced by Algorithm~\ref{alg:mfrlc}'s exploration step, namely
\[
\rho_h^k(s,a) \;=\; \tfrac{1}{k}\sum_{i=0}^{k-1} d_h^{\pi^i}(s)\, U(\mathcal{A})(a),
\]
i.e.\ the average of past on-policy state distributions paired with a uniform
action distribution. We similarly define
$\beta_h^k(s,a) := \tfrac{1}{k}\sum_{i=0}^{k-1}
\mathbb{E}_{\tilde s\sim d_{h-1}^{\pi^i},\,\tilde a\sim U(\mathcal{A})}
[P^\star_{h-1}(s\mid \tilde s,\tilde a)]\,U(\mathcal{A})(a)$ and
$\gamma_h^k(s,a):= \tfrac{1}{k}\sum_{i=0}^{k-1} d_h^{\pi^i}(s,a)$. When the
episode index $k$ is clear from context, we drop it and write
$\rho_h,\beta_h,\gamma_h$. All occurrences of the action variable in $\rho_h$ and
$\beta_h$ correspond to the uniform exploratory actions $\tilde a$ of
Algorithm~\ref{alg:mfrlc} (line~4); we keep the symbol $a$ inside expectations for
notational compactness, with the understanding that $a\sim U(\mathcal{A})$.

\paragraph{Probability budget.}
Throughout the analysis we set the RepLearn confidence parameter
$\delta=p/4$, so that union bounding the RepLearn guarantee
(Lemma~\ref{lem:aux-replearn}), the concentration event
$\mathcal E$ (Lemma~\ref{lem:concentration}), and the martingale
deviation (Lemma~\ref{lem:proof-azuma}) yields total probability at least $1-p$ as
claimed in Theorem~\ref{thm:mfrlc}.

\paragraph{Discriminator class.}
Recall the discriminator class $\mathcal{V}_h = \mathcal{F}_h\cup\mathcal{G}_h$
designed as follows:
\begin{align*}
\mathcal{F}_h &:= \left\{
s\mapsto \mathbb{E}_{a\sim \pi^{k-1}_h(\cdot|s)}\!\big[\phi(s,a)^\top\theta-\phi'(s,a)^\top\theta'\big]:
\phi,\phi'\in\Phi_{h+1},\ \max(\|\theta\|_\infty,\|\theta'\|_\infty)\le 1
\right\},\\
\mathcal{G}_h &:= \bigg\{
s\mapsto \max_{a\in\mathcal A}\!\Big(
\tfrac{r_{h+1}(s,a)+g_{h+1}(s,a)+\min\{w^\top\phi(s,a),\,2\}}{2H+1}+\phi(s,a)^\top w'
\Big):\\
&\qquad\qquad\phi\in\Phi_{h+1},\ \|w\|_{\infty}\le c,\ \|w'\|_{\infty}\le 1
\bigg\}.
\end{align*}
The class $\mathcal{F}_h$ captures, via linear function approximation, all
functions of the form
\[
f(s) \;=\; \mathbb{E}_{a\sim \pi^{k-1}_h(\cdot|s)}\!\Big[\mathbb{E}_{s'\sim \hat P_h(\cdot|s,a)}[V^{\pi^\star,Y_k}_{P^\star,h+1}(s')]
- \mathbb{E}_{s'\sim P^\star_h(\cdot|s,a)}[V^{\pi^\star,Y_k}_{P^\star,h+1}(s')]\Big],
\]
which appears in the proof of Lemma~\ref{lem:optimism} after a uniform-action importance-sampling step that converts an $\mathbb{E}_{a\sim\pi^\star_h}$ expectation into an $\mathbb{E}_{a\sim\pi^{k-1}_h}$ expectation, contributing a $\sqrt{|\mathcal A|}$ factor that is already tracked in our final $|\mathcal A|^{3/2}$ rate. The class $\mathcal{G}_h$
captures (after rescaling by $(1+Y_k)$) the linear part of the composite optimistic action-value
\[
g(s) \;=\; \tfrac{1}{(2H+1)(1+Y_k)}\,Q^{\pi^k,Y_k}_{\hat P,\,\bar r^{(k)}+\hat b^{(k)},\,h+1}(s),
\]
which appears in the proof of Lemma~\ref{lem:policygap}. The non-linear softmax operator and UCB bonus produced by Algorithm~\ref{alg:lsvi} are not contained in $\mathcal{G}_h$; their contribution is handled separately via Lemma~\ref{lem:proof-softmax} (the softmax-vs-max gap, contributing $H\log|\mathcal{A}|/\alpha$) and Lemma~\ref{lem:lsvi-prop} (the LSVI-UCB residual, dominated by $\sum_k\epsilon_k$). With the simpler
definition above (matching the main text), elements of $\mathcal{G}_h$ are bounded
by $1+Y_k\le 1+\xi$ rather than $1$; this enters as a multiplicative
$(1+\xi)$ factor in the RepLearn statistical error wherever
$\mathcal{G}_h$ is used, which we track explicitly below.

\medskip

We first establish two ``one-step'' inequalities that decompose the on-policy
value of an arbitrary function $g_h$ into a bilinear feature term plus a residual.

\begin{lemma}[One-step inequality for the learned model]\label{lem:proof-onestep-hat}
Consider functions $g_h:\mathcal{S}\times\mathcal{A}\to\mathbb{R}$ with
$\|g_h\|_\infty\le B$. Assume that
$s\mapsto \mathbb{E}_{a\sim \pi_h(\cdot|s)}[g_{h+1}(s,a)]\in\mathcal{F}_h$.
Conditioned on the Representation Learning Guarantee, for any policy $\pi$,
\begin{align*}
\sum_{h=0}^{H-1}\mathbb{E}_{(s,a)\sim d_{\hat{P},h}^\pi}[g_h(s,a)]
&\leq\sum_{h=1}^{H-1}\mathbb{E}_{(\tilde{s},\tilde{a})\sim d_{\hat{P},h-1}^\pi}\!
\left[\|\hat{\phi}_{h-1}(\tilde{s},\tilde{a})\|_{\Sigma_{\rho_{h-1},\hat{\phi}_{h-1}}^{-1}}\right.\\
&\qquad\left.\cdot\sqrt{k|\mathcal{A}|^{2}\mathbb{E}_{(s,a)\sim\beta_{h}}\!\left[g_{h}^2(s,a)\right]+B^2\lambda_kd+k|\mathcal{A}|^2\zeta_k}\right] \\
&\quad+\sqrt{|\mathcal{A}|\,\mathbb{E}_{(s,a)\sim\rho_0}[g_0^2(s,a)]}.
\end{align*}
\end{lemma}

\begin{proof}
We consider the inequality in two cases.

\textbf{Case 1: $h=0$.}
\begin{align*}
\mathbb{E}_{(s,a)\sim d_{\hat{P},0}^\pi}[g_0(s,a)]
&=\mathbb{E}_{s\sim d_0,\,a\sim\pi_0(\cdot|s)}[g_0(s,a)]
\;\leq\; \sqrt{|\mathcal{A}|\,\mathbb{E}_{(s,a)\sim\rho_0}\!\left[g_0^2(s,a)\right]},
\end{align*}
where the inequality uses Cauchy--Schwarz with the importance ratio
$\frac{d_0(s)\pi_0(a|s)}{d_0(s)U(a)}\le|\mathcal{A}|$, since the actions in the
replay buffer are sampled uniformly.

\textbf{Case 2: $h=1,\dots,H-1$.}
\begin{equation}\label{eq:onestep-hat}
\begin{aligned}
\mathbb{E}_{(s,a)\sim d_{\hat{P},h}^{\pi}}[g_{h}(s,a)]
&=\mathbb{E}_{(\tilde{s},\tilde{a})\sim d_{\hat{P},h-1}^{\pi}}\!\left[\hat\phi_{h-1}(\tilde s,\tilde a)^\top\!\!\int\!\!\sum_a \hat\mu_{h-1}(s)\pi_h(a|s) g_h(s,a)\,ds\right]\\
&\leq\mathbb{E}_{(\tilde{s},\tilde{a})\sim d_{\hat{P},h-1}^\pi}\!\bigg[\|\hat{\phi}_{h-1}\|_{\Sigma_{\rho_{h-1},\hat{\phi}_{h-1}}^{-1}}\\
&\qquad\cdot\Big\|\int\!\sum_a\hat{\mu}_{h-1}(s)\pi_{h}(a|s)g_h(s,a)ds\Big\|_{\Sigma_{\rho_{h-1},\hat{\phi}_{h-1}}}\bigg],
\end{aligned}
\end{equation}
where the inequality is Cauchy--Schwarz with respect to the positive-definite
matrix $\Sigma_{\rho_{h-1},\hat\phi_{h-1}} =
k\mathbb{E}_{\rho_{h-1}}[\hat\phi_{h-1}\hat\phi_{h-1}^\top]+\lambda_k I$.

We bound the Gram-norm term inside~\eqref{eq:onestep-hat}:
\begin{align*}
&\left\|\int\!\sum_a\hat{\mu}_h(s)\pi_h(a|s)g_{h+1}(s,a)\,ds\right\|^{2}_{\Sigma_{\gamma_h,\hat{\phi}_h}} \\
&\leq k\,\mathbb{E}_{(\tilde{s},\tilde{a})\sim\gamma_h}\!\left[\Big(\!\!\int\!\sum_a\hat{\mu}_h(s)^\top\hat{\phi}_h(\tilde{s},\tilde{a})\pi_h(a|s)g_{h+1}(s,a)ds\Big)^{\!2}\right] + B^2\lambda_k d \\
&\qquad\text{(using $\|\sum_a \pi_h(a|s)g_{h+1}\|_\infty\le B$ and $\|\!\int\!\hat\mu_h(s)f(s)ds\|_2\le\sqrt d$)}\\
&= k\,\mathbb{E}_{(\tilde{s},\tilde{a})\sim\gamma_h}\!\left[\Big(\mathbb{E}_{s\sim \hat P_h(\tilde s,\tilde a),\,a\sim\pi_h}[g_{h+1}(s,a)]\Big)^{\!2}\right] + B^2\lambda_k d \\
&\leq k\,\mathbb{E}_{(\tilde{s},\tilde{a})\sim\gamma_h}\!\left[\Big(\mathbb{E}_{s\sim P^\star_h(\tilde s,\tilde a),\,a\sim\pi_h}[g_{h+1}(s,a)]\Big)^{\!2}\right] + B^2\lambda_k d + k|\mathcal A|^2\zeta_k\\
&\qquad\text{(by the Representation Learning Guarantee)}\\
&\leq k|\mathcal{A}|^{2}\,\mathbb{E}_{(s,a)\sim\beta_{h+1}}\!\left[g_{h+1}^2(s,a)\right]+B^2\lambda_k d + k|\mathcal A|^2\zeta_k,
\end{align*}
where the last step uses uniform-action importance sampling and Jensen's
inequality. Substituting back into~\eqref{eq:onestep-hat} and summing over
$h=1,\dots,H-1$, then adding the Case~1 term, completes the proof.
\end{proof}

\begin{lemma}[One-step inequality for the true model]\label{lem:proof-onestep-true}
Consider functions $g_h:\mathcal S\times\mathcal A\to\mathbb R$ with
$\|g_h\|_\infty\leq B$. For any policy $\pi$,
\begin{align*}
\sum_{h=0}^{H-1}\mathbb{E}_{(s,a)\sim d_{P^\star,h}^\pi}[g_h(s,a)]
&\leq \sum_{h=1}^{H-1}\mathbb{E}_{(\tilde{s},\tilde{a})\sim d_{P^\star,h-1}^{\pi}}\!\left[\|\phi_{h-1}^\star(\tilde{s},\tilde{a})\|_{\Sigma_{\gamma_{h-1},\phi_{h-1}^\star}^{-1}}\right.\\
&\qquad\left.\cdot\sqrt{k|\mathcal{A}|\,\mathbb{E}_{(s,a)\sim\rho_{h}}\!\left[g_{h}^2(s,a)\right]+B^2\lambda_kd}\right]\\
&\quad+\sqrt{|\mathcal{A}|\,\mathbb{E}_{(s,a)\sim\rho_0}[g_0^2(s,a)]}.
\end{align*}
\end{lemma}

\begin{proof}
The argument is identical to the proof of Lemma~\ref{lem:proof-onestep-hat},
except that we work with the true low-rank decomposition
$P_h^\star(s'|s,a) = \langle\phi_h^\star(s,a),\mu_h^\star(s')\rangle$, so the
Representation Learning residual $k|\mathcal A|^2\zeta_k$ does not appear. The
factor $|\mathcal A|^2$ in Lemma~\ref{lem:proof-onestep-hat} is replaced by
$|\mathcal A|$ because the additional importance-sampling step from $\hat P$ to
$P^\star$ is unnecessary here.
\end{proof}

\medskip

We now establish the composite optimism guarantee, which extends
Lemma~\ref{lem:optimism} to the Lagrangian objective.

\noindent\textbf{Lemma~\ref{lem:optimism}} \textit{(Restated; composite optimism).}
\textit{Consider an episode $k\in\{1,\dots,K\}$. Conditioned on the Representation
Learning Guarantee, with probability at least $1-\delta$,}
\[
Q^{\pi^\star,Y_k}_{\hat P_k,\,\bar r^{(k)} + \hat b^{(k)}}(s,a)\;-\;Q^{\pi^\star,Y_k}_{P^\star,\,\bar r^{(k)}}(s,a)
\;\geq\; -(1+Y_k)\sqrt{|\mathcal{A}|\zeta_k}\;-\;\rho^{\mathrm{opt}}_k,
\]
\textit{where $\rho^{\mathrm{opt}}_k$ is a non-negative residual term satisfying
$\sum_{k=1}^K \rho^{\mathrm{opt}}_k \le (1+\xi)\sum_{k=1}^K \epsilon_k$
(see Lemma~\ref{lem:policygap} for the cumulative bound on $\sum_k\epsilon_k$).}

\begin{proof}
By the simulation lemma (Lemma~\ref{lem:aux-simulation}) applied to the auxiliary
MDP with stage reward $\bar r^{(k)}$, kernel $\hat P_k$, and additive bonus
$\hat b^{(k)} = (1+Y_k)\hat b$,
\begin{align*}
&Q^{\pi^\star,Y_k}_{\hat P_k,\,\bar r^{(k)} + \hat b^{(k)}}(s,a) - Q^{\pi^\star,Y_k}_{P^\star,\,\bar r^{(k)}}(s,a) \\
&\quad = \sum_{h=0}^{H-1}\mathbb{E}_{(s_h,a_h)\sim d_{\hat P,h}^{\pi^\star}}\!\left[\hat b^{(k)}_h(s_h,a_h) + \mathbb{E}_{s'\sim\hat P_h}[V^{\pi^\star,Y_k}_{P^\star,h+1}(s')] - \mathbb{E}_{s'\sim P^\star_h}[V^{\pi^\star,Y_k}_{P^\star,h+1}(s')]\right].
\end{align*}
Define
\[
f_h(s,a) \;:=\; \tfrac{1}{2H+1}\Big(\mathbb{E}_{s'\sim \hat P_h}[V^{\pi^\star,Y_k}_{P^\star,h+1}(s')] - \mathbb{E}_{s'\sim P^\star_h}[V^{\pi^\star,Y_k}_{P^\star,h+1}(s')]\Big).
\]
Since $V^{\pi^\star,Y_k}_{P^\star,h+1}/(1+Y_k)\in[0,H]$ and the inner expectations
are linear in $\hat\phi_h$ and $\phi^\star_h$ respectively,
$\mathbb{E}_{a\sim\pi^{k-1}_h}[(2H+1)f_h(s,a)/(1+Y_k)]\in\mathcal F_h$. The Representation Learning Guarantee bounds the squared error against $\mathbb{E}_{a\sim\pi^{k-1}_h}[\cdot]$; we transfer it to $\mathbb{E}_{a\sim\pi^\star_h}[\cdot]$ via uniform-action importance sampling: for any non-negative $u(s,a)$,
\[
\mathbb{E}_{a\sim\pi^\star_h(\cdot|s)}[u(s,a)] \;\le\; |\mathcal{A}|\cdot \mathbb{E}_{a\sim U(\mathcal{A})}[u(s,a)] \;=\; |\mathcal{A}|\cdot \mathbb{E}_{a\sim\pi^{k-1}_h}\!\big[u(s,a)\cdot\tfrac{1}{|\mathcal A|\pi^{k-1}_h(a|s)}\big],
\]
where the second equality holds whenever $\pi^{k-1}_h$ has full support; for softmax policies, $\pi^{k-1}_h(a|s)\ge \exp(-2\alpha H)/|\mathcal A|$, so the importance ratio is bounded. Combined with Cauchy--Schwarz, this yields
\[
\mathbb{E}_{(s,a)\sim\rho_h}[f_h^2]\le (1+Y_k)^2|\mathcal A|\,\zeta_k,\qquad \mathbb{E}_{(s,a)\sim\beta_h}[f_h^2]\le (1+Y_k)^2|\mathcal A|\,\zeta_k,
\]
where the additional $|\mathcal A|$ factor (relative to a direct $\zeta_k$ bound) comes from the importance-sampling transfer between $\pi^\star$ and $\pi^{k-1}$, and is absorbed into the $|\mathcal A|^{3/2}$ scaling of the final rate.

Note that $f_h$ is a signed quantity. Applying Lemma~\ref{lem:proof-onestep-hat}
to both $g_h := (2H+1) f_h$ and $g_h := -(2H+1) f_h$ yields a bound on
$\big|\sum_{h=0}^{H-1}\mathbb E_{d_{\hat P,h}^{\pi^\star}}[(2H+1) f_h]\big|$:
\begin{align*}
&\Big|\sum_{h=0}^{H-1}\mathbb{E}_{(s_h,a_h)\sim d_{\hat P,h}^{\pi^\star}}[(2H+1)f_h(s_h,a_h)]\Big| \\
&\le (2H+1)(1+Y_k)\bigg[\sum_{h=1}^{H-1}\mathbb{E}\Big[\|\hat\phi_{h-1}\|_{\Sigma^{-1}_{\rho_{h-1},\hat\phi_{h-1}}}\!\sqrt{k|\mathcal A|^3\zeta_k+4H^2\lambda_k d+k|\mathcal A|^2\zeta_k}\,\Big]\\
&\qquad+ \sqrt{|\mathcal A|^2\zeta_k}\,\bigg].
\end{align*}
The first sum (over $h=1,\dots,H-1$) is the per-episode RepLearn residual that we
denote $\rho^{\mathrm{opt}}_k$:
\[
\rho^{\mathrm{opt}}_k\;:=\;(2H+1)(1+Y_k)\sum_{h=1}^{H-1}\mathbb{E}\!\left[\|\hat\phi_{h-1}\|_{\Sigma^{-1}_{\rho_{h-1},\hat\phi_{h-1}}}\!\sqrt{2k|\mathcal A|^3\zeta_k+4H^2\lambda_k d}\right].
\]
The bonus $\hat b_h(s,a)=\beta\|\hat\phi_h\|_{\Sigma^{-1}_{\rho_h,\hat\phi_h}}$
with $\beta=C_1 dH\sqrt\iota$ does \emph{not} dominate $\rho^{\mathrm{opt}}_k$
per episode (since $\sqrt{k\zeta_k}\!=\!\widetilde O(d k^{1/4})$ has a $k$-dependent
factor). Instead, $\rho^{\mathrm{opt}}_k$ is carried as a residual into the
\emph{cumulative} representation error of Lemma~\ref{lem:policygap}.

To sum $\rho^{\mathrm{opt}}_k$ over $k$, the inner expectation is taken under the trajectory measure $d_{\hat P,h-1}^{\pi^\star}$ of the unexecuted optimal policy $\pi^\star$, while the Gram matrix $\Sigma_{\rho_{h-1},\hat\phi_{h-1}}$ is built from data sampled under the executed policies $\pi^0,\dots,\pi^{k-1}$. We bridge the two via uniform-action importance sampling: since the actions in $\rho_{h-1}$ are uniform, for any non-negative function $u(s,a)$,
\[
\mathbb{E}_{(s,a)\sim d_{\hat P,h-1}^{\pi^\star}}[u(s,a)] \;\le\; |\mathcal{A}|\cdot \mathbb{E}_{s\sim d_{\hat P,h-1}^{\pi^\star},\,a\sim U(\mathcal A)}[u(s,a)] \;\le\; |\mathcal{A}|\cdot \mathbb{E}_{(s,a)\sim\rho_{h-1}^{\pi^\star}}[u(s,a)],
\]
where $\rho_{h-1}^{\pi^\star}$ is the $\pi^\star$-induced state distribution paired with uniform action. Combined with Cauchy--Schwarz applied to $\sum_k\|\hat\phi_{h-1}\|_{\Sigma^{-1}_{\rho_{h-1}}}$, this reduces the sum to an elliptical-potential expression on the realized data distribution, which is controlled by Lemma~\ref{lem:aux-detlemma}:
\[
\sum_{k=1}^K \rho^{\mathrm{opt}}_k \;\le\; (1+\xi)\sum_{k=1}^K \epsilon_k \;=\;\widetilde O\!\left(\frac{H^3 d^2 K^{3/4}|\mathcal A|^{3/2}}{\gamma}\right).
\]
The lower bound $-(1+Y_k)\sqrt{|\mathcal A|\zeta_k}-\rho^{\mathrm{opt}}_k$ then
comes from the residual Case-1 term plus the residual sum.
\end{proof}

\begin{remark}\label{rem:composite-bonus}
Operationally, Algorithm~\ref{alg:lsvi} computes a single bonus
$\hat b_h(s,a)=\beta\|\hat\phi_h\|_{\Sigma^{-1}_{\rho_h,\hat\phi_h}}$ per step,
which is added to both the reward and cost optimistic $Q$-functions
($Q^k_{r,h}$ and $Q^k_{g,h}$). The composite $Q$-function $Q^k_{r,h}+Y_k Q^k_{g,h}$
therefore inherits a composite bonus $(1+Y_k)\hat b_h = \hat b^{(k)}_h$, exactly
as used in Lemma~\ref{lem:optimism}. The bonus magnitude $\beta=C_1 dH\sqrt\iota$
is tuned to the LSVI-UCB concentration scale (Lemma~\ref{lem:concentration}) and
is independent of $k$; the RepLearn statistical residual is handled separately
through $\rho^{\mathrm{opt}}_k$ above and absorbed into $\sum_k\epsilon_k$.
\end{remark}

\medskip

We next establish the softmax sub-optimality lemma for the \emph{composite}
planning problem, which is the actual MDP that the policy update step in
Algorithm~\ref{alg:mfrlc} approximately solves.

\begin{lemma}[Composite softmax sub-optimality gap]\label{lem:proof-softmax}
Fix $Y\ge 0$ and a finite-horizon planning problem with transition kernel
$\hat P$, stage reward $\bar r := r + Y g$, additive bonus
$\hat b^{(Y)} := (1+Y)\hat b$, and horizon $H$. Let $\{\bar Q_h^\star\}_{h=1}^H$
be the optimal $Q$-functions under $(\hat P,\, \bar r + \hat b^{(Y)},\, H)$,
defined by
\begin{gather*}
\bar Q_h^\star(s,a) := \bar r_h(s,a)+\hat b^{(Y)}_h(s,a) + \mathbb E_{s'\sim \hat P_h(\cdot|s,a)}[\bar V_{h+1}^\star(s')],\\
\bar V_h^\star(s):=\max_{a\in\mathcal A} \bar Q_h^\star(s,a),\qquad \bar V_{H+1}^\star(s)\equiv 0.
\end{gather*}
Define the softmax policy
\[
\hat\pi_h(a|s) \;=\; \frac{\exp(\alpha \bar Q_h^\star(s,a))}{\sum_{a'\in\mathcal A}\exp(\alpha \bar Q_h^\star(s,a'))}.
\]
Then for any start state $s_1$ and any policy $\pi$,
\[
V_{\hat P,\,\bar r+\hat b^{(Y)}}^{\pi}(s_1) \;\le\; V_{\hat P,\,\bar r+\hat b^{(Y)}}^{\hat\pi}(s_1) \;+\; \frac{H\log|\mathcal A|}{\alpha}.
\]
\end{lemma}

\begin{proof}
Identical to the proof of the standard softmax sub-optimality bound (e.g., the
unconstrained version), with the composite stage reward $\bar r+\hat b^{(Y)}$
replacing the original reward. We prove by backward induction that for all $h$
and $s$,
\[
\bar V_h^\star(s) \;\le\; V_{h,\hat P,\,\bar r+\hat b^{(Y)}}^{\hat\pi}(s) + \tfrac{(H-h+1)\log|\mathcal A|}{\alpha}.
\]
The base $h=H+1$ is trivial. For the inductive step, the log-sum-exp inequality
gives $\max_a \bar Q^\star_h(s,a) \le \mathbb E_{a\sim\hat\pi_h}[\bar Q^\star_h(s,a)] + \tfrac{\log|\mathcal A|}{\alpha}$,
which combined with the inductive hypothesis on $\bar V^\star_{h+1}$ yields the
claim. Setting $h=1$ and using $V^\pi_{\hat P,\bar r+\hat b^{(Y)}}\le \bar V^\star_1$
completes the proof.
\end{proof}

\begin{remark}[Why composite, not reward-only]\label{rem:why-composite}
The policy $\pi^k$ produced by Algorithm~\ref{alg:lsvi} is the softmax of
$Q^k_{r,h} + Y_k Q^k_{g,h}$, i.e.\ the LSVI-UCB estimate of the optimal
$Q$-function for the composite planning MDP $(\hat P, \bar r^{(k)}+\hat b^{(k)})$.
It is \emph{not} the softmax of the optimal $Q$-function for $(\hat P, r+\hat b)$
alone. Consequently, applying Lemma~\ref{lem:proof-softmax} to the reward-only
objective would not be justified, because $\pi^k$ is not constructed from the
reward-only optimal $Q$. The composite formulation above resolves this issue:
$\pi^k$ \emph{is} (an approximation to) the softmax of the composite optimal $Q$,
so Lemma~\ref{lem:proof-softmax} applies directly with $Y=Y_k$.
\end{remark}

\medskip

We now bound the composite $Q$-function gap on the true MDP, which is the central
result of this section. Following the convention of Section~\ref{sec:formulation},
we state the result in cumulative form, since this is the form actually used
downstream and the form genuinely produced by the elliptical-potential argument.

\noindent\textbf{Lemma~\ref{lem:policygap}} \textit{(Restated; composite policy
gap, cumulative form).} \textit{Conditioned on the Representation Learning
Guarantee, with probability at least $1-\delta$,}
\[
\sum_{k=1}^K\!\Big[
Q^{\pi^\star,Y_k}_{P^\star,\,\bar r^{(k)}}(s_1,\,\cdot)
\;-\; Q^{\pi^k,Y_k}_{P^\star,\,\bar r^{(k)}}(s_1,\,\cdot)
\Big]
\;\le\; (1+\xi)\sum_{k=1}^K \epsilon_k \;+\; \frac{KH\log|\mathcal A|}{\alpha},
\]
\textit{where the cumulative representation error satisfies}
\[
\sum_{k=1}^K\epsilon_k \;\le\; C\,H^2 d^2 K^{3/4}|\mathcal{A}|^{3/2}
\log^{1/4}\!\big(dK|\Phi|/\delta\big).
\]

\begin{proof}
Throughout, write $\bar r := \bar r^{(k)} = r+Y_k g$ and
$\hat b^{(k)} = (1+Y_k)\hat b$. We chain three inequalities at the per-episode
level, then sum.

\textbf{Step 1: Composite optimism.} By Lemma~\ref{lem:optimism} applied with the
composite reward,
\[
Q^{\pi^\star,Y_k}_{P^\star,\bar r}(s_1,a) \;\le\; Q^{\pi^\star,Y_k}_{\hat P,\,\bar r+\hat b^{(k)}}(s_1,a) + (1+Y_k)\sqrt{|\mathcal A|\zeta_k} + \rho^{\mathrm{opt}}_k,
\]
where the residual $\rho^{\mathrm{opt}}_k$ enters $\sum_k\epsilon_k$ in the
cumulative bound below.

\textbf{Step 2: Composite softmax sub-optimality.} By
Lemma~\ref{lem:proof-softmax} applied with $Y=Y_k$, since $\pi^k$ is (the
LSVI-UCB approximation to) the softmax of the optimal composite $Q$-function
under $(\hat P, \bar r+\hat b^{(k)})$,
\[
Q^{\pi^\star,Y_k}_{\hat P,\,\bar r+\hat b^{(k)}}(s_1,a) \;\le\; Q^{\pi^k,Y_k}_{\hat P,\,\bar r+\hat b^{(k)}}(s_1,a) + \frac{H\log|\mathcal A|}{\alpha}.
\]

\textbf{Step 3: From planning model $\hat P$ back to true model $P^\star$.}
Combining Steps 1--2 and using the simulation lemma (Lemma~\ref{lem:aux-simulation})
on the auxiliary MDP with kernel $\hat P$, reward $\bar r$, and bonus
$\hat b^{(k)}$,
\begin{equation}\label{eq:proof-policygap}
\begin{aligned}
&Q^{\pi^\star,Y_k}_{P^\star,\bar r}(s_1,a) - Q^{\pi^k,Y_k}_{P^\star,\bar r}(s_1,a) \\
&\le \sum_{h=0}^{H-1}\mathbb E_{(s_h,a_h)\sim d^{\pi^k}_{P^\star,h}}\!\!\left[\hat b^{(k)}_h(s_h,a_h) + \mathbb E_{s'\sim\hat P_h}[V^{\pi^k,Y_k}_{\hat P,\bar r+\hat b^{(k)},h+1}(s')] - \mathbb E_{s'\sim P^\star_h}[V^{\pi^k,Y_k}_{\hat P,\bar r+\hat b^{(k)},h+1}(s')]\right] \\
&\quad + (1+Y_k)\sqrt{|\mathcal A|\zeta_k} + \rho^{\mathrm{opt}}_k + \frac{H\log|\mathcal A|}{\alpha}.
\end{aligned}
\end{equation}
Define
\[
f_h(s,a) \;:=\; \mathbb E_{s'\sim\hat P_h(\cdot|s,a)}\!\big[V^{\pi^k,Y_k}_{\hat P,\bar r+\hat b^{(k)},h+1}(s')\big] - \mathbb E_{s'\sim P^\star_h(\cdot|s,a)}\!\big[V^{\pi^k,Y_k}_{\hat P,\bar r+\hat b^{(k)},h+1}(s')\big].
\]
The composite value $V^{\pi^k,Y_k}_{\hat P,\bar r+\hat b^{(k)},h+1}$ is bounded
in $[0,(1+Y_k)(2H+1)H]$, so $\frac{1}{(2H+1)(1+Y_k)}f_h$ matches the linear part of an element of $\mathcal G_h$ up to a multiplicative $(1+Y_k)$ factor; the softmax operator and UCB bonus appearing in $V^{\pi^k,Y_k}_{\hat P,\bar r+\hat b^{(k)},h+1}$ are not in $\mathcal G_h$, but their contribution is independently bounded as $H\log|\mathcal A|/\alpha$ via Lemma~\ref{lem:proof-softmax} and as the LSVI-UCB residual $\mathrm{lsvi}_K \le \tfrac{1}{4}(1+\xi)\sum_k\epsilon_k$ via Lemma~\ref{lem:lsvi-prop}, both of which we already track in the master bound. Hence by the Representation Learning Guarantee applied to the linear part of $f_h$,
\[
\mathbb E_{(s,a)\sim\rho_h}[f_h^2] \le (2H+1)^2(1+Y_k)^2\zeta_k,\qquad \mathbb E_{(s,a)\sim\beta_h}[f_h^2] \le (2H+1)^2(1+Y_k)^2\zeta_k.
\]

We rewrite~\eqref{eq:proof-policygap} as
\begin{equation}\label{eq:proof-policygap-decomp}
\begin{aligned}
Q^{\pi^\star,Y_k}_{P^\star,\bar r}(s_1,a) - Q^{\pi^k,Y_k}_{P^\star,\bar r}(s_1,a)
&\leq \underbrace{\sum_{h=0}^{H-1}\mathbb E_{d^{\pi^k}_{P^\star,h}}[\hat b^{(k)}_h(s_h,a_h)]}_{(\mathrm a)}
+ \underbrace{\sum_{h=0}^{H-1}\mathbb E_{d^{\pi^k}_{P^\star,h}}[f_h(s_h,a_h)]}_{(\mathrm b)} \\
&\quad + (1+Y_k)\sqrt{|\mathcal A|\zeta_k} + \rho^{\mathrm{opt}}_k + \frac{H\log|\mathcal A|}{\alpha}.
\end{aligned}
\end{equation}

\textbf{Bounding term (a).} Applying Lemma~\ref{lem:proof-onestep-true} with
$g_h = \hat b^{(k)}_h = (1+Y_k)\beta\|\hat\phi_h\|_{\Sigma^{-1}_{\rho_h,\hat\phi_h}}$
and $B = O((1+Y_k)\beta\sqrt d)$,
\begin{align*}
(\mathrm a)
&\le \sum_{h=1}^{H-1}\mathbb E_{d^{\pi^k}_{P^\star,h-1}}\!\!\left[\|\phi^\star_{h-1}\|_{\Sigma^{-1}_{\gamma_{h-1},\phi^\star_{h-1}}}\!\sqrt{k|\mathcal A|\,\mathbb E_{\rho_h}[(\hat b^{(k)}_h)^2] + (1+Y_k)^2 4H^2\lambda_k d}\right]\\
&\quad + \sqrt{|\mathcal A|\,\mathbb E_{\rho_0}[(\hat b^{(k)}_0)^2]}.
\end{align*}
Using $\mathbb E_{\rho_h}[(\hat b^{(k)}_h)^2] = (1+Y_k)^2\beta^2\,\mathbb E_{\rho_h}[\|\hat\phi_h\|^2_{\Sigma^{-1}_{\rho_h,\hat\phi_h}}] \le (1+Y_k)^2\beta^2 d/k$
(by the trace bound $k\,\mathbb E_{\rho_h}[\|\hat\phi\|^2_{\Sigma^{-1}}]\le d$),
\[
(\mathrm a) \;\le\; (1+Y_k)\!\left[\sum_{h=1}^{H-1}\mathbb E_{d^{\pi^k}_{P^\star,h-1}}\!\!\left[\|\phi^\star_{h-1}\|_{\Sigma^{-1}_{\gamma_{h-1},\phi^\star_{h-1}}}\!\sqrt{|\mathcal A|\beta^2 d + 4H^2\lambda_k d}\right] + \sqrt{|\mathcal A|\beta^2 d/k}\right].
\]

\textbf{Bounding term (b).} Applying Lemma~\ref{lem:proof-onestep-true} with
$g_h = f_h$ and $B = O((1+Y_k)H)$,
\begin{align*}
(\mathrm b) \;\le\; (1+Y_k)\bigg[&\sum_{h=1}^{H-1}\mathbb E_{d^{\pi^k}_{P^\star,h-1}}\!\Big[\|\phi^\star_{h-1}\|_{\Sigma^{-1}_{\gamma_{h-1},\phi^\star_{h-1}}}\!\sqrt{k|\mathcal A|(2H+1)^2\zeta_k + 4H^2\lambda_k d}\,\Big]\\
&+ \sqrt{|\mathcal A|(2H+1)^2\zeta_k}\,\bigg].
\end{align*}

\textbf{Summing over $k$ via the elliptical potential lemma.} Using
$\gamma^k_h(s,a) = \tfrac{1}{k}\sum_{i=0}^{k-1}d^{\pi^i}_h(s,a)$ and the executed-policy mixture $d^{\pi^k}_{P^\star,h-1}$ on the LHS, we apply uniform-action importance sampling to bound
\[
\mathbb{E}_{(s,a)\sim d^{\pi^k}_{P^\star,h-1}}\!\big[\|\phi^\star_{h-1}\|_{\Sigma^{-1}_{\gamma^k_{h-1},\phi^\star_{h-1}}}\big] \;\le\; \sqrt{|\mathcal A|}\cdot\mathbb{E}_{(s,a)\sim\gamma^k_{h-1}}\!\big[\|\phi^\star_{h-1}\|_{\Sigma^{-1}_{\gamma^k_{h-1},\phi^\star_{h-1}}}\big],
\]
since the actions in $\gamma^k_{h-1}$ follow the executed policies' marginals (paired with the on-policy state distribution) and the importance ratio between $d^{\pi^k}$ and $\gamma^k_{h-1}$ on the action variable is bounded by $|\mathcal A|$. Then
\begin{align*}
\sum_{k=1}^K \mathbb E_{\gamma^k_{h-1}}\!\!\left[\|\phi^\star_{h-1}\|_{\Sigma^{-1}_{\gamma^k_{h-1},\phi^\star_{h-1}}}\right]
&\le \sqrt{K\sum_{k=1}^K\mathbb E_{\gamma^k_{h-1}}[(\phi^\star_{h-1})^\top \Sigma^{-1}_{\gamma^k_{h-1},\phi^\star_{h-1}}\phi^\star_{h-1}]}\\
&\le \sqrt{dK\ln(1+\tfrac{K}{d\lambda_1})},
\end{align*}
where the second inequality uses Lemma~\ref{lem:aux-detlemma} (matrix determinant
lemma) on the matched data distribution. The $\sqrt{|\mathcal A|}$ factor from importance sampling combines with the $\sqrt{|\mathcal A|}$ inside the integrand to give the $|\mathcal A|^{3/2}$ scaling of the final bound.

\textbf{Combining.} Plugging in $\zeta_k = O(d^2\sqrt{\log(k|\Phi|/\delta)/k})$
and $\lambda_k = O(d\log(|\Phi|k/\delta))$, summing terms $(\mathrm a)+(\mathrm b)+\rho^{\mathrm{opt}}_k$
over $k$, and pulling the $(1+Y_k)$ factor out (which is bounded by $(1+\xi)$),
we obtain
\[
\sum_{k=1}^K\!\left[Q^{\pi^\star,Y_k}_{P^\star,\bar r}(s_1,a) - Q^{\pi^k,Y_k}_{P^\star,\bar r}(s_1,a)\right]
\;\le\; (1+\xi)\sum_{k=1}^K \epsilon_k + \frac{KH\log|\mathcal A|}{\alpha},
\]
where the cumulative representation error satisfies
\[
\sum_{k=1}^K \epsilon_k \;\le\; C\,H^2 d^{3/2}\sqrt K\,|\mathcal A|^{1/2}\,\sqrt{\log(dK|\Phi|/\delta)} \;+\; C'\,H^2 d^2 K^{3/4}|\mathcal A|^{3/2}\log^{1/4}(dK|\Phi|/\delta).
\]
The dominant term is the second one, of order
$\widetilde O(H^2 d^2 K^{3/4}|\mathcal A|^{3/2})$, completing the proof.
\end{proof}

\begin{remark}[Per-episode rate]\label{rem:per-episode}
The cumulative bound $\sum_k\epsilon_k = O(K^{3/4})$ corresponds to an average
per-episode rate $\bar\epsilon_K = \tfrac{1}{K}\sum_k\epsilon_k = \widetilde
O(H^2 d^2 K^{-1/4}|\mathcal A|^{3/2})$. Throughout the analysis, only the cumulative
quantity $(1+\xi)\sum_k\epsilon_k$ is used (e.g., in
Lemma~\ref{lem:pseudoregret} and the master inequality), so we work directly
with the cumulative form.
\end{remark}

\medskip

We now establish an auxiliary lemma that quantifies how close the LSVI-UCB
output $V^{(k),Y_k}_1(s_1)$ is to the true Bellman-optimal value
$\bar V^\star_1(s_1)$ on the planning model $(\hat P,\,\bar r^{(k)}+\hat b^{(k)})$.
This is the LSVI-UCB analog of the per-step backup bound (Lemma~\ref{lem:backup}),
but applied at the start state $s_1$ and going \emph{from} the LSVI iterate
\emph{to} the planning-model optimal value, rather than the other direction.

\begin{lemma}[LSVI-UCB output vs.~planning-model optimum]\label{lem:lsvi-prop}
On the event $\mathcal E$ of Lemma~\ref{lem:concentration}, for every episode
$k\in[K]$ and every $s\in\mathcal S$,
\[
\big|\,\bar V^\star_1(s) - \bar V^{(k),Y_k}_1(s)\,\big| \;\le\; \sum_{h=1}^H \mathbb E_{(s_h,a_h)\sim d^{\tilde\pi^k}_{\hat P,h}\big[\,\cdot\,\big|\,s_1=s\big]}\!\Big[(1+Y_k)\,C_3 dH\sqrt\chi\,\|\hat\phi^k_h(s_h,a_h)\|_{(\Sigma^k_h)^{-1}}\Big],
\]
where $\bar V^{(k),Y_k}_h(s) := \max_a Q^{(k),Y_k}_h(s,a)$ is the max-version of
the LSVI iterate, $\tilde\pi^k$ is an optimal policy of the planning model $\hat P$ with stage reward $(1+Y_k)C_3dH\sqrt\chi\,\|\hat\phi^k_h(s,a)\|_{(\Sigma^k_h)^{-1}}$, and
$C_3,\chi$ are as in Lemma~\ref{lem:backup}. Consequently, applying
Lemma~\ref{lem:proof-onestep-true} with the bonus integrand and summing over
$k\in[K]$ via the elliptical-potential lemma,
\[
\mathrm{lsvi}_K \;:=\; \sum_{k=1}^K \big|\bar V^\star_1(s_1) - \bar V^{(k),Y_k}_1(s_1)\big| \;\le\; \widetilde O\!\left(\frac{d^{3/2}H^2\sqrt K\,|\mathcal A|^{1/2}}{\gamma}\right) \;\le\; \tfrac{1}{4}(1+\xi)\sum_{k=1}^K \epsilon_k.
\]
The bound also holds for the softmax variant $V^{(k),Y_k}_1(s) = \sum_a \pi^k_1(a|s)Q^{(k),Y_k}_1(s,a)$
in place of $\bar V^{(k),Y_k}_1$, up to an additive $H\log|\mathcal A|/\alpha$ term
(softmax sub-optimality on the LSVI iterate itself).
\end{lemma}

\begin{proof}
We prove the per-state bound by backward induction on $h$, then derive the cumulative bound by elliptical-potential summation.

\paragraph{Step 1: Per-step LSVI concentration.}
By Lemma~\ref{lem:concentration} (event $\mathcal{E}$), for each $j\in\{r,g\}$, $h\in[H]$, and $k\in[K]$, the LSVI weight $w^k_{j,h}$ satisfies, for every $(s,a)\in\mathcal S\times\mathcal A$:
\begin{equation}\label{eq:lsvi-conc-perobj}
\Big|\langle\hat\phi^k_h(s,a),w^k_{j,h}\rangle - \big(c_{j,h}(s,a) + \mathbb E_{\hat P_h}[V^k_{j,h+1}](s,a)\big)\Big| \le C_3\, dH\sqrt{\chi}\,\|\hat\phi^k_h(s,a)\|_{(\Sigma^k_h)^{-1}},
\end{equation}
where $C_3 = O(\sqrt\lambda + C_2)$ absorbs the regularization and concentration constants from Lemma~\ref{lem:backup}, and the kernel-mismatch contribution between $P^\star$ and $\hat P$ is bundled into $C_3$ via the Representation Learning Guarantee (Lemma~\ref{lem:aux-replearn}). The reward $c_{r,h} = r_h$ and cost $c_{g,h} = g_h$ targets are deterministic.

\paragraph{Step 2: Composite LSVI concentration.}
Recall $Q^{(k),Y_k}_h(s,a) = Q^k_{r,h}(s,a) + Y_k Q^k_{g,h}(s,a)$ where each $Q^k_{j,h}$ is the LSVI iterate from Algorithm~\ref{alg:lsvi} (truncated by $H$ but we ignore truncation here as it only tightens the bound). The composite stage reward is $\bar r^{(k)}_h = r_h + Y_k g_h$, and the composite bonus is $\hat b^{(k)}_h = (1+Y_k)\hat b^k_h$ where $\hat b^k_h(s,a) = \beta\|\hat\phi^k_h(s,a)\|_{(\Sigma^k_h)^{-1}}$.

By Algorithm~\ref{alg:lsvi} lines~7--8, the LSVI iterate satisfies
\[
Q^k_{j,h}(s,a) = \langle\hat\phi^k_h(s,a),w^k_{j,h}\rangle + \hat b^k_h(s,a),
\]
ignoring the truncation $\min\{\cdot,H\}$ (which removes only over-estimates and does not affect the upper bound we derive). Combining with~\eqref{eq:lsvi-conc-perobj}, weighted by $1$ and $Y_k$ for the reward and cost components respectively:
\begin{align*}
\big|Q^{(k),Y_k}_h(s,a) &- \big[\bar r^{(k)}_h(s,a) + \hat b^{(k)}_h(s,a) + \mathbb E_{\hat P_h}\big[V^k_{r,h+1} + Y_k V^k_{g,h+1}\big](s,a)\big]\big| \\
&\le (1+Y_k)\, C_3\, dH\sqrt{\chi}\,\|\hat\phi^k_h(s,a)\|_{(\Sigma^k_h)^{-1}}.
\end{align*}
Since $V^{(k),Y_k}_{h+1}(s) := V^k_{r,h+1}(s) + Y_k V^k_{g,h+1}(s)$ is the composite LSVI value (softmax form) and $\bar V^{(k),Y_k}_{h+1}(s) = \max_a Q^{(k),Y_k}_{h+1}(s,a)$ is the max form, we work with the max form throughout this proof and handle the softmax-vs-max gap in Step~5. Replacing $V^k_{r,h+1} + Y_k V^k_{g,h+1}$ by $\bar V^{(k),Y_k}_{h+1}$ in the displayed inequality (which holds with the same RHS by the same per-objective argument applied to the max-version of the LSVI recursion), we obtain
\begin{equation}\label{eq:composite-conc}
\begin{aligned}
&\Big|Q^{(k),Y_k}_h(s,a) - \big[\bar r^{(k)}_h(s,a) + \hat b^{(k)}_h(s,a) + \mathbb E_{\hat P_h}[\bar V^{(k),Y_k}_{h+1}](s,a)\big]\Big|\\
&\qquad\le (1+Y_k)\,C_3\,dH\sqrt\chi\,\|\hat\phi^k_h(s,a)\|_{(\Sigma^k_h)^{-1}}.
\end{aligned}
\end{equation}

\paragraph{Step 3: Backward induction on the per-state gap.}
Define
\[
R_h(s) := \big|\bar V^\star_h(s) - \bar V^{(k),Y_k}_h(s)\big|, \quad h\in[H+1],
\]
where $\bar V^\star_{H+1} \equiv 0$ and $\bar V^{(k),Y_k}_{H+1}\equiv 0$, so $R_{H+1}(s) = 0$.

Write $c_h(s,a) := (1+Y_k)\,C_3\,dH\sqrt\chi\,\|\hat\phi^k_h(s,a)\|_{(\Sigma^k_h)^{-1}}$, and let $W_h$ be the optimal value function of the planning model $\hat P$ with stage reward $c_h$, that is, $W_{H+1}\equiv 0$ and:
\[
W_h(s) = \max_{a\in\mathcal A}\Big[c_h(s,a) + \mathbb E_{s'\sim\hat P_h(\cdot|s,a)}[W_{h+1}(s')]\Big].
\]
Let $\tilde\pi^k_h(s)$ be a maximizer in this recursion. Then $W_h(s)$ is the expected cumulative stage reward $c$ of $\tilde\pi^k$ on $\hat P$ from $s_h=s$. We claim that for all $h\in[H+1]$ and all $s\in\mathcal S$:
\begin{equation}\label{eq:induction-claim}
R_h(s) \le W_h(s) = \mathbb E_{(s_t,a_t)_{t=h}^{H}\sim d^{\tilde\pi^k}_{\hat P, h:H}(\,\cdot\,|s_h=s)}\bigg[\sum_{t=h}^H c_t(s_t,a_t)\bigg],
\end{equation}
where $d^{\tilde\pi^k}_{\hat P, h:H}(\,\cdot\,|s_h=s)$ denotes the joint distribution of $(s_h,a_h),\ldots,(s_H,a_H)$ under policy $\tilde\pi^k$ and kernel $\hat P$, conditioned on $s_h=s$.

\textit{Base case} ($h = H+1$): $R_{H+1}(s) = 0 = W_{H+1}(s)$.

\textit{Inductive step}: Assume~\eqref{eq:induction-claim} holds for $h+1$. By the Bellman equation for $\bar V^\star_h$ on the planning model $(\hat P, \bar r^{(k)}+\hat b^{(k)})$:
\[
\bar Q^\star_h(s,a) = \bar r^{(k)}_h(s,a) + \hat b^{(k)}_h(s,a) + \mathbb E_{\hat P_h}[\bar V^\star_{h+1}](s,a).
\]
Subtracting~\eqref{eq:composite-conc} and applying the triangle inequality, for every $a\in\mathcal A$:
\[
\big|\bar Q^\star_h(s,a) - Q^{(k),Y_k}_h(s,a)\big| \le \mathbb E_{s'\sim\hat P_h(\cdot|s,a)}[R_{h+1}(s')] + c_h(s,a).
\]
Since $\bar V^\star_h(s) = \max_a \bar Q^\star_h(s,a)$ and $\bar V^{(k),Y_k}_h(s) = \max_a Q^{(k),Y_k}_h(s,a)$, the contraction property $|\max_a f(a) - \max_a g(a)| \le \max_a |f(a) - g(a)|$ of the maximum over a finite set, followed by the inductive hypothesis $R_{h+1}\le W_{h+1}$, gives:
\begin{align*}
R_h(s) &\le \max_{a\in\mathcal A}\Big[\mathbb E_{s'\sim\hat P_h(\cdot|s,a)}[R_{h+1}(s')] + c_h(s,a)\Big]\\
&\le \max_{a\in\mathcal A}\Big[\mathbb E_{s'\sim\hat P_h(\cdot|s,a)}[W_{h+1}(s')] + c_h(s,a)\Big] = W_h(s).
\end{align*}
This establishes~\eqref{eq:induction-claim} for $h$ and completes the induction. Setting $h=1$, we obtain the per-state bound stated in the lemma.

\paragraph{Step 4: Cumulative bound via elliptical-potential summation.}
Setting $s = s_1$ in the per-state bound and summing over $k\in[K]$:
\begin{equation}\label{eq:lsvi-cumul}
\mathrm{lsvi}_K = \sum_{k=1}^K R_1(s_1) \le \sum_{k=1}^K \sum_{h=1}^H (1+Y_k)\,C_3\,dH\sqrt\chi\,\mathbb E_{(s_h,a_h)\sim d^{\tilde\pi^k}_{\hat P,h}}\big[\|\hat\phi^k_h(s_h,a_h)\|_{(\Sigma^k_h)^{-1}}\big].
\end{equation}

To control each summand, we apply Lemma~\ref{lem:proof-onestep-true} to the function $g_h(s,a) := C_3 dH\sqrt\chi\,\|\hat\phi^k_h(s,a)\|_{(\Sigma^k_h)^{-1}}$. Note that $\|g_h\|_\infty \le C_3 dH\sqrt\chi/\sqrt{\lambda} = O(dH\sqrt\chi)$ since $\Sigma^k_h \succeq \lambda I$. By Lemma~\ref{lem:proof-onestep-true} (applied with policy $\pi = \tilde\pi^k$ and kernel $P^\star$),\footnote{Lemma~\ref{lem:proof-onestep-true} bounds expectations under $d^{\pi}_{P^\star}$ in terms of empirical distributions $\rho^k_h$. Here we apply it with $\hat P$ in place of $P^\star$; the kernel-mismatch contribution is again absorbed into $C_3$ via the Representation Learning Guarantee, exactly as in the proof of Lemma~\ref{lem:policygap}, term $(\mathrm a)$.}
\[
\sum_{h=1}^H \mathbb E_{d^{\tilde\pi^k}_{\hat P,h}}\big[\|\hat\phi^k_h\|_{(\Sigma^k_h)^{-1}}\big] \le \sqrt{|\mathcal A|}\cdot\sum_{h=1}^H \mathbb E_{(s,a)\sim\rho^k_h}\big[\|\hat\phi^k_h(s,a)\|^2_{(\Sigma^k_h)^{-1}}\big]^{1/2} + (\text{lower-order terms}),
\]
where the factor $\sqrt{|\mathcal A|}$ comes from the importance-sampling step from $d^{\tilde\pi^k}_{\hat P,h}$ to $\rho^k_h$ (uniform-action distribution), and the lower-order terms arise from the regularization $\lambda_k$ as in Lemmas~\ref{lem:proof-onestep-hat}--\ref{lem:proof-onestep-true}.

By Cauchy--Schwarz,
\[
\sum_{k=1}^K \mathbb E_{(s,a)\sim\rho^k_h}\big[\|\hat\phi^k_h(s,a)\|^2_{(\Sigma^k_h)^{-1}}\big]^{1/2} \le \sqrt{K\cdot\sum_{k=1}^K \mathbb E_{\rho^k_h}\big[\|\hat\phi^k_h\|^2_{(\Sigma^k_h)^{-1}}\big]}.
\]
By Lemma~\ref{lem:aux-detlemma} (matrix determinant / elliptical potential lemma),
\[
\sum_{k=1}^K \mathbb E_{\rho^k_h}\big[\|\hat\phi^k_h\|^2_{(\Sigma^k_h)^{-1}}\big] \;\le\; 2\ln\frac{\det(\Sigma^{K+1}_h)}{\det(\Sigma^1_h)} \;\le\; 2 d\ln\big(1 + K/(d\lambda)\big) \;=\; \widetilde O(d).
\]
This is the rigorous form of the elliptical-potential bound. Substituting,
\[
\sum_{k=1}^K \mathbb E_{\rho^k_h}\big[\|\hat\phi^k_h\|^2_{(\Sigma^k_h)^{-1}}\big]^{1/2} \;\le\; \sqrt{K\cdot \widetilde O(d)} \;=\; \widetilde O(\sqrt{dK}).
\]

Putting these together, summing over $h\in[H]$ and noting $1+Y_k \le 1+\xi$:
\begin{align*}
\mathrm{lsvi}_K &\le (1+\xi)\,C_3\,dH\sqrt\chi\,\sqrt{|\mathcal A|}\,\sum_{h=1}^H \widetilde O(\sqrt{dK}) + \text{(lower-order)} \\
&= \widetilde O\big((1+\xi)\,d^{3/2} H^2 \sqrt K\,|\mathcal A|^{1/2}\big) \\
&= \widetilde O\!\left(\frac{d^{3/2} H^2 \sqrt K\,|\mathcal A|^{1/2}}{\gamma}\right),
\end{align*}
where the last equality uses $1+\xi = 1 + 2H/\gamma = O(H/\gamma)$ (treating $H$ as part of the polynomial factor).

\paragraph{Step 5: Domination by $\sum_k\epsilon_k$.}
By Lemma~\ref{lem:policygap}, the cumulative representation error satisfies
\[
\sum_{k=1}^K \epsilon_k \;\ge\; C\,H^2\,d^2\,K^{3/4}\,|\mathcal A|^{3/2}\,\log^{1/4}(dK|\Phi|/\delta)
\]
for some constant $C > 0$. Comparing exponents of $K$, $d$, and $|\mathcal A|$:
\begin{itemize}
\item $K$: $\sqrt K = K^{1/2}$ vs.\ $K^{3/4}$ -- the latter dominates for $K \ge 1$.
\item $d$: $d^{3/2}$ vs.\ $d^2$ -- the latter dominates for $d \ge 1$.
\item $|\mathcal A|$: $|\mathcal A|^{1/2}$ vs.\ $|\mathcal A|^{3/2}$ -- the latter dominates for $|\mathcal A| \ge 1$.
\end{itemize}
All exponents go in our favor, so for any $K \ge 1$ and any $d, |\mathcal A| \ge 1$,
\[
\widetilde O\!\left(\frac{d^{3/2} H^2 \sqrt K\,|\mathcal A|^{1/2}}{\gamma}\right) \le \tfrac{1}{4}\,(1+\xi)\sum_{k=1}^K \epsilon_k,
\]
where the constant $\tfrac{1}{4}$ is achieved by absorbing the dimension/cardinality factors into the leading constant of $\sum_k\epsilon_k$ (the constant $C$ in Lemma~\ref{lem:policygap} is large enough to accommodate this; equivalently, the relation holds for all sufficiently large $K$, and the small-$K$ regime is handled by the regret bound's trivial $H K$ ceiling).

\paragraph{Step 6: Softmax variant.}
The bound stated above is for $\bar V^{(k),Y_k}_1(s) := \max_a Q^{(k),Y_k}_1(s,a)$. Algorithm~\ref{alg:lsvi} actually outputs the softmax variant
\[
V^{(k),Y_k}_1(s) := \sum_a \pi^k_1(a|s)\,Q^{(k),Y_k}_1(s,a), \quad \pi^k_1(a|\cdot) = \mathrm{softmax}_\alpha(Q^{(k),Y_k}_1(\cdot,a)).
\]
Applying Lemma~\ref{lem:proof-softmax} (softmax sub-optimality) to the planning problem $(\hat P, \bar r^{(k)} + \hat b^{(k)})$ at horizon $1$ yields\footnote{Lemma~\ref{lem:proof-softmax} is stated for the optimal $Q$-function. Applied to LSVI iterates, the same telescoping argument gives the same per-step $\log|\mathcal A|/\alpha$ slack, summing to $H\log|\mathcal A|/\alpha$.}
\[
\big|\bar V^{(k),Y_k}_1(s) - V^{(k),Y_k}_1(s)\big| \le \frac{H\log|\mathcal A|}{\alpha}.
\]
Combining with the per-state bound for $\bar V^{(k),Y_k}_1$ via the triangle inequality:
\[
\big|\bar V^\star_1(s) - V^{(k),Y_k}_1(s)\big| \le \big|\bar V^\star_1(s) - \bar V^{(k),Y_k}_1(s)\big| + \frac{H\log|\mathcal A|}{\alpha},
\]
which gives the softmax-variant bound asserted in the lemma. Summing over $k$, the additional term contributes $KH\log|\mathcal A|/\alpha$, which (with $\alpha = K\log|\mathcal A|/(2(1+\xi+H))$) equals $2H(1+\xi+H) = O\!\left(\frac{H^2}{\gamma}\right)$, also dominated by $\sum_k\epsilon_k$.
\end{proof}

\medskip

We now derive the cumulative pseudo regret bound used in Step~1 of the proof of
Theorem~\ref{thm:mfrlc}. Recall from the master decomposition~\eqref{eq:decomp}
that the Pseudo regret quantifies the gap between $\pi^\star$ and the LSVI-UCB
optimistic estimate $V^k$, while the gap between $V^k$ and $V^{\pi^k}_{P^\star}$
is handled separately as the Estimation error in Appendix~\ref{app:estimation}.

\noindent\textbf{Lemma~\ref{lem:pseudoregret}} \textit{(Pseudo regret bound).}
\textit{Conditioned on the Representation Learning Guarantee and the event
$\mathcal E$, with probability at least $1-\delta$,}
\begin{align*}
\sum_{k=1}^K\Big[\big(V_{r,1}^{\pi^\star}(s_1)+Y_kV_{g,1}^{\pi^\star}(s_1)\big)
- \big(V_{r,1}^{k}(s_1)+Y_kV_{g,1}^{k}(s_1)\big)\Big]
\;\leq\; \frac{KH\log|\mathcal{A}|}{\alpha} + (1+\xi)\sum_{k=1}^K \epsilon_k.
\end{align*}

\begin{proof}
Using the composite notation $V^{\pi,Y_k}_{P^\star,1}(s_1) = V^\pi_{r,1}(s_1) + Y_k V^\pi_{g,1}(s_1)$
and writing $V^{(k),Y_k}_1(s_1) := V^k_{r,1}(s_1) + Y_k V^k_{g,1}(s_1)$ for the
LSVI-UCB optimistic estimate of the composite value, the per-episode cumulant on
the LHS is $V^{\pi^\star,Y_k}_{P^\star,1}(s_1) - V^{(k),Y_k}_1(s_1)$. We chain
three inequalities at the per-episode level, then sum.

\textbf{Step 1: Bridge LSVI iterate $V^{(k),Y_k}_1$ to planning-model optimum
$\bar V^\star_1$.} Let $\bar V^\star_h(s) := \max_a \bar Q^\star_h(s,a)$ denote
the value function of the true Bellman-optimal $Q$-function on the planning
model $(\hat P,\bar r^{(k)}+\hat b^{(k)})$, as defined in
Lemma~\ref{lem:proof-softmax}. The LSVI-UCB output is
$V^{(k),Y_k}_1(s_1) = \sum_a \pi^k_1(a|s_1) Q^{(k),Y_k}_1(s_1,a)$, where
$Q^{(k),Y_k}_h := Q^k_{r,h} + Y_k Q^k_{g,h}$ is the LSVI estimate of the
composite Bellman-optimal $Q$-function. By Lemma~\ref{lem:lsvi-prop}, on event
$\mathcal E$,
\begin{equation}\label{eq:lsvi-bridge}
\big|\bar V^\star_1(s_1) - V^{(k),Y_k}_1(s_1)\big| \;\le\; \mathrm{lsvi}^{(k)} + \frac{H\log|\mathcal A|}{\alpha},
\end{equation}
where $\mathrm{lsvi}^{(k)}$ is the per-episode LSVI residual of
Lemma~\ref{lem:lsvi-prop} (it sums to $\mathrm{lsvi}_K \le \tfrac{1}{4}(1+\xi)\sum_k\epsilon_k$),
and the additional $H\log|\mathcal A|/\alpha$ accounts for the softmax-vs-max
discrepancy between $V^{(k),Y_k}_1$ (softmax of $Q^{(k),Y_k}_1$) and
$\bar V^{(k),Y_k}_1 := \max_a Q^{(k),Y_k}_1(s_1,\cdot)$.

\textbf{Step 2: Optimism from $\pi^\star$ on $P^\star$ to $\bar V^\star_1$ on
$\hat P+\hat b^{(k)}$.}
By Lemma~\ref{lem:optimism} (composite optimism), promoted to the $V$-function
level via the simulation lemma at $h=1$ (as derived in Step~3 of the proof of
Lemma~\ref{lem:policygap}, equations~\eqref{eq:proof-policygap}--\eqref{eq:proof-policygap-decomp},
applied with $\pi=\pi^\star$ in place of $\pi^k$),
\[
V^{\pi^\star,Y_k}_{P^\star,\bar r^{(k)}}(s_1) \;\le\; V^{\pi^\star,Y_k}_{\hat P,\,\bar r^{(k)}+\hat b^{(k)}}(s_1) + (1+Y_k)\sqrt{|\mathcal A|\zeta_k} + \rho^{\mathrm{opt}}_k.
\]
Since $\bar V^\star_1(s_1) = \max_a \bar Q^\star_1(s_1,a) \ge \sum_a \pi^\star_1(a|s_1)\bar Q^\star_1(s_1,a) = V^{\pi^\star,Y_k}_{\hat P,\,\bar r^{(k)}+\hat b^{(k)}}(s_1)$,
combining gives
\begin{equation}\label{eq:opt-to-barV}
V^{\pi^\star,Y_k}_{P^\star,\bar r^{(k)}}(s_1) \;\le\; \bar V^\star_1(s_1) + (1+Y_k)\sqrt{|\mathcal A|\zeta_k} + \rho^{\mathrm{opt}}_k.
\end{equation}

\textbf{Step 3: Bridge $\bar V^\star_1$ to $V^{(k),Y_k}_1$.}
Combining~\eqref{eq:opt-to-barV} with~\eqref{eq:lsvi-bridge},
\[
V^{\pi^\star,Y_k}_{P^\star,\bar r^{(k)}}(s_1) - V^{(k),Y_k}_1(s_1)
\;\le\; (1+Y_k)\sqrt{|\mathcal A|\zeta_k} + \rho^{\mathrm{opt}}_k + \mathrm{lsvi}^{(k)} + \frac{H\log|\mathcal A|}{\alpha}.
\]

\textbf{Step 4: Summing over $k\in[K]$.}
Using $1+Y_k\le 1+\xi$ and the cumulative bounds
$\sum_k(1+Y_k)\sqrt{|\mathcal A|\zeta_k} \le (1+\xi)\sum_k\sqrt{|\mathcal A|\zeta_k}$,
$\sum_k\rho^{\mathrm{opt}}_k \le (1+\xi)\sum_k\epsilon_k$ (Lemma~\ref{lem:optimism}),
and $\sum_k\mathrm{lsvi}^{(k)} = \mathrm{lsvi}_K \le \tfrac{1}{4}(1+\xi)\sum_k\epsilon_k$
(Lemma~\ref{lem:lsvi-prop}), the first three residual terms are absorbed into
$(1+\xi)\sum_k\epsilon_k$ via the same elliptical-potential argument used in
Lemma~\ref{lem:policygap} (the $\sqrt{|\mathcal A|\zeta_k}$ contribution is
strictly smaller than $\rho^{\mathrm{opt}}_k$ and is dominated). Therefore
\[
\sum_{k=1}^K\!\big(V^{\pi^\star,Y_k}_{P^\star,1}(s_1) - V^{(k),Y_k}_1(s_1)\big)
\;\le\; (1+\xi)\sum_{k=1}^K\epsilon_k + \frac{KH\log|\mathcal A|}{\alpha},
\]
which is the claimed bound. The argument uses (i) composite optimism
(Lemma~\ref{lem:optimism}) and $\max\ge\text{average}$ to bridge
$V^{\pi^\star}_{P^\star,\bar r}$ to $\bar V^\star$ on $(\hat P, \bar r+\hat b^{(k)})$;
(ii) LSVI-UCB self-normalized concentration (Lemma~\ref{lem:concentration}) to
bridge $\bar V^\star$ to the LSVI iterate $V^{(k),Y_k}_1$, with the LSVI residual
quantified in Lemma~\ref{lem:lsvi-prop}; and (iii) the simulation lemma
(Lemma~\ref{lem:aux-simulation}) and Representation Learning Guarantee
(Lemma~\ref{lem:aux-replearn}) to control kernel mismatch between $\hat P$ and
$P^\star$. It does \emph{not} require any Bellman-optimality property of
$\pi^\star$ for the composite objective, which would be incorrect since
$\pi^\star$ is the optimal policy for the \emph{constrained} problem and is not
in general greedy with respect to $r+Y_k g$ for $Y_k\ne Y^\star$.
\end{proof}
\section{Bounding Estimation Error}
\label{app:estimation}

\begin{definition}\label{def:bellman}
For $j \in \{r, g\}$, write $c_{j,h}$ for the per-objective stage cost
($c_{r,h} := r_h$, $c_{g,h} := g_h$), and denote the expected value of the next-state
value function under policy $\pi$ as
\[
\mathbb P^\star_h V^\pi_{j,h+1}(s, a) := \mathbb{E}_{s' \sim P^\star_h(\cdot \mid s, a)}\!\left[V^\pi_{j,h+1}(s')\right].
\]
With this shorthand, the Bellman equation for policy $\pi$ reads
$Q^\pi_{j,h}(s, a) = c_{j,h}(s,a) + \mathbb P^\star_h V^\pi_{j,h+1}(s, a)$,
and the value function satisfies
$V^\pi_{j,h}(s) = \sum_{a \in \mathcal A} \pi_h(a \mid s)\, Q^\pi_{j,h}(s, a)$.
\end{definition}

\begin{definition}\label{def:martingale}
For any step $h$ and episode $k$, define
\begin{align*}
D_{j,h,1}^k &= \big\langle Q_{j,h}^k(s_h^k, \cdot) - Q_{j,h}^{\pi^k}(s_h^k, \cdot),\, \pi_h^k(\cdot \mid s_h^k) \big\rangle - \big( Q_{j,h}^k(s_h^k, a_h^k) - Q_{j,h}^{\pi^k}(s_h^k, a_h^k) \big),\\
D_{j,h,2}^k &= \mathbb{P}^\star_h\!\big( V_{j,h+1}^k - V_{j,h+1}^{\pi^k} \big)(s_h^k, a_h^k) - \big( V_{j,h+1}^k - V_{j,h+1}^{\pi^k} \big)(s_{h+1}^k).
\end{align*}
By construction, $D^k_{j,h,1}$ has zero mean conditioned on $s^k_h$, and $D^k_{j,h,2}$
has zero mean conditioned on $(s^k_h, a^k_h)$. The actions $a^k_h$ here are the
on-policy actions sampled in line~10 of Algorithm~\ref{alg:mfrlc}, independent of
the uniform-action data $\tilde a^k_h$ used to build $\Sigma_h^k$ (line~4).
\end{definition}

\medskip

\noindent\textbf{Lemma~\ref{lem:concentration}} \textit{(Restated; concentration of
empirical Bellman residual).}
\textit{There exists a constant $C_2$ such that for any fixed $p\in(0,1)$, the event
$\mathcal E$ defined as}
\begin{align*}
\mathcal E := \bigcap_{k\in[K],\,h\in[H],\,j\in\{r,g\}}\!\Bigg\{
\Big\|\sum_{\tau=1}^{k-1}\hat\phi_h^\tau\!\big[V_{j,h+1}^k(s_{h+1}^\tau)-\mathbb{P}^\star_h V_{j,h+1}^k(s_h^\tau,\tilde a_h^\tau)\big]\Big\|_{(\Sigma_h^k)^{-1}} \leq C_2\,dH\sqrt{\chi}
\Bigg\}
\end{align*}
\textit{holds with $\chi = \log\!\big(4(C_1+1)\log|\mathcal{A}|\,dT|\Phi|/p\big)$, and
$\Pr(\mathcal{E})\geq 1-p/4$.}

\begin{proof}
The challenge is that both $\hat\phi^\tau_h$ (which is the current-episode feature $\hat\phi^k_h$ evaluated at past data $(s_h^\tau,\tilde a_h^\tau)$) and $V^k_{j,h+1}$ are data-dependent: neither is $\mathcal F_{\tau-1}$-measurable for $\tau<k$, so a direct application of self-normalized concentration to the adaptive sequence $(\hat\phi^\tau_h)_\tau$ does not apply. We address this by uniformly covering the entire optimistic value-function class together with the feature class, then applying the fixed-feature concentration of \citet[Lemma~B.3]{ghosh2022provably} to each element of the cover.

\textbf{Optimistic value class.} By construction (Algorithm~\ref{alg:lsvi}), each
$V^k_{j,h+1}$ takes the form
\[
V_{j,h+1}^k(s) \;=\; \sum_a \pi^k_{h+1}(a\mid s)\min\!\big\{\langle\hat\phi^k_{h+1}(s,a),w\rangle + \tilde\beta\sqrt{\hat\phi^k_{h+1}(s,a)^\top\Lambda^{-1}\hat\phi^k_{h+1}(s,a)},\,H\big\},
\]
parameterized by $\hat\phi^k_{h+1}\in\Phi_{h+1}$, a weight $w\in\mathbb R^d$ with
$\|w\|_2\le 2H\sqrt d$, a regularization-Gram matrix $\Lambda\succeq I$ with
$\|\Lambda\|\le \lambda+K$, a bonus parameter $\tilde\beta\in[0,\beta]$, and a
softmax temperature dependent on $Y_k\in[0,\xi]$. Let $\mathcal V_{h+1}$ denote the
class of all such functions over the bounded parameter space.

\textbf{Covering.} For $\varepsilon=1/(KH)$, take an $\varepsilon$-cover of:
$\Phi_{h+1}$ (size $|\Phi|$ when $\Phi$ is finite, or $\mathcal N_\varepsilon(\Phi)$
in general); the $\ell_2$-ball $\{w:\|w\|_2\le 2H\sqrt d\}$ (size
$(8H\sqrt d/\varepsilon)^d$); the matrix ball $\{\Lambda:\,I\preceq\Lambda,\,\|\Lambda\|\le\lambda+K\}$
(size $(\lambda+K)^{d^2}/\varepsilon^{d^2}$); the bonus interval $[0,\beta]$ (size
$\beta/\varepsilon$); and $[0,\xi]$ (size $\xi/\varepsilon$). The product gives a
cover $\mathcal V_{h+1,\varepsilon}$ of $\mathcal V_{h+1}$ with
\[
\log|\mathcal V_{h+1,\varepsilon}| \;\le\; \log|\Phi| + (d^2+d+2)\log\!\big((K+\beta+\xi+\lambda)/\varepsilon\big) \;\le\; \log|\Phi| + O(d^2\log(KH d)).
\]
\textbf{Fixed-feature concentration on the cover.} For each fixed $V\in\mathcal V_{h+1,\varepsilon}$ \emph{and} each fixed $\phi\in\Phi$ in the cover, the self-normalized concentration of \citet[Lemma~B.3]{ghosh2022provably} applies to the sequence $(\phi(s_h^\tau,\tilde a_h^\tau))_{\tau\le k-1}$, since $\phi$ is fixed (not data-dependent) and the historical data $(s_h^\tau,\tilde a_h^\tau,s_{h+1}^\tau)_{\tau\le k-1}$ generates a standard adaptive sequence with respect to its natural filtration. This gives the stated inequality with probability at least $1-p_0$, where $p_0 = p/(4|\mathcal V_{h+1,\varepsilon}||\Phi|HK\cdot 2)$.

\textbf{Discretization error.} For any realized $(\hat\phi^k_h, V^k_{j,h+1})$, let $(\bar\phi,\bar V)$ be the nearest neighbor in $\Phi\times\mathcal V_{h+1,\varepsilon}$ (so $\|\hat\phi^k_h-\bar\phi\|_\infty\le\varepsilon$ and $\|V^k_{j,h+1}-\bar V\|_\infty\le\varepsilon$). The combined approximation error contributes at most $K\varepsilon\cdot 2\sqrt d = 2\sqrt d/H$ in the self-normalized norm, which is dominated by the main concentration term.

\textbf{Union bound.} Taking a union bound over $V\in\mathcal V_{h+1,\varepsilon}$, $\phi\in\Phi$, $h\in[H]$, $j\in\{r,g\}$, and $k\in[K]$, the event $\mathcal E$ holds with probability at least $1-p/4$. Substituting $p_0$ and absorbing constants into $C_2$, the threshold becomes
$\chi = \log\!\big(4(C_1+1)\log|\mathcal A|\,dT|\Phi|/p\big) + O(d^2\log(KHd))$,
which we absorb as the polylog factor $\log(\cdot)$ in the statement.
\end{proof}

\medskip
\noindent\textbf{Lemma~\ref{lem:backup}} \textit{(Restated; one-step $Q$-backup error).}
\textit{Let $\beta=C_1\,dH\sqrt{\iota}$ with
$\iota=\log^4\!\big(\log|\mathcal{A}|\,dT|\Phi|/p\big)$. Conditioned on the event
$\mathcal E$ (Lemma~\ref{lem:concentration}), for every episode $k\in[K]$, step
$h\in[H]$, $j\in\{r,g\}$, and $(s,a)\in\mathcal{S}\times\mathcal{A}$, there exist
weights $w_{j,h}^{\pi^k}\in\mathbb{R}^d$ approximately projecting $Q^{\pi^k}_{j,h}$
onto $\mathrm{span}(\hat\phi^k_h)$ such that}
\[
Q_{j,h}^k(s,a)-Q_{j,h}^{\pi^k}(s,a)
\;=\; \mathbb{P}^\star_h(V_{j,h+1}^k-V_{j,h+1}^{\pi^k})(s,a)+\Delta_h^k(s,a)+\eta_h^k(s,a),
\]
\textit{where the per-episode bonus residual satisfies
$|\Delta_h^k(s,a)|\leq\beta\sqrt{\hat\phi_h^k(s,a)^\top(\Sigma_h^k)^{-1}\hat\phi_h^k(s,a)}$,
and the projection residual $\eta_h^k$ satisfies
$\mathbb E_{(s,\tilde a)\sim\rho^k_h}[(\eta_h^k(s,\tilde a))^2]\le 4H^2\zeta_k$ with cumulative bound
\[
\sum_{k=1}^K\sum_{h=1}^H \big|\mathbb E_{d^{\pi^k}_{P^\star,h}}[\eta_h^k]\big| \;\le\; \widetilde O\!\big(H^2 d K^{3/4}\sqrt{|\mathcal A|}\big),
\]
which is dominated by $\sum_k\epsilon_k$ from Lemma~\ref{lem:policygap} and does not
affect the leading rate.}

\begin{proof}
\textbf{Existence of $w_{j,h}^{\pi^k}$.} By the Representation Learning Guarantee
(Lemma~\ref{lem:aux-replearn}), $Q^{\pi^k}_{j,h}$---being the Bellman backup of
a function in $\mathcal V_h$---is approximated within $\zeta_k$ by some linear
function of $\hat\phi^k_h$ on the data distribution $\rho^k_h$. By
Lemma~\ref{lem:aux-lowrank} there exists $w_{j,h}^{\pi^k}\in\mathbb{R}^d$ with
$\|w_{j,h}^{\pi^k}\|_2\le 2H\sqrt{d}$ such that
\begin{equation}\label{eq:proj-Q}
Q^{\pi^k}_{j,h}(s,a) \;=\; \langle\hat\phi^k_h(s,a),\, w_{j,h}^{\pi^k}\rangle + \xi_h^{\pi^k}(s,a),
\qquad \mathbb E_{(s,\tilde a)\sim\rho^k_h}[(\xi_h^{\pi^k})^2]\le 4H^2\zeta_k.
\end{equation}
Crucially, $\xi_h^{\pi^k}(s,a)$ is a deterministic representation-approximation residual: it depends on $(s,a)$, $\hat\phi^k_h$, and $w_{j,h}^{\pi^k}$, but is not a martingale increment. We will track it explicitly throughout the derivation and absorb it into $\eta^k_h$ at the end.

\textbf{Linear $Q$-function representation.} Let
$w_{j,h}^k = (\Sigma_h^k)^{-1} \sum_{\tau=1}^{k-1} \hat\phi_h^\tau\!\left[c_{j,h}^\tau + V_{j,h+1}^k(s_{h+1}^\tau)\right]$
be the LSVI weights (Algorithm~\ref{alg:lsvi}, lines 4--5). Then
\[
Q_{j,h}^k(s,a)-Q_{j,h}^{\pi^k}(s,a) \;=\; \langle\hat\phi_h^k(s,a),\,w_{j,h}^k-w_{j,h}^{\pi^k}\rangle - \xi_h^{\pi^k}(s,a).
\]

To expand $w_{j,h}^k-w_{j,h}^{\pi^k}$, we use the Bellman equation for $Q^{\pi^k}_{j,h}$ on $P^\star$ together with~\eqref{eq:proj-Q}: for each historical sample $\tau<k$,
\[
c_{j,h}^\tau + \mathbb P^\star_h V^{\pi^k}_{j,h+1}(s_h^\tau,\tilde a_h^\tau) \;=\; \langle\hat\phi^\tau_h, w_{j,h}^{\pi^k}\rangle + \xi_h^{\pi^k}(s_h^\tau,\tilde a_h^\tau).
\]
Combining with the definition of $\Sigma_h^k$,
\begin{equation}\label{eq:weight-gap}
\begin{aligned}
w_{j,h}^{k}-w_{j,h}^{\pi^k}
&=\underbrace{-\lambda(\Sigma_h^k)^{-1}w_{j,h}^{\pi^k}}_{\mathbf{q}_1}
+\underbrace{(\Sigma_h^k)^{-1}\!\!\sum_{\tau=1}^{k-1}\hat\phi_h^\tau\!\big[V_{j,h+1}^k(s_{h+1}^\tau)-\mathbb{P}^\star_hV_{j,h+1}^k(s_h^\tau,\tilde a_h^\tau)\big]}_{\mathbf{q}_2} \\
&\quad+\underbrace{(\Sigma_h^k)^{-1}\!\!\sum_{\tau=1}^{k-1}\hat\phi_h^\tau\!\big[\mathbb{P}^\star_hV_{j,h+1}^k(s_h^\tau,\tilde a_h^\tau)-\mathbb{P}^\star_hV_{j,h+1}^{\pi^k}(s_h^\tau,\tilde a_h^\tau)\big]}_{\mathbf{q}_3}
\\
&\quad+\underbrace{(\Sigma_h^k)^{-1}\!\!\sum_{\tau=1}^{k-1}\hat\phi_h^\tau\,\xi_h^{\pi^k}(s_h^\tau,\tilde a_h^\tau)}_{\mathbf{q}_4}.
\end{aligned}
\end{equation}
Compared to the standard linear-MDP analysis~\citep{ghosh2022provably}, the additional term $\mathbf q_4$ collects the deterministic representation-approximation residuals $\xi_h^{\pi^k}$ along the trajectory; we bound this term separately below as it is not a mean-zero martingale.

\textbf{Term $\mathbf q_1$.} Using $\|w_{j,h}^{\pi^k}\|_2\leq 2H\sqrt d$:
\[
|\langle\hat\phi_h^k(s,a),\mathbf{q}_1\rangle| \le 2H\sqrt{d\lambda}\,\sqrt{\hat\phi_h^k(s,a)^\top(\Sigma_h^k)^{-1}\hat\phi_h^k(s,a)}.
\]

\textbf{Term $\mathbf q_2$.} On event $\mathcal E$ (Lemma~\ref{lem:concentration}):
\[
|\langle\hat\phi_h^k(s,a),\mathbf{q}_2\rangle| \leq C_2\,dH\sqrt{\chi}\,\sqrt{\hat\phi_h^k(s,a)^\top(\Sigma_h^k)^{-1}\hat\phi_h^k(s,a)}.
\]

\textbf{Term $\mathbf q_3$.} By the low-rank decomposition of $P^\star$
(Definition~\ref{def:lowrank}), for any $V:\mathcal S\to[0,2H]$ there exists
$w_V\in\mathbb R^d$ with $\|w_V\|_2\le 2H\sqrt d$ such that
$\mathbb P^\star_h V(s,a)=\langle\phi^\star_h(s,a),w_V\rangle$. By the
Representation Learning guarantee, $\hat\phi^k_h$ approximates $\phi^\star_h$ on
$\rho^k_h$ up to $\zeta_k$; hence there exists $\hat w_V$ with
$\|\hat w_V\|_2\le 2H\sqrt d$ such that
\[
\mathbb P^\star_h V(s,a) \;=\; \langle\hat\phi^k_h(s,a),\hat w_V\rangle + \tilde\eta_h(s,a;V),
\qquad \mathbb E_{(s,\tilde a)\sim\rho^k_h}[\tilde\eta_h^2]\le 4H^2\zeta_k.
\]
Applying this with $V=V^k_{j,h+1}-V^{\pi^k}_{j,h+1}\in[0,2H]$ and substituting
into $\mathbf q_3$,
\begin{align*}
\langle\hat\phi_h^k(s,a),\mathbf q_3\rangle
&= \mathbb P^\star_h(V^k_{j,h+1}-V^{\pi^k}_{j,h+1})(s,a) - \tilde\eta_h(s,a;V^k_{j,h+1}-V^{\pi^k}_{j,h+1})\\
&\quad - \langle\hat\phi^k_h(s,a),\lambda(\Sigma^k_h)^{-1}\hat w_V\rangle.
\end{align*}
The regularization term is bounded by
$2H\sqrt{d\lambda}\|\hat\phi^k_h(s,a)\|_{(\Sigma^k_h)^{-1}}$.

\textbf{Term $\mathbf q_4$ (representation-residual aggregation).} The summation $\sum_{\tau<k}\hat\phi_h^\tau\,\xi_h^{\pi^k}(s_h^\tau,\tilde a_h^\tau)$ is not a mean-zero martingale, so we bound it directly by Cauchy--Schwarz in the $\Sigma^k_h$-induced inner product. Since for any vectors $\{u_\tau\}_{\tau<k}$ and scalars $\{c_\tau\}_{\tau<k}$,
\[
\big\|(\Sigma^k_h)^{-1}\sum_{\tau<k}u_\tau c_\tau\big\|_{\Sigma^k_h}^2 \;=\; \big\langle\sum_{\tau<k}u_\tau c_\tau,\,(\Sigma^k_h)^{-1}\sum_{\tau<k}u_\tau c_\tau\big\rangle \;\le\; \big(\sum_{\tau<k}c_\tau^2\big)\cdot\sum_{\tau<k}\|u_\tau\|^2_{(\Sigma^k_h)^{-1}},
\]
applying this with $u_\tau = \hat\phi_h^\tau$ and $c_\tau = \xi_h^{\pi^k}(s_h^\tau,\tilde a_h^\tau)$, then converting via $\hat\phi_h^k(s,a)^\top(\Sigma^k_h)^{-1}\!\cdot$:
\[
|\langle\hat\phi_h^k(s,a),\mathbf q_4\rangle| \;\le\; \sqrt{\hat\phi_h^k(s,a)^\top(\Sigma_h^k)^{-1}\hat\phi_h^k(s,a)}\cdot\sqrt{\sum_{\tau<k}\xi_h^{\pi^k}(s_h^\tau,\tilde a_h^\tau)^2\cdot\sum_{\tau<k}\|\hat\phi_h^\tau\|^2_{(\Sigma^k_h)^{-1}}}.
\]
Using the elliptical-potential trace bound $\sum_{\tau<k}\|\hat\phi_h^\tau\|^2_{(\Sigma^k_h)^{-1}}\le d$ and $\sum_{\tau<k}\xi_h^{\pi^k}(s_h^\tau,\tilde a_h^\tau)^2\le 4kH^2\zeta_k$ (with high probability over the empirical realization of $\xi^{\pi^k}_h$ along the trajectory; here we use that the data points $(s_h^\tau,\tilde a_h^\tau)$ are drawn under $\rho^k_h$ where the second-moment bound~\eqref{eq:proj-Q} holds), we obtain
\[
|\langle\hat\phi_h^k(s,a),\mathbf q_4\rangle| \;\le\; 2H\sqrt{dk\zeta_k}\cdot\sqrt{\hat\phi_h^k(s,a)^\top(\Sigma_h^k)^{-1}\hat\phi_h^k(s,a)}.
\]
With $\zeta_k = \widetilde O(d^2/\sqrt k)$, the prefactor evaluates to $\widetilde O(Hd^{3/2}k^{1/4})$, which is dominated by the bonus term $\beta = C_1 dH\sqrt\iota = \widetilde O(dH)$ in the cumulative sum after the $K^{3/4}$ averaging on the right-hand side. We therefore absorb $\mathbf q_4$ into the $\Delta^k_h$ bonus residual below by adjusting the constant $C_3$.

\textbf{Combining $\mathbf q_1, \mathbf q_2, \mathbf q_3, \mathbf q_4$ and the projection residual $\xi^{\pi^k}_h$.} Setting
\begin{gather*}
\Delta^k_h(s,a) \;:=\; \langle\hat\phi^k_h(s,a),\mathbf q_1+\mathbf q_2+\mathbf q_4\rangle
- \langle\hat\phi^k_h(s,a),\lambda(\Sigma^k_h)^{-1}\hat w_V\rangle,\\
\eta^k_h(s,a) \;:=\; -\tilde\eta_h(s,a;\cdot) - \xi^{\pi^k}_h(s,a),
\end{gather*}
we obtain
\[
Q^k_{j,h}(s,a) - Q^{\pi^k}_{j,h}(s,a) \;=\; \mathbb P^\star_h(V^k_{j,h+1}-V^{\pi^k}_{j,h+1})(s,a) + \Delta^k_h(s,a) + \eta^k_h(s,a),
\]
with $|\Delta^k_h(s,a)|\le C_3\,dH\sqrt{\chi}\,\|\hat\phi^k_h(s,a)\|_{(\Sigma^k_h)^{-1}}$
where $C_3 = O(\sqrt\lambda + C_2 + \sqrt{dk\zeta_k}/(dH))$ (the third summand absorbs $\mathbf q_4$), and
$\mathbb E_{\rho^k_h}[(\eta^k_h)^2]\le 8H^2\zeta_k$ (combining the two squared
residuals).

\textbf{Cumulative residual bound.} By Cauchy--Schwarz and importance sampling
from $\rho^k_h$ (uniform-action) to $d^{\pi^k}_{P^\star,h}$ (importance ratio
$\le|\mathcal A|$),
\[
\big|\mathbb E_{d^{\pi^k}_{P^\star,h}}[\eta^k_h]\big| \le \sqrt{|\mathcal A|\,\mathbb E_{\rho^k_h}[(\eta^k_h)^2]} \le 2\sqrt{2H^2|\mathcal A|\zeta_k}.
\]
Summing over $k,h$ and using $\zeta_k = \widetilde O(d^2/\sqrt k)$,
\[
\sum_{k=1}^K\sum_{h=1}^H \big|\mathbb E_{d^{\pi^k}_{P^\star,h}}[\eta^k_h]\big| \;\le\; \widetilde O\!\big(H^2 d K^{3/4}\sqrt{|\mathcal A|}\big),
\]
which is dominated by $\sum_k\epsilon_k = \widetilde O(H^2 d^2 K^{3/4}|\mathcal A|^{3/2})$
from Lemma~\ref{lem:policygap} and does not change the leading rate.

\textbf{Choice of $C_1$.} Comparing $\chi=\iota+\log(C_1+1)$ to $\iota$, we need
$C_3\sqrt\chi\le C_1\sqrt\iota$. Choosing $C_1$ proportional to $C_3$ works whenever
$\iota\ge\log(2C_3+1)$, which holds for our parameter ranges. Therefore the
$\Delta^k_h$ bound holds with $\beta=C_1 dH\sqrt\iota$.
\end{proof}

\begin{lemma}\label{lem:proof-unroll}
Conditioned on the event $\mathcal{E}$, for every $k\in[K]$ and $j\in\{r,g\}$,
\begin{align*}
V_{j,1}^k(s_1)-V_{j,1}^{\pi^k}(s_1) &\leq \sum_{h=1}^{H}\!\big(D_{j,h,1}^{k}+D_{j,h,2}^{k}\big)\\
&\quad+\sum_{h=1}^H \beta\sqrt{\hat\phi_h^k(s_h^k,a_h^k)^\top(\Sigma_h^k)^{-1}\hat\phi_h^k(s_h^k,a_h^k)} + \sum_{h=1}^H \eta^k_h(s^k_h,a^k_h).
\end{align*}
\end{lemma}

\begin{proof}
Apply Lemma~\ref{lem:backup} with $\pi=\pi^k$. Combined with the V--Q identity
$V^k_{j,h}(s) = \sum_a \pi^k_h(a|s) Q^k_{j,h}(s,a)$ and the same identity for
$V^{\pi^k}_{j,h}$, plus Definition~\ref{def:martingale}, recursion from $h=1$ to $H$
gives the result with the additional residual $\sum_h\eta^k_h$ term.
\end{proof}

\begin{lemma}\label{lem:proof-azuma}
Fix $p\in(0,1)$. With probability at least $1-p/4$, for any $j\in\{r,g\}$,
\[
\Big|\sum_{k=1}^K\sum_{h=1}^H\!\big(D_{j,h,1}^k+D_{j,h,2}^k\big)\Big| \;\leq\; \sqrt{16TH^2\log(8/p)}.
\]
\end{lemma}

\begin{proof}
Standard Azuma--Hoeffding on the martingale $M^k_{j,h,m}=\sum D^\tau_{j,i,l}$ in
the natural filtration; details omitted.
\end{proof}

\noindent\textbf{Lemma~\ref{lem:estimation}} \textit{(Restated; estimation error).}
\textit{With probability at least $1-p/2$,}
\begin{align*}
&\sum_{k=1}^K\!\big(V_{r,1}^k(s_1)-V_{r,1}^{\pi^k}(s_1)\big)+Y\sum_{k=1}^K\!\big(V_{g,1}^k(s_1)-V_{g,1}^{\pi^k}(s_1)\big)\\
&\qquad\leq O\!\big((Y+1)\sqrt{d^3 H^3 T \iota^2}\big) + (1+\xi)\sum_{k=1}^K\epsilon_k,
\end{align*}
\textit{where $\iota = \log^4\!\big(\log|\mathcal{A}|\,dT|\Phi|/p\big)$ and $T=KH$.}

\begin{proof}
By Lemma~\ref{lem:proof-unroll},
\begin{equation}\label{eq:est-decomp}
\begin{aligned}
\sum_{k=1}^{K}\!\big(V_{j,1}^{k}(s_{1}) - V_{j,1}^{\pi^k}(s_{1})\big) \;\leq\;& \underbrace{\sum_{k,h}\!\big(D_{j,h,1}^{k}+D_{j,h,2}^{k}\big)}_{\text{(I)}}\\
&+ \underbrace{\sum_{k,h} \beta\sqrt{\hat\phi_h^k(s_h^k,a_h^k)^\top(\Sigma_h^k)^{-1}\hat\phi_h^k(s_h^k,a_h^k)}}_{\text{(II)}} + \underbrace{\sum_{k,h} \eta^k_h(s^k_h,a^k_h)}_{\text{(III)}}.
\end{aligned}
\end{equation}

\textbf{Bounding (I).} By Lemma~\ref{lem:proof-azuma}, with probability $\geq 1-p/4$,
$|(\mathrm I)|\leq\sqrt{16TH^2\log(8/p)}$.

\textbf{Bounding (II).} By the elliptical potential lemma
(Lemma~\ref{lem:aux-elliptical}),
$\sum_k\hat\phi^k_h(s^k_h,a^k_h)^\top(\Sigma^k_h)^{-1}\hat\phi^k_h(s^k_h,a^k_h)\le 2d\iota$.
Cauchy--Schwarz gives
$\sum_{k,h}\sqrt{\cdots}\le H\sqrt{2dK\iota}$, so with $\beta=C_1 dH\sqrt\iota$:
\[
(\mathrm{II}) \;\le\; \beta H\sqrt{2dK\iota} \;=\; O\!\big(\sqrt{d^3 H^3 T\iota^2}\big).
\]

\textbf{Bounding (III).} By Lemma~\ref{lem:backup}'s cumulative residual bound,
\[
\sum_{k,h}|\mathbb E[\eta^k_h]| \;=\; \widetilde O\!\big(H^2 d K^{3/4}\sqrt{|\mathcal A|}\big),
\]
plus a martingale concentration on $\sum_{k,h}(\eta^k_h - \mathbb E[\eta^k_h])$
giving an additional $\widetilde O(d H^{5/2} K^{1/4})$ term. The dominant of these
is $\widetilde O(H^2 d K^{3/4}\sqrt{|\mathcal A|})$, which is of the same order as
the cumulative RepLearn error $\sum_k\epsilon_k$ in Lemma~\ref{lem:policygap} (and
strictly smaller, since $\sum_k\epsilon_k = \widetilde O(H^2 d^2 K^{3/4}|\mathcal A|^{3/2})$
has larger $d$ and $|\mathcal A|$ exponents). We therefore retain $(\mathrm{III})$ explicitly as a $(1+\xi)\sum_k\epsilon_k$ term in the statement, rather than silently absorbing it.

\textbf{Combining.} With probability at least $1-p/2$ (union bound over
Lemma~\ref{lem:concentration} and Lemma~\ref{lem:proof-azuma}, each with
probability $\ge 1-p/4$), applying the bounds to both $j=r,g$,
\[
\sum_k\!\big(V_{r,1}^k(s_1)-V_{r,1}^{\pi^k}(s_1)\big) + Y\!\sum_k\!\big(V_{g,1}^k(s_1)-V_{g,1}^{\pi^k}(s_1)\big) \;\le\; O\!\big((Y+1)\sqrt{d^3H^3T\iota^2}\big) + (1+\xi)\sum_{k=1}^K\epsilon_k. \qedhere
\]
\end{proof}
\section{Bounding the Regret and Violation}
\label{app:violation}

Combining the inequalities from Lemma~\ref{lem:proof-decomp},
Lemma~\ref{lem:pseudoregret}, and Lemma~\ref{lem:estimation}:
\begin{align*}
&\sum_{k=1}^{K}\!\big(V_{r,1}^{\pi^\star}(s_1) - V_{r,1}^{\pi^k}(s_1)\big) + Y\sum_{k=1}^{K}\!\big(b - V_{g,1}^{\pi^k}(s_1)\big) \\
&\quad\leq \frac{Y^2}{2\eta} + \frac{\eta H^2 K}{2} + \frac{KH\log|\mathcal{A}|}{\alpha}
+ 2(1+\xi)\sum_{k=1}^K \epsilon_k 
+ O\!\big((Y+1)\sqrt{d^3 H^3 T \iota^2}\big),
\end{align*}
where the cumulative representation error satisfies
$\sum_{k=1}^K \epsilon_k = \widetilde O\!\big(H^2 d^{2} K^{3/4} |\mathcal{A}|^{3/2}\log^{1/4}(dK|\Phi|/\delta)\big)$
(Lemma~\ref{lem:policygap}). Throughout this section we work with
$\xi=2H/\gamma$.

\paragraph{Regret bound.}
Setting $Y = 0$ with $\eta = \xi/\sqrt{KH^2}$ and
$\alpha = \frac{K\log|\mathcal A|}{2(1 + \xi + H)}$, we obtain
\begin{equation}\label{eq:regret-master}
\sum_{k=1}^{K}\!\big(V_{r,1}^{\pi^\star}(s_1) - V_{r,1}^{\pi^k}(s_1)\big) 
\leq \frac{\eta H^2 K}{2} + \frac{KH\log|\mathcal{A}|}{\alpha}
+ 2(1+\xi)\sum_{k=1}^K \epsilon_k
+ O\!\big(\sqrt{d^3 H^3 T \iota^2}\big).
\end{equation}
With these parameter choices we evaluate each term:
\begin{itemize}
\item $\frac{\eta H^2 K}{2} = \frac{\xi H \sqrt K}{2} = \frac{H^2\sqrt{K}}{\gamma}$
      (subdominant in $K$ relative to $K^{3/4}$ when multiplied by the other
      $H,d,|\mathcal A|$ factors).
\item $\frac{KH\log|\mathcal A|}{\alpha} = 2H(1+\xi+H)\log|\mathcal A|
      = O\!\left(\frac{H^2}{\gamma}\right)\log|\mathcal A|$ (constant in $K$).
\item $O(\sqrt{d^3 H^3 T \iota^2}) = O(d^{3/2}H^2\sqrt K\,\iota)$
      (also subdominant to $K^{3/4}$).
\item $2(1+\xi)\sum_k\epsilon_k = 2\left(1+\frac{2H}{\gamma}\right)\cdot
      \widetilde O(H^2 d^2 K^{3/4}|\mathcal A|^{3/2})
      = \widetilde O\!\left(\frac{H^3 d^2 K^{3/4}|\mathcal A|^{3/2}}{\gamma}\right)$.
\end{itemize}
The last term dominates. Therefore,
\[
\mathrm{Regret}(K) \leq \widetilde O\!\left(\frac{H^3 d^{2} K^{3/4} |\mathcal{A}|^{3/2}}{\gamma}\,\log^{1/4}\!\left(\frac{dK|\Phi|}{p}\right)\right),
\]
matching the regret statement of Theorem~\ref{thm:mfrlc}.

\paragraph{Violation bound.}
Substituting $\eta = \xi/\sqrt{KH^2}$ and
$\alpha = \frac{K\log|\mathcal A|}{2(1+\xi+H)}$ into the master inequality and
setting $Y=\xi$:
\begin{equation}\label{eq:master-vio}
\sum_{k=1}^{K}\!\big(V_{r,1}^{\pi^\star}(s_1) - V_{r,1}^{\pi^k}(s_1)\big) 
+ \xi\sum_{k=1}^{K}\!\big(b - V_{g,1}^{\pi^k}(s_1)\big) 
\leq (1 + \xi)\,O\!\big(\sqrt{d^3 H^3 T \iota^2}\big) + 2(1+\xi)\sum_{k=1}^K \epsilon_k.
\end{equation}

\textbf{Step 1: Averaged policy.}
Define the mixture policy $\pi'$ via the averaged occupancy measure
$d^{\pi'}_h := \tfrac{1}{K}\sum_{k=1}^K d^{\pi^k}_h$. Since the value of any
policy is linear in its occupancy measure (for fixed reward / cost / horizon),
we have the \emph{exact identities}
\begin{equation}\label{eq:avg-policy}
V_{r,1}^{\pi'}(s_1) \;=\; \frac{1}{K}\sum_{k=1}^K V_{r,1}^{\pi^k}(s_1),
\qquad
V_{g,1}^{\pi'}(s_1) \;=\; \frac{1}{K}\sum_{k=1}^K V_{g,1}^{\pi^k}(s_1).
\end{equation}

\begin{remark}[Analysis-only construction]\label{rem:avg-policy}
The mixture policy $\pi'$ is a purely analytical device used in the
strong-duality conversion below; the algorithm itself never computes or executes
$\pi'$. By a standard occupancy-measure argument
\citep[Sec.~6.9]{puterman1994markov}, any convex combination of state-action
occupancies of Markov policies is itself the occupancy of some Markov policy, so
$\pi'$ is well-defined.
\end{remark}

Such a $\pi'$ exists by the linearity of occupancy measures. Dividing
\eqref{eq:master-vio} by $K$ and applying~\eqref{eq:avg-policy}:
\begin{equation}\label{eq:avg-bound}
\big(V_{r,1}^{\pi^\star}(s_1) - V_{r,1}^{\pi'}(s_1)\big) + \xi\big(b - V_{g,1}^{\pi'}(s_1)\big) 
\leq \frac{(1+\xi)\,O(\sqrt{d^3H^3T\iota^2})}{K} + 2(1+\xi)\,\overline\epsilon_K,
\end{equation}
where $\overline\epsilon_K := \frac{1}{K}\sum_{k=1}^K \epsilon_k =
\widetilde O\!\big(H^2 d^{2} K^{-1/4} |\mathcal{A}|^{3/2}\log^{1/4}(dK|\Phi|/\delta)\big)$.

\textbf{Step 2: Strong duality conversion.}
By Lemma~\ref{lem:dualbound}, $\xi = 2H/\gamma \ge 2Y^\star$, so the constant
$C := \xi$ in Lemma~\ref{lem:aux-strongdual} (strong duality conversion of
\citealt{ding2021provably}) satisfies the required condition $C \ge 2Y^\star$.
To apply Lemma~\ref{lem:aux-strongdual}, we need a bound of the form
\[
\big(V_{r,1}^{\pi^\star}(s_1) - V_{r,1}^{\pi'}(s_1)\big) + C\,\big(b - V_{g,1}^{\pi'}(s_1)\big)_+ \;\le\; \delta'.
\]
We obtain this in two cases, depending on the sign of the reward gap.

\emph{Case 1: $V^{\pi^\star}_{r,1}(s_1) \ge V^{\pi'}_{r,1}(s_1)$ (reward gap non-negative).} Replacing the linear cost gap in~\eqref{eq:avg-bound} by its positive part, $(b - V_{g,1}^{\pi'}(s_1))\le (b - V_{g,1}^{\pi'}(s_1))_+$, only enlarges the LHS, so
\[
\big(V_{r,1}^{\pi^\star}(s_1) - V_{r,1}^{\pi'}(s_1)\big) + \xi\,\big(b - V_{g,1}^{\pi'}(s_1)\big)_+ \;\le\; \frac{(1+\xi)\,O(\sqrt{d^3H^3T\iota^2})}{K} + 2(1+\xi)\,\overline\epsilon_K \;=:\; \delta'_{(1)}.
\]

\emph{Case 2: $V^{\pi^\star}_{r,1}(s_1) < V^{\pi'}_{r,1}(s_1)$ (reward gap negative).} The reward gap is non-positive, so
\[
\big(V_{r,1}^{\pi^\star}(s_1) - V_{r,1}^{\pi'}(s_1)\big) + \xi\,\big(b - V_{g,1}^{\pi'}(s_1)\big)_+ \;\le\; \xi\,\big(b - V_{g,1}^{\pi'}(s_1)\big)_+.
\]
To bound the RHS, we apply Lemma~\ref{lem:aux-strongdual} in its weaker contrapositive form: feasibility of $\pi^\star$ at level $b$ together with the regret-form inequality
\[
V_{r,1}^{\pi^\star}(s_1) - V_{r,1}^{\pi'}(s_1) \;\le\; \delta_1, \qquad \delta_1 := \frac{\eta H^2}{2} + \frac{H\log|\mathcal A|}{\alpha} + 2(1+\xi)\overline\epsilon_K + \frac{O(\sqrt{d^3H^3T\iota^2})}{K},
\]
(which holds with $\delta_1$ as a valid upper bound regardless of sign, since~\eqref{eq:regret-master} divided by $K$ gives this for any feasible or infeasible $\pi'$) implies that the cost gap $(b - V_{g,1}^{\pi'}(s_1))_+$ is itself bounded by an $\widetilde O(\delta_1 + \delta_2)$ quantity via the same argument as in Case 1, applied to a slightly perturbed comparator. Concretely, in this case the LHS of the displayed inequality is already non-positive plus a non-negative cost term, so it is bounded above by
\[
\big(V_{r,1}^{\pi^\star}(s_1) - V_{r,1}^{\pi'}(s_1)\big) + \xi\,\big(b - V_{g,1}^{\pi'}(s_1)\big)_+ \;\le\; \xi\,\big(b - V_{g,1}^{\pi'}(s_1)\big)_+ \;\le\; \delta_1 + \delta'_{(1)} \;=:\; \delta'_{(2)},
\]
where the second inequality uses~\eqref{eq:avg-bound} with the cost-gap positive part bound, plus the reward-gap bound $\delta_1$.

In both cases, setting $\delta' := \max\{\delta'_{(1)}, \delta'_{(2)}\} = \delta_1 + \delta'_{(1)}$, we obtain
\begin{equation}\label{eq:avg-pp}
\big(V_{r,1}^{\pi^\star}(s_1) - V_{r,1}^{\pi'}(s_1)\big) + \xi\,\big(b - V_{g,1}^{\pi'}(s_1)\big)_+ \;\le\; \delta'.
\end{equation}
Applying Lemma~\ref{lem:aux-strongdual} to~\eqref{eq:avg-pp} with $C = \xi$:
\begin{equation}\label{eq:avg-violation}
\big(b - V_{g,1}^{\pi'}(s_1)\big)_+ \;\leq\; \frac{2\delta'}{\xi}.
\end{equation}
Evaluating $\delta'$ under the parameter choices: $\delta_1 = \widetilde O\!\left(\frac{H^3 d^2 K^{-1/4}|\mathcal A|^{3/2}}{\gamma}\right)$ (dominated by the $2(1+\xi)\overline\epsilon_K$ term, which scales as $\frac{H}{\gamma}\cdot H^2 d^2 K^{-1/4}|\mathcal A|^{3/2}$), and $\delta'_{(1)}$ has the same order. Hence $\delta' = \widetilde O\!\left(\frac{H^3 d^2 K^{-1/4}|\mathcal A|^{3/2}}{\gamma}\right)$, and dividing by $\xi = 2H/\gamma$:
\[
\frac{2\delta'}{\xi} \;=\; \widetilde O\!\left(\frac{H^3 d^2 K^{-1/4}|\mathcal A|^{3/2}/\gamma}{H/\gamma}\right) \;=\; \widetilde O\!\big(H^2 d^2 K^{-1/4}|\mathcal A|^{3/2}\big).
\]
Note that the $1/\gamma$ factors cancel between $\delta'$ (which has $H^3/\gamma$) and $\xi$ (which has $H/\gamma$), leaving an $H^2$ scaling without the Slater-gap penalty in the denominator at this stage. The Slater-gap dependence reappears in the final cumulative violation bound below through the factor $K$ that multiplies the per-episode rate, yielding the same $H^3/\gamma$ scaling as in the regret bound.

\textbf{Step 3: Multiplying back to cumulative violation.}
Using the second identity of~\eqref{eq:avg-policy},
\[
\big(b - V_{g,1}^{\pi'}(s_1)\big)_+ 
\;=\; \Big(b - \tfrac{1}{K}\!\sum_{k=1}^K V_{g,1}^{\pi^k}(s_1)\Big)_+ 
\;=\; \tfrac{1}{K}\Big(\sum_{k=1}^K\!\big(b - V_{g,1}^{\pi^k}(s_1)\big)\Big)_+,
\]
which is an \emph{equality} (not Jensen's inequality), since $(\cdot)_+$ is
positively homogeneous and $K>0$. Multiplying~\eqref{eq:avg-violation} by $K$ and substituting the order of $\delta'$ from Step 2:
\begin{align*}
\mathrm{Violation}(K) &= \Big(\sum_{k=1}^K\!\big(b - V_{g,1}^{\pi^k}(s_1)\big)\Big)_+ 
\;\leq\; K\cdot\widetilde O\!\big(H^2 d^2 K^{-1/4}|\mathcal A|^{3/2}\big)\\
&\le \widetilde O\!\left(\frac{H^3 d^{2} K^{3/4} |\mathcal{A}|^{3/2}}{\gamma}\log^{1/4}\!\left(\frac{dK|\Phi|}{p}\right)\right),
\end{align*}
where the final $1/\gamma$ factor reappears because the dominant contribution to $\overline\epsilon_K$ inside $\delta'$ was tracked as $H^2 d^2 K^{-1/4}|\mathcal A|^{3/2}$, and once multiplied by $K$, the original $H^3/\gamma$ scaling of $\sum_k\epsilon_k$ from Lemma~\ref{lem:policygap} is recovered. This matches the violation statement of Theorem~\ref{thm:mfrlc}.

\begin{remark}[Source of the $K^{3/4}$ dependence]\label{rem:rate}
The $K^{3/4}$ term comes only from representation-learning error: the
projected-dual terms and the LSVI estimation error are $\widetilde O(\sqrt K)$,
and the softmax term has no polynomial dependence on $K$.
Algorithm~\ref{alg:replearn} guarantees a squared prediction residual
$\zeta_k=\widetilde O(d^2/\sqrt k)$ (Lemma~\ref{lem:aux-replearn}), which
enters the one-step policy gap through its square root, so
$\sqrt{k\zeta_k}=\widetilde O(dk^{1/4})$. Summing with the elliptical-potential
lemma gives:
\[
\sum_{k=1}^K \sqrt{k\zeta_k}\;\mathbb{E}\big[\|\hat\phi_{h-1}\|_{\Sigma^{-1}_{\rho_{h-1},\hat\phi_{h-1}}}\big]
\;\le\;\widetilde O\big(dK^{1/4}\big)\cdot\widetilde O\big(\sqrt{dK}\big)
\;=\;\widetilde O\big(d^{3/2}K^{3/4}\big).
\]
A sharper representation-learning residual, or an analysis that avoids the
square-root transfer from the squared residual to the policy gap, could
improve the dependence on $K$.
\end{remark}

\section{Auxiliary Lemmas}
\label{app:auxlem}

\begin{lemma}[\citealp{zhang2022efficient}]\label{lem:aux-replearn}
Let $\rho_h^k$ denote the data distribution of $\mathcal{D}_h$ at the start of episode 
$k$ as defined in Section~\ref{subsec:replearn}, and let 
$\hat\phi_h^k = \textsc{RepLearn}(\mathcal{D}_h, \mathcal{D}_h', \Phi_h, \mathcal{V}_h, \lambda_k)$
be the output of Algorithm~\ref{alg:replearn}. For any $v\in\mathcal{V}_h$, define
\[
\hat w_v \;:=\; \arg\min_w \, \mathbb{E}_{(s,\tilde a, s')\sim\mathcal{D}_h}\!\Big[\big(w^\top \hat\phi_h^k(s,\tilde a) - v(s')\big)^2\Big] + \lambda_k \|w\|_2^2.
\]
Fix $\delta\in(0,1)$. Then with probability at least $1-\delta$,
\[
\max_{v\in\mathcal{V}_h}\,\mathbb{E}_{(s,\tilde a)\sim\rho_h^k}\!\left[\left(\hat w_v^\top\hat\phi_h^k(s,\tilde a) - \mathbb{E}_{s'\sim P_h^\star(\cdot|s,\tilde a)}[v(s')]\right)^{\!2}\right] \le \zeta_k = O\!\left(d^2\sqrt{\log(k|\Phi|/\delta)/k}\right).
\]
\end{lemma}

\begin{lemma}[Variance proxy; adapted from \citealp{modi2024model}]\label{lem:aux-varproxy}
Let $\mathcal{D}_h'$ be a buffer of independent samples 
$(s, \tilde a, s'')$ with $s'' \sim P_h^\star(\cdot \mid s, \tilde a)$. For any 
$v: \mathcal{S}\to\mathbb{R}$, $\tilde\phi\in\Phi_h$, and $\tilde w\in\mathbb{R}^d$,
\[
\mathbb{E}_{\mathcal{D}_h'}\!\Big[\big(\tilde w^\top \tilde\phi(s,\tilde a) - v(s'')\big)^2\Big]
\;\geq\;
\mathbb{E}_{(s,\tilde a)\sim\rho_h^k}\!\big[\mathbb{V}_{s'\sim P_h^\star}[v(s')]\big],
\]
with equality when $\tilde\phi$ and $\tilde w$ are chosen optimally. Hence subtracting 
the optimized empirical loss on $\mathcal{D}_h'$ from the empirical loss on $\mathcal{D}_h$
yields an unbiased estimator of the population objective in~\eqref{eq:repobj} (up to
the $\lambda_k\|w\|^2$ regularization).
\end{lemma}

\begin{lemma}[Adapted from \citealp{agarwal2020flambe}]\label{lem:aux-lowrank}
For any low-rank MDP with feature dimension $d$ and any function 
$f: \mathcal{S} \to [0, B]$, there exists $w_f^\star \in \mathbb R^d$ 
with $\|w_f^\star\|_2 \le B\sqrt d$ such that, for all $(s_h, a_h)$,
\[
\mathbb E[f(s_{h+1}) \mid s_h, a_h] = \langle \phi_h^\star(s_h, a_h),\, w_f^\star \rangle.
\]
\end{lemma}

\begin{lemma}[Simulation Lemma; \citealp{agarwal2020flambe}]\label{lem:aux-simulation}
Consider two MDPs $(P', r+b)$ and $(P, r)$. For any policy $\pi$,
\begin{align*}
V_{P',r+b}^\pi - V_{P,r}^\pi
&= \sum_{h=0}^{H-1}\mathbb E_{(s_h,a_h)\sim d^\pi_{P',h}}\!\Big[b_h(s_h,a_h)+\mathbb E_{s'\sim P_h'(\cdot|s_h,a_h)}\![V_{P,r,h+1}^\pi(s')]\\
&\qquad\qquad - \mathbb E_{s'\sim P_h(\cdot|s_h,a_h)}\![V_{P,r,h+1}^\pi(s')]\Big]\\
&= \sum_{h=0}^{H-1}\mathbb E_{(s_h,a_h)\sim d^\pi_{P,h}}\!\Big[b_h(s_h,a_h)+\mathbb E_{s'\sim P_h'(\cdot|s_h,a_h)}\![V_{P,r+b,h+1}^\pi(s')]\\
&\qquad\qquad - \mathbb E_{s'\sim P_h(\cdot|s_h,a_h)}\![V_{P,r+b,h+1}^\pi(s')]\Big].
\end{align*}
\end{lemma}

\begin{lemma}[Elliptical Potential; \citealp{abbasi2011improved}]\label{lem:aux-elliptical}
Let $\{\phi_h^k\}_{k\geq 1}$ be a sequence in $\mathbb{R}^d$ with $\|\phi_h^k\|_2 \le 1$. Define $\Sigma_h^{k+1} := \Sigma_h^1 + \sum_{\tau=1}^k \phi_h^\tau (\phi_h^\tau)^\top$ where the smallest eigenvalue of $\Sigma_h^1$ is at least $1$. Then
\[
\log\!\left[\frac{\det(\Sigma_h^{K+1})}{\det(\Sigma_h^1)}\right] \;\le\; \sum_{k=1}^K (\phi_h^k)^\top (\Sigma_h^k)^{-1} \phi_h^k \;\le\; 2\log\!\left[\frac{\det(\Sigma_h^{K+1})}{\det(\Sigma_h^1)}\right].
\]
\end{lemma}

\begin{lemma}[\citealp{ding2021provably}]\label{lem:aux-strongdual}
Let $Y^\star$ be the optimal dual variable, and let $C \geq 2Y^\star$. If for some policy $\pi$,
\[
V_{r,1}^{\pi^\star}(s_1) - V_{r,1}^\pi(s_1) + C\,[b - V_{g,1}^\pi(s_1)]_+ \le \delta,
\]
then $[b - V_{g,1}^\pi(s_1)]_+ \le 2\delta/C$.
\end{lemma}

\begin{lemma}[\citealp{dani2008stochastic}]\label{lem:aux-detlemma}
Consider the process $M_n = M_{n-1} + G_n$ for $n=1,\dots,N$, with $M_0 = \lambda_0 I$, where each $G_n$ is positive semidefinite with eigenvalues upper-bounded by $1$. Then
\[
2\ln\det(M_N) - 2\ln\det(\lambda_0 I) \;\geq\; \sum_{n=1}^N \mathrm{Tr}(G_n M_{n-1}^{-1}).
\]
\end{lemma}

\newpage
\section*{NeurIPS Paper Checklist}

\begin{enumerate}

\item {\bf Claims}
    \item[] Question: Do the main claims made in the abstract and introduction accurately reflect the paper's contributions and scope?
    \item[] Answer: \answerYes{}
    \item[] Justification: The abstract and introduction state the setting, the algorithm, and the high-probability upper bounds of Theorem~\ref{thm:mfrlc}, together with their scope: deployment regret, oracle-based representation learning, and no matching lower bound. Section~\ref{sec:experiments} states that the experiments use a surrogate that is not covered by the theorem.

\item {\bf Limitations}
    \item[] Question: Does the paper discuss the limitations of the work performed by the authors?
    \item[] Answer: \answerYes{}
    \item[] Justification: Section~\ref{sec:limitations} discusses rate optimality, computational tractability of \textsc{RepLearn}, interaction accounting, the theory--practice gap, the single benchmark with an observed failure mode, and the structural assumptions.

\item {\bf Theory assumptions and proofs}
    \item[] Question: For each theoretical result, does the paper provide the full set of assumptions and a complete (and correct) proof?
    \item[] Answer: \answerYes{}
    \item[] Justification: Definition~\ref{def:lowrank} and Assumptions~\ref{assum:realiz}--\ref{assum:slater} state all assumptions. Theorem~\ref{thm:mfrlc} is proved in Appendices~\ref{app:decomp}--\ref{app:violation}, with auxiliary lemmas in Appendix~\ref{app:auxlem}, and Section~\ref{sec:analysis} gives a proof sketch.

    \item {\bf Experimental result reproducibility}
    \item[] Question: Does the paper fully disclose all the information needed to reproduce the main experimental results of the paper to the extent that it affects the main claims and/or conclusions of the paper (regardless of whether the code and data are provided or not)?
    \item[] Answer: \answerYes{}
    \item[] Justification: Appendix~\ref{app:experiments} specifies the environment, architectures, the contrastive surrogate used in place of Algorithm~\ref{alg:replearn}, the training protocol, hyperparameters (Table~\ref{tab:hyperparams}), and baseline adaptations.

\item {\bf Open access to data and code}
    \item[] Question: Does the paper provide open access to the data and code, with sufficient instructions to faithfully reproduce the main experimental results, as described in supplemental material?
    \item[] Answer: \answerNo{}
    \item[] Justification: Code is not released with the paper. Appendix~\ref{app:experiments} specifies the CLCL environment, architectures, training protocol, and hyperparameters in sufficient detail for reimplementation.

\item {\bf Experimental setting/details}
    \item[] Question: Does the paper specify all the training and test details (e.g., data splits, hyperparameters, how they were chosen, type of optimizer) necessary to understand the results?
    \item[] Answer: \answerYes{}
    \item[] Justification: Appendices~\ref{app:training}--\ref{app:hyperparams} list the training protocol and all hyperparameters.

\item {\bf Experiment statistical significance}
    \item[] Question: Does the paper report error bars suitably and correctly defined or other appropriate information about the statistical significance of the experiments?
    \item[] Answer: \answerYes{}
    \item[] Justification: All curves and table entries report the mean and standard error over 5 independent runs with the seeds listed in Table~\ref{tab:hyperparams}.

\item {\bf Experiments compute resources}
    \item[] Question: For each experiment, does the paper provide sufficient information on the computer resources (type of compute workers, memory, time of execution) needed to reproduce the experiments?
    \item[] Answer: \answerYes{}
    \item[] Justification: Appendix~\ref{app:training} reports the hardware (one NVIDIA RTX 4060, 8GB) used for all experiments.

\item {\bf Code of ethics}
    \item[] Question: Does the research conducted in the paper conform, in every respect, with the NeurIPS Code of Ethics \url{https://neurips.cc/public/EthicsGuidelines}?
    \item[] Answer: \answerYes{}
    \item[] Justification: The work uses only a synthetic simulated environment; no human subjects or sensitive data are involved.

\item {\bf Broader impacts}
    \item[] Question: Does the paper discuss both potential positive societal impacts and negative societal impacts of the work performed?
    \item[] Answer: \answerNA{}
    \item[] Justification: This is foundational work on safe reinforcement learning evaluated in a synthetic environment; we see no direct path to negative societal impact.

\item {\bf Safeguards}
    \item[] Question: Does the paper describe safeguards that have been put in place for responsible release of data or models that have a high risk for misuse (e.g., pre-trained language models, image generators, or scraped datasets)?
    \item[] Answer: \answerNA{}
    \item[] Justification: The paper releases no model or dataset with a high risk of misuse.

\item {\bf Licenses for existing assets}
    \item[] Question: Are the creators or original owners of assets (e.g., code, data, models), used in the paper, properly credited and are the license and terms of use explicitly mentioned and properly respected?
    \item[] Answer: \answerYes{}
    \item[] Justification: All baselines (\textsc{CPO}, \textsc{LAG-PG}, \textsc{HOMER}, \textsc{BRIEE}) are cited with their original papers.

\item {\bf New assets}
    \item[] Question: Are new assets introduced in the paper well documented and is the documentation provided alongside the assets?
    \item[] Answer: \answerNA{}
    \item[] Justification: The paper does not release new assets; the CLCL environment is fully specified in Appendix~\ref{app:setup}.

\item {\bf Crowdsourcing and research with human subjects}
    \item[] Question: For crowdsourcing experiments and research with human subjects, does the paper include the full text of instructions given to participants and screenshots, if applicable, as well as details about compensation (if any)? 
    \item[] Answer: \answerNA{}
    \item[] Justification: The paper does not involve crowdsourcing or human subject experiments.

\item {\bf Institutional review board (IRB) approvals or equivalent for research with human subjects}
    \item[] Question: Does the paper describe potential risks incurred by study participants, whether such risks were disclosed to the subjects, and whether Institutional Review Board (IRB) approvals (or an equivalent approval/review based on the requirements of your country or institution) were obtained?
    \item[] Answer: \answerNA{}
    \item[] Justification: No human subjects are involved.

\item {\bf Declaration of LLM usage}
    \item[] Question: Does the paper describe the usage of LLMs if it is an important, original, or non-standard component of the core methods in this research? Note that if the LLM is used only for writing, editing, or formatting purposes and does \emph{not} impact the core methodology, scientific rigor, or originality of the research, declaration is not required.
    \item[] Answer: \answerNA{}
    \item[] Justification: LLMs are not a component of the proposed method or the experiments.

\end{enumerate}

\end{document}